\pdfoutput=1
\RequirePackage[T1]{fontenc}
\documentclass[12pt]{article}
\usepackage[skip=1em plus 0.2em minus 0.1em]{parskip}
\usepackage[height=8.85in,width=6.45in]{geometry}

\usepackage[utf8]{inputenc}
\usepackage{amsmath}
\usepackage{amssymb}
\usepackage{mathtools}
\numberwithin{equation}{section}
\usepackage{slashed}
\usepackage{braket}
\usepackage{enumitem}
\usepackage[svgnames]{xcolor}
\usepackage[colorlinks,citecolor=DarkGreen,linkcolor=FireBrick,urlcolor=FireBrick,linktocpage,unicode]{hyperref}

\usepackage{tocloft}

\usepackage{footmisc}

\allowdisplaybreaks
\usepackage{graphicx}
\usepackage{tikz}
\usetikzlibrary{arrows.meta,calc,backgrounds,shapes.geometric}
\usepackage{tikz-cd}
\usepackage{times}
 
\usepackage{bm}
\usepackage{physics}
\usepackage{xcolor}
\usepackage{natbib}
\usepackage{mdframed}
\usepackage{nicefrac}
\usepackage{booktabs}
\usepackage{lipsum}
\usepackage{titlesec}
\usepackage{wrapfig}
\usepackage{authblk}
\usepackage{blindtext}
\usepackage[font=small]{caption}

\usepackage{algorithm}
\usepackage{algpseudocode}
\newcommand{\commentsymbol}{$\triangleright$}
\algrenewcommand\algorithmiccomment[1]{\hfill {\footnotesize \commentsymbol{} #1}}

\usepackage{titletoc}
\usepackage{todonotes}
\usepackage{setspace}
\usepackage{dsfont} 
\newcommand{\one}{\mathds{1}}

\usepackage[nameinlink,capitalize,noabbrev]{cleveref}

\usepackage{CJK}

\usepackage{array}
\usepackage{bbm}
\usepackage{makecell}

\usepackage{dashbox}
\usepackage{xcolor}
\usepackage{colortbl}
\definecolor{lightyellow}{rgb}{1.0, 0.95, 0.7}
\definecolor{Blue}{rgb}{0, 0, 0.8}
\definecolor{blue}{rgb}{0,0,1}
\definecolor{darkgreen}{rgb}{0,0.40,0}
\definecolor{firebrick}{rgb}{0.698,0.133,0.133}
\newcommand*{\red}[1]{\textcolor{red}{#1}}

\newcommand*{\blue}[1]{\textcolor{blue}{#1}}

\definecolor{colorA}{rgb}{1,0,0}
\definecolor{colorB}{rgb}{0,0.3,1}
\definecolor{colorC}{rgb}{0.9,0.8,0.2}
\definecolor{colorD}{rgb}{0,0.65,0}
\definecolor{lesslightgray}{rgb}{0.5,0.5,0.5}
\definecolor{light-gray}{gray}{0.95}

\let\tilde\widetilde
\let\hat\widehat

\newcommand{\argmin}{\mathop{\mathrm{argmin}}}

\def\R{\mathbb{R}}

\let\cite\citep 

\usepackage{amsthm}
\usepackage[many]{tcolorbox}
\usepackage{etoolbox}
\newlength{\legacytheoremtopsep}
\newcommand{\localtheoremlistskips}{%
  \setlength{\topsep}{\legacytheoremtopsep}%
  \setlength{\partopsep}{\topsep}%
}
\makeatletter
\newcommand{\template@registermathspacing}[1]{%
  \AddToHook{env/#1/before}[template-local-spacing]{%
    \edef\template@outerlistdepth{\number\@listdepth}%
  }%
  \DeclareHookRule{env/#1/before}{template-local-spacing}{before}{tcolorbox}%
  \AtBeginEnvironment{#1}{%
    \ifnum\template@outerlistdepth=0\relax
      \localtheoremlistskips
    \fi
  }%
}
\forcsvlist{\template@registermathspacing}{theorem,lemma,corollary,proposition,definition,assumption,remark,problem,property,hypothesis,fact,example,question,claim,proof}
\makeatother
\makeatletter
\newcommand{\template@titlelistskips}{%
  \ifdefstring{\template@parentenv}{titlepage}{\localtheoremlistskips}{}%
}
\BeforeBeginEnvironment{center}{\let\template@parentenv\@currenvir}
\AtBeginEnvironment{center}{\template@titlelistskips}
\BeforeBeginEnvironment{flushright}{\let\template@parentenv\@currenvir}
\AtBeginEnvironment{flushright}{\template@titlelistskips}
\makeatother
\BeforeBeginEnvironment{proof}{\par\vspace{-1\baselineskip}}
\AfterEndEnvironment  {proof}{\par\vspace{-1.5\baselineskip}}

\newtheoremstyle{theoremstyle}
  {.5\baselineskip} 
  {.5\baselineskip} 
  {}                  
  {}                  
  {\bfseries}        
  {.}                 
  {1em}               
  {}                  

\theoremstyle{theoremstyle}
\newtheorem{theorem}{Theorem}[section]
\newtheorem{lemma}{Lemma}[section]
\newtheorem{corollary}{Corollary}[theorem]
\newtheorem{proposition}{Proposition}[section]
\newtheorem{definition}{Definition}[section]

\newtheorem{remark}{Remark}[section]

\tcolorboxenvironment{theorem}{
  breakable,
  colback=black!10,
  colframe=white,%
  width=\dimexpr\linewidth+10pt\relax,%
  enlarge left by=-5pt,%
  enlarge right by=-5pt,%
  boxsep=5pt,%
  boxrule=0pt,
  left=0pt,right=0pt,top=0pt,bottom=0pt,
  sharp corners,
  before skip=0.5\baselineskip, 
  after skip=0.5\baselineskip,  
  fonttitle=\bfseries, 
  coltitle=black 
}
\tcolorboxenvironment{remark}{
  blanker,
  breakable,
  before skip=.8\baselineskip,  
  after  skip=.8\baselineskip   
}

\tcolorboxenvironment{proposition}{
  breakable,
  colback=black!10,
  colframe=white,%
  width=\dimexpr\linewidth+10pt\relax,%
  enlarge left by=-5pt,%
  enlarge right by=-5pt,%
  boxsep=5pt,%
  boxrule=0pt,
  left=0pt,right=0pt,top=0pt,bottom=0pt,
  sharp corners,
  before skip=0.5\baselineskip, 
  after skip=0.5\baselineskip,  
  fonttitle=\bfseries, 
  coltitle=black 
}

\tcolorboxenvironment{lemma}{
  breakable,
  colback=black!10,
  colframe=white,%
  width=\dimexpr\linewidth+10pt\relax,%
  enlarge left by=-5pt,%
  enlarge right by=-5pt,%
  boxsep=5pt,%
  boxrule=0pt,
  left=0pt,right=0pt,top=0pt,bottom=0pt,
  sharp corners,
  before skip=0.5\baselineskip, 
  after skip=0.5\baselineskip,  
  fonttitle=\bfseries, 
  coltitle=black 
}

\tcolorboxenvironment{corollary}{
  breakable,
  colback=black!10,
  colframe=white,%
  width=\dimexpr\linewidth+10pt\relax,%
  enlarge left by=-5pt,%
  enlarge right by=-5pt,%
  boxsep=5pt,%
  boxrule=0pt,
  left=0pt,right=0pt,top=0pt,bottom=0pt,
  sharp corners,
  before skip=0.5\baselineskip, 
  after skip=0.5\baselineskip,  
  fonttitle=\bfseries, 
  coltitle=black 
}

\tcolorboxenvironment{definition}{
  breakable,
  colback=black!10,
  colframe=white,%
  width=\dimexpr\linewidth+10pt\relax,%
  enlarge left by=-5pt,%
  enlarge right by=-5pt,%
  boxsep=5pt,%
  boxrule=0pt,
  left=0pt,right=0pt,top=0pt,bottom=0pt,
  sharp corners,
  before skip=0.5\baselineskip, 
  after skip=0.5\baselineskip,  
  fonttitle=\bfseries, 
  coltitle=black 
}
\tcolorboxenvironment{assumption}{
  breakable,
  colback=black!10,
  colframe=white,%
  width=\dimexpr\linewidth+10pt\relax,%
  enlarge left by=-5pt,%
  enlarge right by=-5pt,%
  boxsep=5pt,%
  boxrule=0pt,
  left=0pt,right=0pt,top=0pt,bottom=0pt,
  sharp corners,
  before skip=0.5\baselineskip, 
  after skip=0.5\baselineskip,  
  fonttitle=\bfseries, 
  coltitle=black 
}

\crefname{theorem}{Theorem}{Theorems}
\crefname{proposition}{Proposition}{Propositions}
\crefname{lemma}{Lemma}{Lemmas}
\crefname{corollary}{Corollary}{Corollaries}
\crefname{definition}{Definition}{Definitions}
\crefname{assumption}{Assumption}{Assumptions}
\crefname{remark}{Remark}{Remarks}
\crefname{problem}{Problem}{Problems}
\crefname{property}{Property}{property}

\tcolorboxenvironment{hypothesis}{
  breakable,
  colback=black!10,
  colframe=white,%
  width=\dimexpr\linewidth+10pt\relax,%
  enlarge left by=-5pt,%
  enlarge right by=-5pt,%
  boxsep=5pt,%
  boxrule=0pt,
  left=0pt,right=0pt,top=0pt,bottom=0pt,
  sharp corners,
  before skip=0.5\baselineskip, 
  after skip=0.5\baselineskip,  
  fonttitle=\bfseries, 
  coltitle=black 
}
\crefname{hypothesis}{Hypothesis}{Hypothesises}

\tcolorboxenvironment{fact}{
  breakable,
  colback=black!10,
  colframe=white,%
  width=\dimexpr\linewidth+10pt\relax,%
  enlarge left by=-5pt,%
  enlarge right by=-5pt,%
  boxsep=5pt,%
  boxrule=0pt,
  left=0pt,right=0pt,top=0pt,bottom=0pt,
  sharp corners,
  before skip=0.5\baselineskip, 
  after skip=0.5\baselineskip,  
  fonttitle=\bfseries, 
  coltitle=black 
}
\crefname{fact}{Fact}{Facts}

\newtheorem{example}{Example}
\tcolorboxenvironment{example}{
  breakable,
  colback=black!10,
  colframe=white,%
  width=\dimexpr\linewidth+10pt\relax,%
  enlarge left by=-5pt,%
  enlarge right by=-5pt,%
  boxsep=5pt,%
  boxrule=0pt,
  left=0pt,right=0pt,top=0pt,bottom=0pt,
  sharp corners,
  before skip=0.5\baselineskip, 
  after skip=0.5\baselineskip,  
  fonttitle=\bfseries, 
  coltitle=black 
}
\crefname{example}{Example}{Examples}

\tcolorboxenvironment{question}{
  breakable,
  colback=black!10,
  colframe=white,%
  width=\dimexpr\linewidth+10pt\relax,%
  enlarge left by=-5pt,%
  enlarge right by=-5pt,%
  boxsep=5pt,%
  boxrule=0pt,
  left=0pt,right=0pt,top=0pt,bottom=0pt,
  sharp corners,
  before skip=0.5\baselineskip, 
  after skip=0.5\baselineskip,  
  fonttitle=\bfseries, 
  coltitle=black 
}

\crefname{question}{Question}{Questions}
\numberwithin{equation}{section}
\numberwithin{theorem}{section}
\numberwithin{proposition}{section}
\numberwithin{definition}{section}
\numberwithin{lemma}{section}
\numberwithin{assumption}{section}
\numberwithin{remark}{section}

\newcommand*{\annot}[1]{\tag*{\footnotesize{\textcolor{black!50}{\big(#1\big)}}}}

\makeatletter
\let\save@mathaccent\mathaccent
\newcommand*\if@single[3]{%
    \setbox0\hbox{${\mathaccent"0362{#1}}^H$}%
    \setbox2\hbox{${\mathaccent"0362{\kern0pt#1}}^H$}%
    \ifdim\ht0=\ht2 #3\else #2\fi
}
\newcommand*\rel@kern[1]{\kern#1\dimexpr\macc@kerna}
\newcommand*\widebar[1]{\@ifnextchar^{{\wide@bar{#1}{0}}}{\wide@bar{#1}{1}}}
\newcommand*\wide@bar[2]{\if@single{#1}{\wide@bar@{#1}{#2}{1}}{\wide@bar@{#1}{#2}{2}}}
\newcommand*\wide@bar@[3]{%
    \begingroup
    \def\mathaccent##1##2{%
        \let\mathaccent\save@mathaccent
        \if#32 \let\macc@nucleus\first@char \fi
        \setbox\z@\hbox{$\macc@style{\macc@nucleus}_{}$}%
        \setbox\tw@\hbox{$\macc@style{\macc@nucleus}{}_{}$}%
        \dimen@\wd\tw@
        \advance\dimen@-\wd\z@
        \divide\dimen@ 3
        \@tempdima\wd\tw@
        \advance\@tempdima-\scriptspace
        \divide\@tempdima 10
        \advance\dimen@-\@tempdima
        \ifdim\dimen@>\z@ \dimen@0pt\fi
        \rel@kern{0.6}\kern-\dimen@
        \if#31
        \overline{\rel@kern{-0.6}\kern\dimen@\macc@nucleus\rel@kern{0.4}\kern\dimen@}%
        \advance\dimen@0.4\dimexpr\macc@kerna
        \let\final@kern#2%
        \ifdim\dimen@<\z@ \let\final@kern1\fi
        \if\final@kern1 \kern-\dimen@\fi
        \else
        \overline{\rel@kern{-0.6}\kern\dimen@#1}%
        \fi
    }%
    \macc@depth\@ne
    \let\math@bgroup\@empty \let\math@egroup\macc@set@skewchar
    \mathsurround\z@ \frozen@everymath{\mathgroup\macc@group\relax}%
    \macc@set@skewchar\relax
    \let\mathaccentV\macc@nested@a
    \if#31
    \macc@nested@a\relax111{#1}%
    \else
    \def\gobble@till@marker##1\endmarker{}%
    \futurelet\first@char\gobble@till@marker#1\endmarker
    \ifcat\noexpand\first@char A\else
    \def\first@char{}%
    \fi
    \macc@nested@a\relax111{\first@char}%
    \fi
    \endgroup
    }
\makeatother
\let\bar\widebar

\usepackage{paper_metadata}
\PaperTitle{A Quantum Duality for Grokking Transformers}

\PaperAffiliation{vanderbilt}{%
  Department of Computer Science, College of Connected Computing, Vanderbilt University, USA}
\PaperAffiliation{ntu_physics}{%
  Department of Physics, National Taiwan University, Taipei, Taiwan}
\PaperAffiliation{nthu_physics}{%
  Department of Physics, National Tsing Hua University, Hsinchu, Taiwan}

\PaperTitleNote{
Code is available at \url{https://github.com/yuanchen2045/grokking_etoc}}

\PaperAuthor[affiliations={vanderbilt,ntu_physics},
  email={b09202059@ntu.edu.tw}]{Yu-An Chen}
\PaperAuthor[affiliations={nthu_physics},
  email={henryliao.physics@gmail.com}]{Henry Liao}
\PaperAuthor[affiliations={vanderbilt},
  email={jhu.vandy@gmail.com}]{Jerry Yao-Chieh Hu}

\setlist[itemize,enumerate]{
  parsep=\parskip,                                   
  itemsep=\dimexpr .3em - \parskip\relax plus 2pt,   
  topsep=\dimexpr 6pt - \parskip\relax plus 1pt minus 1pt,
  partopsep=0pt,
  listparindent=\parindent
}

\begin{document}
\begin{titlepage}

\begin{flushright}
Last Update: \today
\end{flushright}

\vskip 2.5em
\begin{center}

{
\LARGE \bfseries 
\begin{spacing}{1.15} 
\PrintPaperTitle
\end{spacing}
}

\vskip 1em
\PrintPaperAuthors

\vskip 1em

\PrintPaperAffiliations
\PrintPaperNotes


\end{center}

\noindent

We present a quantum duality for Transformers, and use it to study Grokking.
This duality originates from the exact correspondence between the kernel formulation of Transformers and coherent-state representation of bosonic quantum dynamical systems.
We highlight its usefulness through a quantum-chaos-style analysis of Grokking in Transformer training dynamics.
In particular, we introduce \underline{E}qual-\underline{T}ime-\underline{O}rdered \underline{C}orrelator (ETOC) to characterize the memorization-generalization transitions in Grokking.
Our key insight is - ETOC predicts and locates Grokking during training, with  key-key changes \textit{right before} Grokking and query-key changes \textit{right after} Grokking. 
This provides precise characterizations and effective detections of Grokking in real time.
Numerical results on modular addition and multiplication tasks corroborate with our theory.

\begin{figure}[H]
\vspace{.5em}
    \centering
    \includegraphics[width=0.85\linewidth]{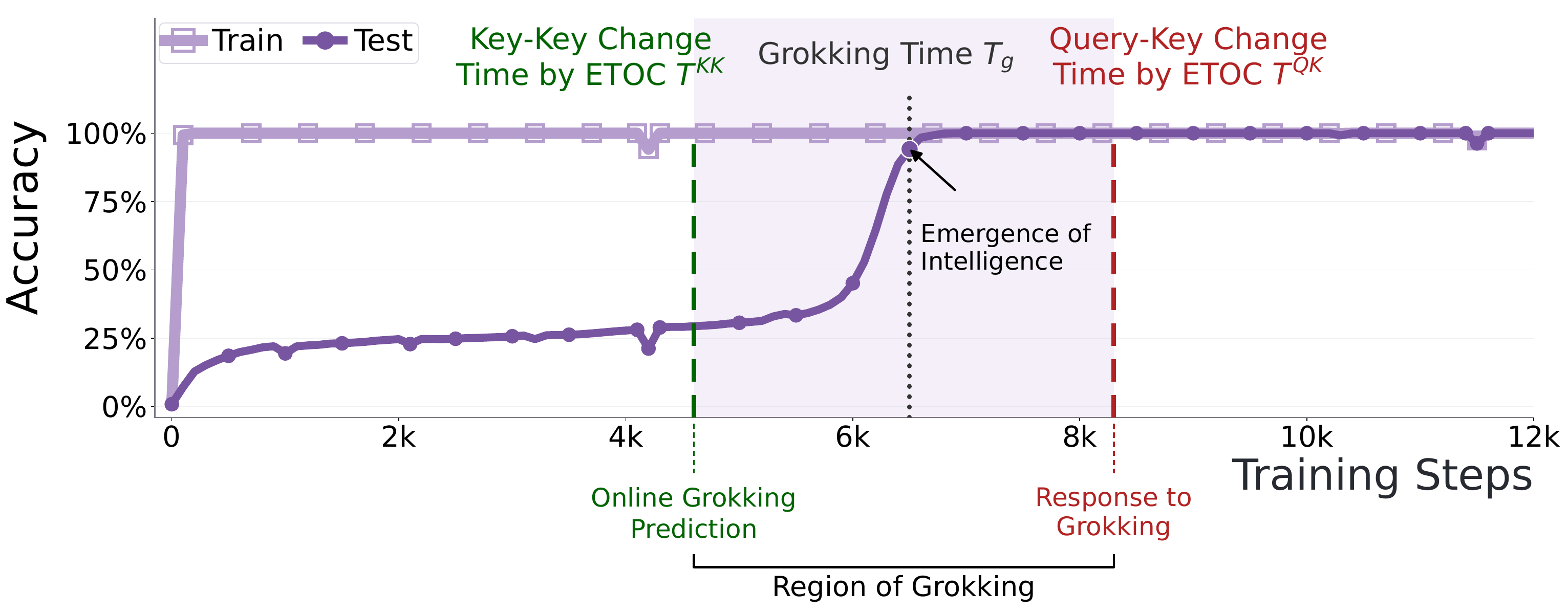}
    \vspace{-.5em}
    \caption{We introduce ETOCs to predict and locate Grokking during Transformer training.}
    \label{fig:grok_abstract}
\end{figure}

\vfill

    
    
    
    

\end{titlepage}

{
\setlength{\parskip}{0em}
\setcounter{tocdepth}{2}
\tableofcontents
}
\setcounter{footnote}{0}


\section{Introduction}
\label{sec:intro}

\begin{figure}[t]
    \centering
    \includegraphics[width=1\linewidth]{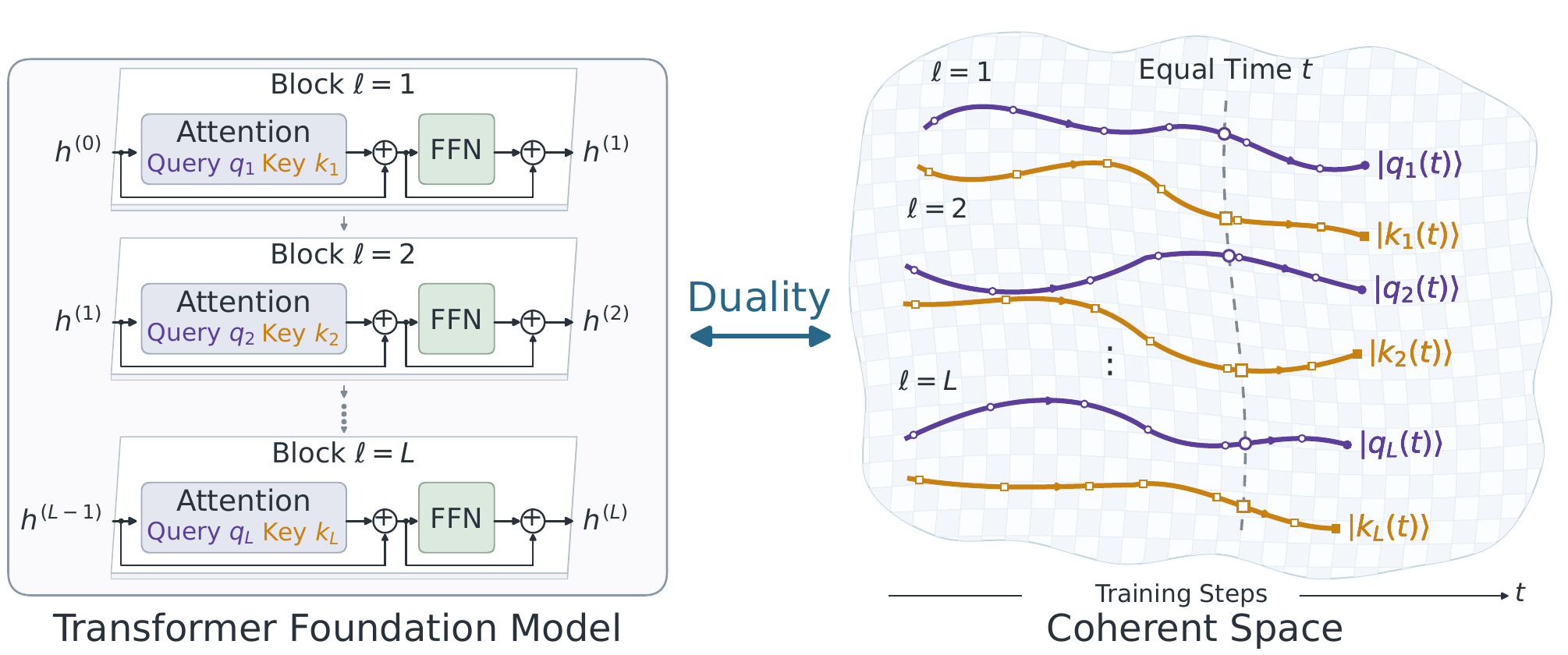}
    \caption{A schematic sketch of the duality between kernel representations of Transformers and coherent-state representations in bosonic quantum systems.}
    \label{fig:intro_schematic}
\end{figure}

We establish a quantum duality for Transformers and use this formulation to study Grokking.
This duality maps Transformer training dynamics to time evolution trajectories from a quantum dynamical system.
Importantly, this invites migration of physics ideas from quantum chaos into quantitative tools for understanding Transformer training.
In this paper, we study Grokking as an example.
Specifically, we introduce \underline{E}qual-\underline{T}ime-\underline{O}rdered \underline{C}orrelator (ETOC) to quantify ordering effects in quantum representations at each training step $t$. 
The time evolution of these ETOCs
characterizes the transition from memorization to generalization and predicts Grokking.
We sketch this idea in \cref{fig:intro_schematic}.

Recent work on Grokking raises two questions: how learned representations characterize generalization and how training dynamics anticipate the transition \cite{power2022grokking}. 
To explain Grokking, \citet{nanda2023progress} verify a Transformer's learned rule for $(a+b)\bmod p$ and track the rule's formation alongside the drop in test loss.
Besides, \citet{kumar2024grokking} connect the delay in generalization to feature learning. 
To predict Grokking, \citet{tan2023understanding} propose indicators from representation responses to small input changes, while \citet{xu2026early}  use a sharp increase in parameter sensitivity due to gradient-update ordering as an early warning of Grokking.
However, these approaches leave open how the relation of queries and keys characterizes, predicts Grokking. 

To address these issues, we apply the quantum duality to define query-key and key-key ETOCs throughout Transformer training dynamics. 
Specifically, query-key ETOCs measure ordering effects between collective changes in queries and keys, while key-key ETOCs measure ordering effects among distinct keys. 
We evaluate these correlators in single-block Transformers with multi-head self-attention on $(a+b)\bmod p$ and $(ab)\bmod p$ tasks. 
Across the two tasks, early change-time estimates place the key-key change right before Grokking and the query-key change right after Grokking. 
Together, the earlier key-key change predicts the approach of Grokking, and the two change times locate the transition from memorization to generalization.

\paragraph{Contributions.}
We establish a quantum duality for Transformers and introduce quantum-chaos-inspired ETOCs to characterize, predict, and locate Grokking. 
Our main contributions are:

\begin{itemize}

    \item \textbf{An Exact Quantum Duality For Transformers.}
    We establish an exact duality between the kernel formulation of attention and coherent-state representations of bosonic quantum systems.
    Specifically, we represent queries and keys as coherent states, recover the attention (\cref{thm:quantum_softmax_rbf}), and discuss its existence and non-uniqueness (\cref{coro:exist_uniqueness}). 
    Through this perspective, attention describes how a query selects and combines stored information. 
    This duality extends across attention heads and Transformer layers (\cref{def:moyal_weyl_transformer,sec:3attention}).
    
    \item \textbf{Characterizing Memorization and Generalization.}
    We introduce \underline{E}qual-\underline{T}ime-\underline{O}rdered \underline{C}orrelators (ETOCs) to provide a precise characterization of attention dynamics from memorization to generalization. 
    Specifically, query-key ETOCs quantify ordering effects between collective changes in queries and keys, while key-key ETOCs quantify ordering effects among distinct keys (\cref{def:query_key_etoc,def:key_key_etoc})\footnote{For quantum objects $\hat{A}$ and $\hat{B}$, the commutator $(\hat{A}\hat{B}-\hat{B}\hat{A})$ measures how reversing their order changes the result. These objects are called operators and are defined in \cref{def:operator}.}.
    
    \item \textbf{Predicting and Locating Grokking in Real Time.}
    We use ETOC change times to predict Grokking and locate the transition from memorization to generalization (\cref{def:etoc_signal_change_time}). 
    Our experiments on $(a+b)\bmod p$ and $(ab)\bmod p$ show: key-key changes occur right before Grokking, while query-key changes occur right after Grokking (\cref{fig:addition,fig:addition-key-key,fig:multiplication,fig:multiplication-key-key}). 
    We assess the stability of these timing estimates during training (\cref{fig:addition-horizon,fig:multiplication-horizon}).
    
\end{itemize}

\paragraph{Related Work.}
Kernel formulations express Transformer attention through inner products between feature vectors \cite{vaswani2017attention,Aronszajn1950TheoryOR,choromanski2020rethinking}. 
On the other hand, coherent-state methods construct kernels from quantum-state inner products \cite{ali:1995zz,schuld2019quantum,schuld2021supervised,abanov2025quantum}. 
This dual structure underlies our quantum duality for Transformers.
Beyond kernel correspondence, quantum-chaos research probes dynamics through ordering effects between quantum objects.
In particular, out-of-time-ordered correlators (OTOCs) quantify these effects and describe how small perturbations spread during time evolution in information scrambling and quantum chaos \cite{maldacena2016bound,rozenbaum2017lyapunov,xu2024scrambling}. 
Drawing on this approach, we introduce ETOCs and characterize how Transformer representations reorganize during Grokking. 
We discuss more related work in \cref{sec:related_work}.

\paragraph{Organization.}
\cref{sec:2background} introduces the kernel-quantum duality. 
\cref{sec:3attention} develops the quantum formulation of Transformers. 
\cref{sec:4grok} defines Grokking and the ETOCs. 
\cref{sec:5exp} examines how ETOCs characterize, locate, and anticipate Grokking in experiments. 
\cref{sec:conclusion} discusses the implications and limitations.
Finally, we defer related work, a brief review on Hilbert space and algebra, Transformer training and detailed proofs in \cref{sec:related_work,sec:brief_review_hilbert_space,sec:transformer_training,sec:proof}.

\section{Preliminaries}
\label{sec:2background}

In this section, we establish the duality between kernel formulation and quantum theory.
Then we consider Moyal-Weyl coherent space as a concrete example to recover Gaussian kernel structures.

\subsection{Kernel-Quantum Space Correspondence}

We begin with the reproducing kernel Hilbert space (RKHS) viewpoint.
We then show how the same structure emerges through quantum states and quantization in quantum theory perspective.

\begin{definition}[Reproducing Kernel Hilbert Space \cite{Aronszajn1950TheoryOR,berlinet2004kernel}]
\label{def:reproducing_kernel_rkhs}
    Let $\mathcal{X}$ be a non-empty set.
    Then
    \begin{itemize}

        \item \textbf{Reproducing kernel.}
        A function $K:\mathcal{X}\times\mathcal{X}\to\mathbb{C}$ is a reproducing kernel of the Hilbert space $\mathcal{H}_K$ if and only if 
        \begin{align}
        \label{eq:reproducing_kernel}
            K(\cdot,x)\in\mathcal{H}_K\quad\text{ and }\quad \langle K(\cdot, x),f\rangle_{\mathcal{H}_K}=f(x),
            \quad \forall x\in\mathcal{X},\quad\forall f\in\mathcal{H}_K,
        \end{align}
        where $\langle\cdot,\cdot\rangle_{\mathcal{H}}$ denotes the inner product in $\mathcal{H}$.
        \eqref{eq:reproducing_kernel} is called the reproducing property.

        \item \textbf{Reproducing kernel Hilbert space (RKHS).}
        A Hilbert space $\mathcal{H}_K$ with a reproducing kernel is a RKHS.
    \end{itemize}
\end{definition}

\begin{remark}
    By Riesz representation theorem (\cref{prp:riesz_rep_thm}), RKHS is also a Hilbert space with bounded functions $f(x)$ (\cref{def:bounded}) for all $f\in\mathcal{H}_K$. 
\end{remark}

In the following, we establish an alternative but equivalent perspective via the operator formalism.

\begin{definition}[Operator \cite{vonneumann1927mathematische}]
\label{def:operator}
    Given a Hilbert space $\mathcal{H}$, we say $\hat{A}:\text{Dom}(\hat{A})\subseteq\mathcal{H}\to\mathcal{H}$ is an operator if it is a linear map. 
    We use hat $\widehat{\hspace{4mm}}$ to denote an operator.
    Given a basis in $\mathcal{H}$, we are then able to express an operator as a matrix. 
    This is how physicists establish a quantum theory  \cite{Griffiths_Schroeter_2018,Sakurai_Napolitano_2020}.
    We use $\text{Mat}(\mathcal{H})$ for operators in matrix representation on $\mathcal{H}$.
\end{definition}

In quantum theory, promoting scalar physical quantities to operators in a Hilbert space is necessary.
It is the minimal mathematical step required to experiments.
In particular, classical/scalar observables do not explain the interference pattern in double-slit experiments, or the incompatibility of position and momentum measurements due to the Heisenberg uncertainty principle \cite{Griffiths_Schroeter_2018,Sakurai_Napolitano_2020,von1955mathematical}.

\begin{remark}
    In quantum theory, physical observables are Hermitian operators, i.e., $\hat{A}^\dagger\coloneq \overline{(\hat{A}^\top)}=\hat{A}$ for an operator $\hat{A}$.
    Their matrix representations are basis-dependent. 
    However, in the infinite-dimensional case, we further require $\text{Dom}(\hat{A})=\text{Dom}(\hat{A}^\dagger)$ due to the boundedness issue \cite{bcHall2020quantumformath}, where $\text{Dom}(\hat{A})$ denotes the domain of $\hat{A}$.
    For more details, please see \cref{sec:brief_review_hilbert_space}.
\end{remark}

To construct the operator formalism concretely, we define several maps as follows.

\begin{definition}[Quantum-State Map and Projection \cite{ali:1995zz,Steinacker:2024unq}]
\label{def:quantum_state_map}
    Let $\mathcal{H}$ be a complex Hilbert space and $\mathcal{X}$ be a non-empty set.  
    Then we define
    \begin{align}
        \text{Quantum-State Map:}&\qquad\psi:\mathcal{X}\to\mathcal{H},\qquad &&x\mapsto\ket{x}\in\mathcal{H},\\
        \text{Projection Map (Projector):}&\qquad \hat{P}:\mathcal{X}\to\mathcal{B}(\mathcal{H}),\quad &&x\mapsto \hat{P}_x\coloneq\ket{x}\bra{x},
    \end{align}
    where $\mathcal{B}(\mathcal{H})$ denotes the space of bounded operators of $\mathcal{H}$ (\cref{def:bounded}), $\hat{\hspace{2mm}}$ denotes matrices, and we define the dual vector of normalized $\ket{x}\in\mathcal{H}$ as the bra state
    \begin{align}
    \label{eq:bra}
        \bra{x}\coloneq\ket{x}^\dagger.
    \end{align} 
    
\end{definition}

From the above definitions and notations, we connect quantum theory with RKHS as follows. 

\begin{proposition}[Quantum-RKHS Duality \cite{ali:1995zz}]
\label{prp:coherent-rkhs}
    Let $x\mapsto\ket{x}\in\mathcal{H}$ be a quantum-state map. 
    We define the complex kernel  via the inner product, i.e., $K_{\mathcal{H}}(x,y)\coloneq\bra{x}\ket{y}$.  Then $K_{\mathcal{H}}$ is a Hermitian positive semi-definite (PSD) kernel with the identification
    \begin{align}
    \label{eq:coherent_rkhs}
        K_{\mathcal{H}}(x,y)=\bra{x}\ket{y}=\langle K(\cdot,x),K(\cdot,y)\rangle_{\mathcal{H}_K}.
    \end{align}  
    For projection map, we have a real PSD kernel
    \begin{align}
        K_{\text{proj}}(x,y)\coloneq\Tr_{\mathcal{H}}(\hat{P}^\dagger_x\hat{P}_y)=|K_{\mathcal{H}}(x,y)|^2.
    \end{align}
    Constructing $\Tr(\cdot)$ is crucial for showing the quantum duality explicitly. 
    We leave the detailed discussion and normalization issue in \cref{prp:normalization_trace}.
\end{proposition}

\begin{proof}
    For arbitrary $c_1,\ldots,c_N\in\mathbb{C}$ and $x_1,\ldots,x_N\in\mathcal{X}$, we have
    \begin{align*}
        \sum_{i,j=1}^{N}\bar{c}_ic_j K_{\mathcal{H}}(x_i,x_j)=\sum_{i,j=1}^{N}\bar{c}_i c_j\bra{x_i}\ket{x_j}=\Big\|\sum_{i=1}^{N}c_i\ket{x_i}\Big\|_{\mathcal{H}}^2\geq 0.
    \end{align*}
    It holds similar to $\sum_{i,j=1}^N\bar{c}_ic_j\Tr_{\mathcal{H}}(\hat{P}_i\hat{P}_j)$. 
    Therefore, $K_{\mathcal{H}}$ is Hermitian PSD.
    This follows from the definition of complex inner product naturally.
    For projection maps, we compute
    \begin{align*}
        \Tr_{\mathcal{H}}(\hat{P}_x\hat{P}_y)=\Tr_{\mathcal{H}}(\ket{x}\bra{x}\ket{y}\bra{y})=\bra{x}\ket{y}\bra{y}\ket{x}=|K_{\mathcal{H}}(x,y)|^2.
    \end{align*}
    Hence, it is a real PSD kernel.
    This completes the proof.
\end{proof}



\subsection{Weyl Quantization and Moyal-Weyl Coherent States}

In this section, we establish the coherent-state/RKHS duality on the Moyal-Weyl quantum plane.
We adopt this example throughout the paper.
In particular, the inner product of coherent states recover an RBF kernel, while its phase depends on the geometry itself.

\begin{definition}[Weyl Quantization Map \cite{weyl1927quantenmechanik}]
\label{def:weyl_quantization}
    A Weyl map on a manifold $\R^{2d}$ is a quantization map $\hat{\mathcal{Q}}:C^\infty(\R^{2d})\to\text{Mat}(\mathcal{H})$ with the form 
    \begin{align}
        \hat{\mathcal{Q}}(f)\coloneq\int_{\R^{2d}}\tilde{f}(k)e^{ik^\top \hat{M}}\frac{\dd^{2d}k}{(2\pi)^{2d}}
        \quad\text{with}
        \quad
        \tilde{f}(k)\coloneq\int_{\R^{2d}}f(x)e^{-ik^\top x}~\dd^{2d}x,
    \end{align}
    where $\text{Mat}(\mathcal{H})$ denotes the matrix representation of operator on $\mathcal{H}$, and $k^\top\hat{M}\coloneq\sum_{a=1}^{2d}k_a\hat{M}^a$ with each element $\hat{M}^a$ being an operator (\cref{def:operator}).
\end{definition}

\begin{remark}
    Here \textit{quantization} refers to the passage from commutative scalar functions to non-commutative matrices, i.e., $fg=gf$ but $\hat{\mathcal{Q}}(f)\hat{\mathcal{Q}}(g)\neq\hat{\mathcal{Q}}(g)\hat{\mathcal{Q}}(f)$ in general.
    Weyl quantization is one useful way to realize this non-commutativity.
    There exist several quantization maps that lead to distinct quantum theory in the same Hilbert space $\mathcal{H}$, but they all coincide in the classical (commutative) limit \cite{berezin1972covariant,berezin1974quantization,perelomov1977coherent,Steinacker:2024unq}.
\end{remark}


\begin{definition}[Moyal-Weyl Quantum Plane \cite{moyal1949quantum,weyl1927quantenmechanik,Steinacker:2024unq}]
\label{def:moyal_weyl_plane}
    A Moyal-Weyl quantum plane $\mathcal{X}=\R^{2d}_\theta$ is the space with coordinates $\{\hat{M}^a\}_{a=1}^{2d}$, where $\hat{M}^a\in\text{Mat}(\mathcal{H})$ satisfying
    \begin{align}
    \label{eq:moyal_weyl_commutation}
        \big[\hat{M}^a,\hat{M}^b\big]=i\theta^{ab}\hat{\one},
    \end{align}
    $\{\hat{M}^a\}_{a=1}^{2d}$ are Hermitian, $[\hat{M}^a,\hat{M}^b]\coloneq \hat{M}^a\hat{M}^b-\hat{M}^b\hat{M}^a$,  $\theta^{ab}$ is invertible real anti-symmetric, and $\hat{\one}$ is the identity operator.
    We further block diagonalize $\theta^{ab}$ and assign the following coordinates for each block,
    \begin{align}
    \label{eq:block_diagonalize_theta}
        O^\top\theta O=\bigoplus_{i=1}^d\begin{pmatrix}0&\theta_i\\-\theta_i&0\end{pmatrix},\qquad\text{with }~(\hat{Q}^i,\hat{P}^i)\coloneq\big((O^\top \hat{M})^{2i-1},(O^\top \hat{M})^{2i}\big),
    \end{align}
    where $\theta_i>0$, and $O$ is a $2d$-dimensional orthogonal matrix.
    These coordinates are called the canonical coordinates or Darboux coordinates. 
    Furthermore, we denote the non-commutative scale as $\varepsilon\coloneq L_{\text{NC}}^2\coloneq[\det((\theta^\top\theta)^{1/2})]^{1/2d}=(\prod_{i=1}^d\theta_i)^{1/d}$.
\end{definition}

This Moyal-Weyl quantum plane is a quantized symplectic flat space $\mathcal{M}$.
The non-commutative structure emerges from the Weyl quantization of symplectic geometry naturally \cite{gracia1995geometric,Steinacker:2024unq}.

\begin{proposition}[Canonical Commutation Relation \cite{born1925quantenmechanik}]
\label{prp:creation_annihilation}
    Consider the canonical coordinates in a $2d$-dimensional Moyal-Weyl quantum plane.
    For each block, we have the commutation relation
    \begin{align*}
        \big[\hat{Q}^i,\hat{P}^j\big]=i\theta_i\delta^{ij}\hat{\one},\qquad\big[\hat{Q}^i,\hat{Q}^j\big]=\big[\hat{P}^i,\hat{P}^j\big]=0.
    \end{align*}
    We call these relations the canonical commutation relations.
    These relations allow us to construct lowering and raising operators as
    \begin{align}
    \label{eq:annihilation_creation}
        \hat{a}_i\coloneq\frac{1}{\sqrt{2\theta_i}}(\hat{Q}^i+i\hat{P}^i),\qquad \hat{a}^\dagger_i=\frac{1}{\sqrt{2\theta_i}}(\hat{Q}^i-i\hat{P}^i),
    \end{align}
    respectively.
    Specifically, they satisfy the following commutation relation
    \begin{align}
    \label{eq:creation_annihilation_commutation_relation}
        [\hat{a}_i,\hat{a}^\dagger_j]=\delta_{ij}\hat{\one}.
    \end{align}
\end{proposition}

\begin{proof}
    Please see \cref{sec:proof_creation_annihilation} for a detailed proof.
\end{proof}

In \cref{def:quantum_state_map}, we define a general quantum-state map $x\mapsto\ket{x}$.
From now on, we consider Moyal-Weyl coherent state as a concrete example for this quantum state $\ket{x}$.

\begin{definition}[Moyal-Weyl Coherent State \cite{glauber1963coherent}]
\label{def:moyal_weyl_coherent_state}
    We define the normalized coherent state in Moyal-Weyl quantum space $\R^{2d}_\theta$ via
    \begin{align}
        \ket{x}\coloneq \hat{U}_x\ket{0}\coloneq\exp\Big\{i\sum_{a,b=1}^{2d}x^a(\theta^{-1})_{ab}\hat{M}^b\Big\}\ket{0},\qquad \hat{a}_i\ket{0}\coloneq0\;\forall i,
    \end{align}
    where $x^a\in\R$, $\hat{M}^a\in\text{Mat}(\mathcal{H})$, and $\ket{0}$ is a chosen normalized reference coherent state.
    This coherent-space translational operator $\hat{U}_x$ belongs to $\mathcal{B}(\mathcal{H})$ and satisfies $\hat{U}^\dagger_x=\hat{U}_{-x}$. 
\end{definition}


In this paper, we use the term \textit{coherent(-state) space} for the Hilbert space spanned by the chosen coherent-state family $\{\ket{x}\}$.
Hence, a Moyal-Weyl quantum plane possesses non-commutative coordinates $\hat{M}^a$ and acts on the coherent space $\mathcal{H}$.




\begin{proposition}[Inner Product in Moyal-Weyl Quantum Plane \cite{glauber1963coherent,Steinacker:2024unq}]
\label{prp:kernel_moyal_weyl_coherent_inner_product}
    The coherent states follow the inner product structure
    \begin{align}
        \bra{x}\ket{y}=\exp\Big\{-\frac{1}{4}\sum_{a,b=1}^{2d}(x-y)_a(\theta^\top\theta)^{-1/2}_{ab}(x-y)_b\Big\}\exp\Big\{-\frac{i}{2}\sum_{a,b=1}^{2d}x_a(\theta^{-1})_{ab}y_b\Big\}.
    \end{align}
\end{proposition}

\begin{proof}
    Please see \cref{sec:proof_kernel_moyal_weyl_coherent_inner_product} for a detailed proof.
\end{proof}

\begin{remark}
    Notably, $\hat{U}_x$ does not satisfy the commutativity, i.e., $\hat{U}_x\hat{U}_y\neq\hat{U}_y\hat{U}_x$.
    This non-commutativity is obvious from the phase in \cref{prp:kernel_moyal_weyl_coherent_inner_product}.
    Importantly, this refers to the Heisenberg uncertainty principle in quantum theory, i.e., $\Delta X^a\Delta X^b\geq\frac{|\theta^{ab}|}{2}$ or $\Delta Q_i\Delta P_i\geq\frac{\theta_i}{2}$ \cite{uncertainty1929robertson,heisenberg1927anschaulichen}, where $\Delta(\cdot)$ denotes standard deviation under the coherent states.
\end{remark}

\cref{prp:kernel_moyal_weyl_coherent_inner_product} has a complex phase.
Notably, after taking the square, $|\bra{x}\ket{y}|^2$ gives a Gaussian.
This Gaussian comes from the non-commutativity between coordinates $\hat{M}^a$.
To investigate the normalization of this Moyal-Weyl quantum plane, we state the proposition below.

\begin{proposition}[Identity Operator and Trace in Moyal-Weyl Quantum Plane \cite{glauber1963coherent,Steinacker:2024unq}]
\label{prp:normalization_trace}
    Consider a Moyal-Weyl quantum plane $\R^{2d}_\theta$ with coherent states $\ket{x}\in\mathcal{H}$.
    We deal with the normalization consistency by counting the number of coherent-states.
    With the non-commutativity scale $\varepsilon=[\det((\theta^\top\theta)^{1/2})]^{1/2d}=(\prod_{i=1}^d\theta_i)^{1/d}$, this gives the identity operator
    \begin{align}
    \label{eq:identity_operator_moyal_weyl}
        \hat{\one}=\int_{\R^{2d}_\theta} \hat{P}_x~\dd\mu(x)=\int_{\R^{2d}}\ket{x}\bra{x}~\frac{\dd^{2d}x}{(2\pi\varepsilon)^d}.
    \end{align}
    Furthermore, for $\hat{A}=\sum_{r=1}^m\ket{u_r}\bra{v_r}\in\text{Mat}(\mathcal{H})$ with $\ket{u_r},\ket{v_r}\in\mathcal{H}$, the trace satisfies
    \begin{align*}
        \Tr_{\mathcal{H}}(\hat{A})=\int_{\R^{2d}}\bra{x}\hat{A}\ket{x}~\frac{\dd^{2d}x}{(2\pi\varepsilon)^d}.
    \end{align*}
\end{proposition}

\begin{proof}
    Please see \cref{sec:proof_normalization_trace} for a detailed proof.
\end{proof}

\section{A Quantum Duality for Transformers}
\label{sec:3attention}

In this section, we show the duality between quantum theory and multi-head self-attention architecture.
Then we plug this structure into every Transformer block.
In this way, we establish the quantum theory for a full Transformer.

\subsection{Quantum Duality for Attention}

Here we introduce the kernel formulation of attention, including the common softmax attention and the RBF Gaussian kernelized attention.
We connect those models within a single quantum coherent space framework explicitly.

\begin{definition}[Attention Mechanism \cite{vaswani2017attention,tsai2019transformer,katharopoulos2020transformers}]
\label{def:attention_mechanism}
    Let query be $q\in\R^{2d}$, key be $k_\mu\in\R^{2d}$, and value be $v_\mu\in\R^{d_v}$ with $\mu=1,\cdots,D$.
    Then the feature map $\phi:\R^{2d}\to\mathcal{H}_K$ gives the kernel
    \begin{align}
        \kappa(q,k)\coloneq\langle\phi(q),\phi(k)\rangle_{\mathcal{H}_K},
    \end{align}
    and the kernel expression of attention
    \begin{align}
    \label{eq:kernelized_attention}
        {\rm Attn}_\kappa(q,k,v)=\frac{\sum_{\mu=1}^D\kappa(q,k_\mu)v_\mu}{\sum_{\nu=1}^D\kappa(q,k_\nu)}\coloneq\sum_{\mu=1}^D\tilde{\kappa}(q,k_\mu)v_\mu,
    \end{align}
    where $\tilde{\kappa}(q,k_\mu)$ is the (normalized) attention weight.
    For this attention weight, we introduce two kinds of attention, softmax and RBF attention weight
    \begin{align}
    \label{eq:softmax_rbf}
        \kappa_{\text{Soft}}(q,k_\mu)\coloneq e^{\beta q^\top k_\mu},\qquad\kappa_{\text{RBF}}\coloneq e^{-\frac{1}{2}(q-k_\mu)^\top\Sigma^{-1}(q-k_\mu)}.
    \end{align}
\end{definition}

From this setup, the following theorem reveals a connection between attention mechanism and quantum theory framework.

\begin{theorem}[Quantum Description of RBF and Softmax]
\label{thm:quantum_softmax_rbf}
    Consider the projection map $\hat{P}_x=\ket{x}\bra{x}$.
    If we define the scaled projection map as $\hat{\mathcal{P}}_x\coloneq\exp(\|x\|_g^2)\hat{P}_x$ with $\|x\|_g^2\coloneq\frac{1}{2}x^\top(\theta^\top\theta)^{-1/2}x$, 
   then we recover the softmax and RBF kernel by
    \begin{align}
    \label{eq:softmax_rbf_kernel}
        \kappa_{\text{Soft}}(q,k_\mu)=\Tr_{\mathcal{H}}\Big(\hat{\mathcal{P}}_q\hat{\mathcal{P}}_{k_\mu}\Big),\quad\kappa_{\text{RBF}}=\Tr_{\mathcal{H}}\Big(\hat{P}_q\hat{P}_{k_\mu}\Big).
    \end{align}
    In particular, the inverse-temperature and covariance matrix correspond to
    \begin{align*}
        \beta\one_{2d}=(\theta^\top\theta)^{-1/2},\quad\Sigma=(\theta^\top\theta)^{1/2}.
    \end{align*}
\end{theorem}

\begin{proof}
    Please see \cref{sec:proof_quantum_softmax_rbf} for a detailed proof.
\end{proof}



\begin{remark}
    Although the weighted projection map $\mathcal{P}_x$ is different from $P_x$, they live in the same Hilbert space $\mathcal{H}$. 
    Hence, this reweighting only changes the feature representation of the attention mechanism.
    The Hilbert space and the commutation relation on a Moyal-Weyl quantum plane \eqref{eq:moyal_weyl_commutation} remain unchanged.
\end{remark}




\begin{remark}
    In an isotropic case, $\Sigma=\sigma^2\one$ and  $\theta_1=\cdots=\theta_d\coloneq\varepsilon$.
    Then the variance of the Gaussian identifies with the non-commutative scale, i.e., $\sigma^2=\varepsilon$.
    For softmax, it is often isotropic.
    But in the sense of \cref{thm:quantum_softmax_rbf}, one can generalize it to $\beta\in\R^{2d\times 2d}$. 
\end{remark}

To further verify the validity of this Moyal-Weyl realization to softmax and RBF kernels, we should discuss the existence of this quantum duality.
We show the statement below.

\begin{proposition}[Existence of Moyal-Weyl Realization]
\label{prp:existence}
    Consider a RBF Gaussian kernel $\kappa_{\text{RBF}}(x,y)=\exp[-\frac{1}{2}(x-y)^\top\Sigma^{-1}(x-y)]$ with covariance matrix $\Sigma\succ0$.
    Then a Moyal-Weyl duality realization with a linear quantum-state map $x\mapsto\ket{Lx}_\theta$ always exists, where
    \begin{align*}
        L^\top(\theta^\top\theta)^{-1/2}L=\Sigma^{-1},
    \end{align*}
    and $\theta$ is even-dimensional, anti-symmetric and invertible matrix.
    If one requires $L=\one$ in the same dimension, then such a realization exists if and only if every eigenspace of $\Sigma$ is even-dimensional. 
    This existence holds for the softmax kernel $\kappa_{\text{soft}}=\exp(\beta x^\top y)$ as well.
\end{proposition}

\begin{proof}
    Please see \cref{sec:proof_existence} for a detailed proof.
\end{proof}

Aside from the existence of Moyal-Weyl realization of a softmax or RBF kernel, we further verify the non-uniqueness of this duality as follows.
That is, different $\theta$ is able to give the same kernel.

\begin{proposition}[Non-Uniqueness of Moyal-Weyl Realization]
\label{prp:non_uniqueness}
    Whenever the realization exists, it is not unique even if $L=\one$.
    That is, if the multiplicity of an eigenvalue $\lambda$ of $\Sigma=(\theta^\top\theta)^{1/2}$ is $2m_\lambda$, then we parametrize the non-commutative structure by
    \begin{align*}
        \Theta(\Sigma)\simeq\prod_\lambda O(2m_\lambda)/U(m_\lambda),\quad\Theta(\Sigma)\coloneq\{\theta:(\theta^\top\theta)^{1/2}=\Sigma\},
    \end{align*}
    where $O(2m_\lambda)$ is a set of orthogonal matrices, and $U(m_\lambda)$ is a set of unitary matrices.
    All these structures induce the same RBF and softmax kernel, but produce distinct phases in general. 
\end{proposition}

\begin{proof}
    Please see \cref{sec:proof_non_uniqueness} for a detailed proof.
\end{proof}



Together, we arrive the following corollary for this quantum duality via Moyal-Weyl realization.

\begin{corollary}[Existed and Non-Unique Quantum Duality]
\label{coro:exist_uniqueness}
    Every RBF or softmax kernel admits a Moyal-Weyl coherent-state realization.
    The kernel determines $L^\top(\theta^\top\theta)^{-1/2}L$ uniquely, but not the underlying non-commutative structure $\theta$.
    Different realizations carry distinct coherent-state phases in general.
    Hence, the duality is exact for kernels, and the underlying quantum geometry has additional Moyal-Weyl realization degree of freedom.
\end{corollary}

Hence, we establish the duality between (i) a machine learning RKHS perspective and (ii) a Moyal-Weyl quantum plane viewpoint with additional, non-unique underlying geometry.
In this way, we view this kernel formulation as the Transformer attention mechanism in quantum theory.


\subsection{Quantum Duality for Transformer}
\label{sec:moyal_weyl_transformer}

\begin{figure}[t]
\centering
\resizebox{\linewidth}{!}{
\begin{tikzpicture}[
    >=Latex, font=\fontsize{10}{12}\selectfont, line width=.65pt,
    block/.style={draw=black!55,fill=black!3,rounded corners=2pt,
        minimum height=.78cm,align=center,inner sep=5pt},
    quantum/.style={block,draw=RoyalBlue!75,fill=RoyalBlue!5},
    add/.style={draw=black!65,circle,fill=white,minimum size=.48cm,inner sep=0pt},
    flow/.style={->,draw=black!75},
    qflow/.style={->,draw=RoyalBlue!85},
    smallnote/.style={font=\fontsize{9}{10.5}\selectfont,align=center}
]
    \node (h) at (0,0) {$h_i$};
    \node[block] (ln1) at (1.65,0) {${\rm LN}_1$};
    \node[block] (proj) at (3.55,0) {$W_{Q/K/V}$};
    \node (qkv) at (5.65,0) {$q_i,\,k_\mu,\,v_\mu$};
    \node[quantum] (states) at (8.00,0) {$\ket{q_i},\,\ket{k_\mu}$};
    \node[quantum,minimum width=3.15cm] (kernel) at (11.05,0)
        {$\kappa_{\text{RBF}}=\Tr_{\mathcal H}(P_{q_i}P_{k_\mu})$\\[2pt]
         ${\kappa}_{\text{Soft}}=\Tr_{\mathcal H}(\mathcal P_{q_i}\mathcal P_{k_\mu})$};
    \node[block] (norm) at (14.50,0) {normalize};

    \begin{scope}[on background layer]
    \node[
        rectangle,
        rounded corners=4pt,
        draw=RoyalBlue!75,
        dashed,
        line width=.9pt,
        fill=RoyalBlue!2,
        minimum width=6.7cm,
        minimum height=2.1cm
    ] at (9.9,0.2) {};
    \end{scope}
    \node[text=RoyalBlue!85!black,font=\fontsize{11}{13}\selectfont\bfseries] at (9.9,1.55)
        {Moyal-Weyl Quantum Plane};
    \node[text=RoyalBlue!85!black,smallnote] at (9.9,.93)
        {$[\hat{M}^a,\hat{M}^b]=i\theta^{ab}\hat\one$};
    \draw[flow] (h) -- (ln1);
    \draw[flow] (ln1) -- (proj);
    \draw[flow] (proj) -- (qkv);
    \draw[qflow] (qkv) -- (states);
    \draw[qflow] (states) -- (kernel);
    \draw[flow] (kernel) -- (norm);

    \node[block] (weighted) at (14.50,-2.45)
        {$\displaystyle\sum_{\mu=1}^D\tilde\kappa\,v_\mu$};
    \node[block] (concat) at (11.05,-2.45) {Concat $+\,W_O$};
    \node[add] (add1) at (8.5,-2.45) {$+$};
    \node[block] (ln2) at (6.35,-2.45) {${\rm LN}_2$};
    \node[block] (ffn) at (4.6,-2.45) {FFN};
    \node[add] (add2) at (2.7,-2.45) {$+$};
    \node (out) at (.65,-2.45) {$h_i^{\text{new}}$};

    \draw[flow] (norm) -- node[right,smallnote] {$\tilde\kappa$} (weighted);
    \draw[flow] (qkv.south) -- (5.65,-1.42) -- (14.05,-1.42)
        -- ($(weighted.north)+(-.45,0)$);
    \node[smallnote,fill=white,inner sep=2pt] at (9.8,-1.42) {$v_\mu$};
    \draw[flow] (weighted) -- (concat);
    \draw[flow] (concat) -- node[above,smallnote] {$a_i$} (add1);
    \draw[flow] (add1) -- node[above,smallnote] {$h'_i$} (ln2);
    \draw[flow] (ln2) -- (ffn);
    \draw[flow] (ffn) -- (add2);
    \draw[flow] (add2) -- (out);
    \node (res1) at (8.5,-3.45) {$h_i$};
    \node (res2) at (2.7,-3.45) {$h'_i$};
    \draw[flow] (res1) -- (add1);
    \draw[flow] (res2) -- (add2);
\end{tikzpicture}
}
\caption{\textbf{Moyal-Weyl Quantum Plane in a Transformer Layer.}
\small{
This illustration sketches how a hidden state $h_{i,\ell,t}$ passes through one Transformer layer and becomes $h_{i,\ell+1,t}$.
Furthermore, the blue block shows how a Moyal-Weyl quantum plane fit into the Transformer architecture.
}}
\end{figure}
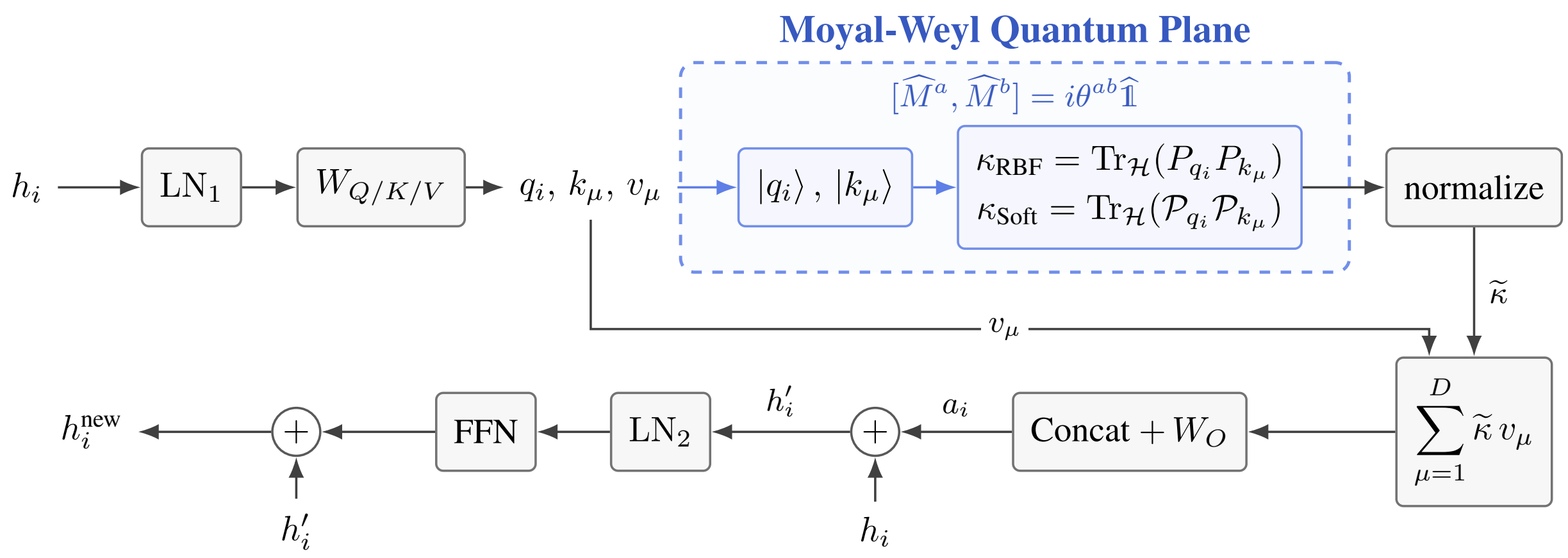

We adopt Moyal-Weyl quantum plane to discuss applications on kernel formulation of attention in Transformers.
Specifically, we define our Transformer architecture \cite{vaswani2017attention} and embed our quantum formulation into it.
For Transformer training, please see \cref{sec:transformer_training} for details.

Let $i$ be the query token index, $\mu$ be the key token index with $i,\mu=[S]$.
Consider an input token sequence $X=(x_1,\cdots,x_S)$, where $X\in\R^{S\times d_{\text{in}}}$ and $d_{\text{in}}$ is the input dimension.
Then we define the embedding $\text{Emb}_t:\R^{d_{\text{in}}}\to\R^{d_{\text{model}}}$ and represent the resulting hidden state as
\begin{align*}
    H_{0,t}(X)=\begin{pmatrix}h_{1,0,t}(X)\\h_{2,0,t}(X)\\\vdots\\h_{S,0,t}(X)\end{pmatrix}\coloneq\begin{pmatrix}\text{Emb}_t(x_1)\\\text{Emb}_t(x_2)\\\vdots\\\text{Emb}_t(x_S)\end{pmatrix}\in\R^{S\times d_{\text{model}}},
\end{align*}
where $t$ labels the iteration, and $\ell=0$ denotes the input layer of a Transformer.
In the following, we construct how hidden states iterate after one Transformer layer \eqref{eq:transformer_update}, i.e., $H_{\ell+1,t}(X)=\mathcal{F}(H_{\ell,t}(X))$.
First, for a full Transformer with multi-head self-attention architecture, we define the $n$-th head query, key, and value as
\begin{align}
    &q_{i,\ell,t}^{(n)}\coloneq W^{(n,\ell,t)}_Q {\rm LN}_1^{(\ell,t)}\big(h_{i,\ell,t}\big)\in\R^{d_h},\\
    &k^{(n)}_{\mu,\ell,t}\coloneq W_K^{(n,\ell,t)}{\rm LN}_1^{(\ell,t)}\big(h_{\mu,\ell,t}\big)\in\R^{d_h},\\
    &v^{(n)}_{\mu,\ell,t}\coloneq W_V^{(n,\ell,t)}{\rm LN}_1^{(\ell,t)}\big(h_{\mu,\ell,t}\big)\in\R^{d_v},
\end{align}
where $d_h$ is the head dimension, $n=[H]$ labels the head, $W_{Q/K}^{(n,\ell,t)}\in\R^{d_h\times d_{{\rm model}}}$ are the query/key projections, $W_V^{(n,\ell,t)}\in\R^{d_v\times d_{{\rm model}}}$ is the value projection, and ${\rm LN}$ is the LayerNorm, i.e.,
\begin{align*}
    {\rm LN}_1^{(\ell,t)}(h^r_{i,\ell,t})\coloneq\gamma^r_{1,\ell,t}\frac{h_{i,\ell,t}^r-\bar{h}_{i,\ell,t}}{\sqrt{s_{i,\ell,t}^2+\epsilon_{{\rm LN}}}}+\beta^r_{1,\ell,t},
\end{align*}
with fixed $\epsilon_{{\rm LN}}>0$.
Here $\bar{h}$ and $s^2_{i,\ell,t}$ act as mean and variance of $h$, i.e.,
\begin{align*}
    \bar{h}_{i,\ell,t}\coloneq\frac{1}{d_{{\rm model}}}\sum_{r=1}^{d_{{\rm model}}}h_{i,\ell,t}^r,\quad s^2_{i,\ell,t}\coloneq\frac{1}{d_{{\rm model}}}\sum_{r=1}^{d_{{\rm model}}}\big(h_{i,\ell,t}^r-\bar{h}_{i,\ell,t}\big)^2.
\end{align*}

In this multi-head self-attention setup, our quantum framework comes in as follows.

\begin{definition}[Moyal-Weyl Quantum Plane in Multi-Head Self-Attention]
\label{def:moyal_weyl_transformer}
    We consider operators $\hat{M}^a\in\text{Mat}(\mathcal{H})$ that forms a Moyal-Weyl quantum plane $\R^{d_h}_\theta$, i.e., 
    \begin{align}
        [\hat{M}^a,\hat{M}^b]=i\theta^{ab}\hat{\one},
    \end{align}
    where $a,b=1,\cdots, d_h$.
    Then the query and key states are coherent states with the following form
    \begin{align}
        &\ket{q^{(n)}_{i,\ell,t}}\coloneq\exp\Big(iq^{(n)\top}_{i,\ell,t}\theta^{-1}\hat{M}\Big)\ket{0}=\exp\Big(iW^{(n,\ell,t)}_Q{\rm LN}_1^{(\ell,t)}\big(h_{i,\ell,t}\big)\theta^{-1}\hat{M}\Big)\ket{0},\\
        &\ket{k^{(n)}_{\mu,\ell,t}}\coloneq\exp\Big(ik^{(n)\top}_{\mu,\ell,t}\theta^{-1}\hat{M}\Big)\ket{0}=\exp\Big(iW^{(n,\ell,t)}_K{\rm LN}_1^{(\ell,t)}\big(h_{\mu,\ell,t}\big)\theta^{-1}\hat{M}\Big)\ket{0}.
    \end{align}
    Then we obtain the $n$-th head softmax or RBF kernelized self-attention weight $\tilde{\kappa}(q^{(n)}_{i,\ell,t},k^{(n)}_{\mu,\ell,t})$ by the inner product \eqref{eq:softmax_rbf_kernel}.
\end{definition}

\cref{def:moyal_weyl_transformer} expresses the attention weights through coherent-state inner product on the Moyal--Weyl quantum plane. 
This attention then gives head outputs and a Transformer performs several non-linear operations to a final output. 
We assemble these operations in the following definition.

By introducing above query, key, value, and attention (\cref{def:attention_mechanism}), we define the output as
\begin{align*}
    a_{i,\ell,t}\coloneq W_{O,\ell,t}{\rm Concat}_{n=1}^H\Big\{\sum_{\mu=1}^D\tilde{\kappa}\big(q^{(n)}_{i,\ell,t},k^{(n)}_{\mu,\ell,t}\big)v^{(n)}_{\mu,\ell,t}\Big\}\in\R^{d_{{\rm model}}},
\end{align*}
which collects total $H$ heads attention and projects to an output $a$ via $W_{O,\ell,t}\in\R^{d_{{\rm model}}\times Hd_v}$.
Then we write $h'$ as the first residual update from this output:
\begin{align*}
    h'_{i,\ell,t}\coloneq h_{i,\ell,t}+a_{i,\ell,t}.
\end{align*}
In this way, it gives the new hidden state after one Transformer layer using one feed-forward network (FFN) and one residual update 
\begin{align}
\label{eq:transformer_update}
    h_{i,\ell+1,t}=h'_{i,\ell,t}+W_{2,\ell,t}\text{ReLU}\Big(W_{1,\ell,t}{\rm LN}_2^{(\ell,t)}\big(h'_{i,\ell,t}\big)+b_{1,\ell,t}\Big)+b_{2,\ell,t},
\end{align}
where $\text{ReLU}(u)\coloneq\max(0,u)$.
Finally, we define the class score and class probability as
\begin{align}
\label{eq:class_score_probability}
    z_{c}(X)=[W_{U,t}h_{i,L,t}(X)]_c,\qquad\pi(c|X)=\frac{e^{z_{c}(X)}}{\sum_{c'\in\mathcal Y}e^{z_{c'}(X)}},
\end{align}
where $W_{U,t}\in\R^{|\mathcal{Y}|\times d_{{\rm model}}}$ gives the class score $z_{c}(X)$ with output class set $\mathcal{Y}$, $\pi(c|X)$ is the probability that assigns input token $X$ to each possible answer $c\in\mathcal{Y}$, and $L$ is the total depth of the Transformer.
Additionally, we leave the Transformer training in \cref{sec:transformer_training}.

\begin{remark}
    If $d_h$ is odd, then we append a zero coordinate to every query and key.
    This constructs an even-dimensional quantum plane in $\R^{d_h+1}$.
    This preserves inner products and hence both softmax and RBF kernels.
\end{remark}

\section{Grokking and Equal-Time-Ordered Correlator (ETOC)}
\label{sec:4grok}

\begin{figure}[t]
    \centering
    \includegraphics[width=1\linewidth]{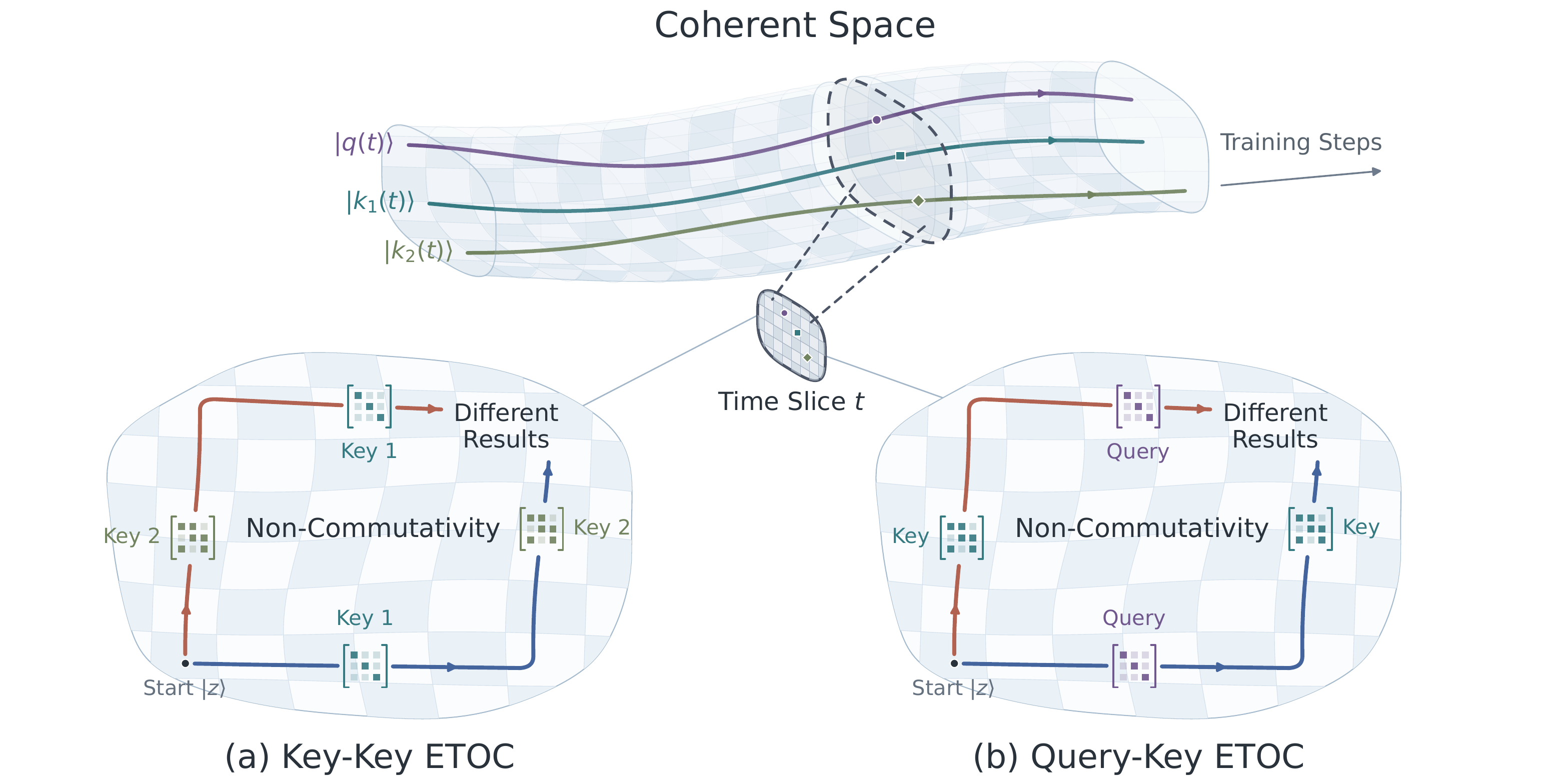}
    \caption{\textbf{Illustration of ETOCs.}
    {\small
    This illustrates the physical meaning of two ETOCs in the underlying coherent space.
    The curves in coherent space represent query and key coherent-state trajectories under Transformer training.
    We evaluate ETOCs by considering a single time slice. 
    \textbf{(a) Key-Key ETOC.} 
    It characterizes the non-commutativity and correlation between key projector operations.
    \textbf{(b) Query-Key ETOC.} 
    It characterizes the non-commutativity and correlation between query and key projector operations.
    }}
    \label{fig:etoc_schematic}
\end{figure}

\begin{figure}[t]
    \centering
    \includegraphics[width=1\linewidth]{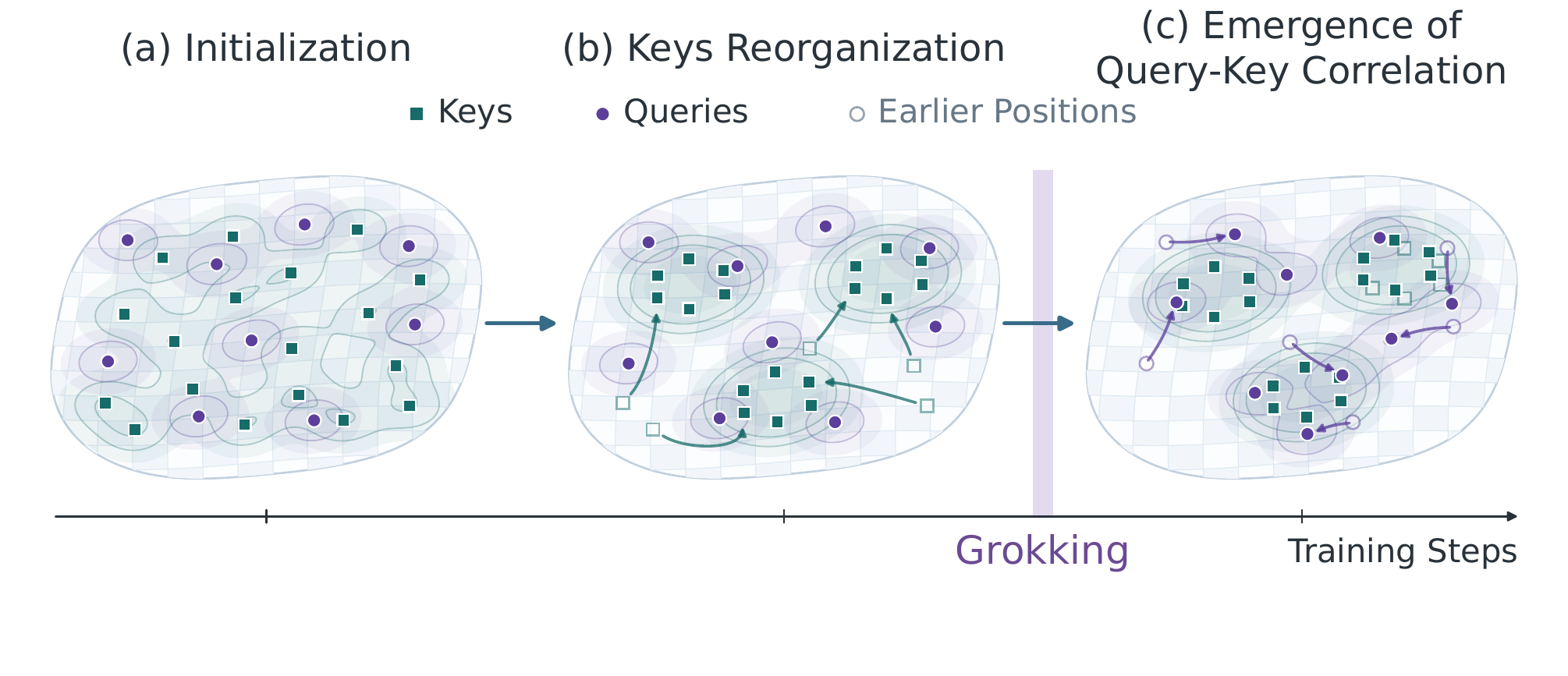}
    \caption{\textbf{Geometric Sketch of Grokking.}
    {\small
    This illustrates \textbf{(a)} the initialization of queries and keys,
    \textbf{(b)} significant key reorganization before Grokking, and
    \textbf{(c)} a later emergence of query-key correlation.
    The purple band marks the Grokking event.
    }}
    \label{fig:etoc_geometry}
\end{figure}

In this section, we adopt the quantum formulation in \cref{sec:3attention} to ask how query and key geometry change in Grokking phenomena \cite{power2022grokking}.
In particular, we introduce equal-time-ordered correlators (ETOCs) to characterize non-commutativity between queries and keys during training.
These measures connect the above question to the experiments in \cref{sec:5exp}.
Please see \cref{fig:etoc_schematic,fig:etoc_geometry} for illustrations of ETOCs and how they connect Grokking, respectively.



\subsection{Grokking Time}

Grokking involves a gap between fitting the training data and generalizing to test data. 
Here we define a common timing rule to measure that gap.

\begin{definition}[Grokking Time]
\label{def:grokking_time}
    We define the Grokking time from test accuracy as
	\begin{align}
		T_g\coloneq\min\Big\{t\mid\operatorname{Acc}_{\mathcal D_{\rm test}}(t)
		\ge1-\delta\Big\},
		\label{eq:grokking-time}
	\end{align}
    where $\text{Acc}_{\mathcal{D}_{\text{test}}}(t)$ denotes the accuracy of the output at update $t$, and $\delta$ is an artificial selected threshold to define Grokking.
\end{definition}

This shows a route from memorization to generalization: the model moves from fitting training examples with poor test performance to learning representations that support unseen examples.

\begin{remark}
    \citet{power2022grokking} introduce Grokking by measuring the first update that accuracy reach $99\%$.
    \citet{levi2024grokking} instead define Grokking time as the difference between the training and generalization accuracy with threshold $95\%$ across times.
    Thus, to clarify, we denote $T_g$ as the former case, rather than to subtract the training crossing time.
\end{remark}

In the following, we establish two quantities from our quantum formulation to characterize this transition.

\subsection{Key-Key Equal-Time-Ordered Correlator (ETOC)}

In this subsection, we define an OTOC-inspired quantity, equal-time-ordered correlator (ETOC),
to characterize the non-commutativity between key projectors at a fixed step $t$.

\begin{definition}[Key-Key ETOC]
\label{def:key_key_etoc}
    Similar to \cref{def:query_key_etoc}, we consider ETOC between distinct keys.
    Let $\mathcal{A}_\mu\subset\R^{d_{\text{in}}}$ be a fixed set of tokens at position $\mu$.
    Then for $\alpha,\alpha'\in\mathcal{A}_\mu$ and $\alpha\neq\alpha'$, we define the key-key ETOC as
    \begin{align}
    \label{eq:key_key_etoc}
        C^{KK}_{\mu,n,\ell,t}(\alpha,\alpha') \coloneq & ~
        \Tr_{\mathcal{H}}\Big(\Big[\hat{P}_{k_{\mu,\ell,t}^{(n)}[\alpha]},\hat{P}_{k_{\mu,\ell,t}^{(n)}[\alpha']}\Big]^\dagger\Big[\hat{P}_{k_{\mu,\ell,t}^{(n)}[\alpha]},\hat{P}_{k_{\mu,\ell,t}^{(n)}[\alpha']}\Big]\Big)\\
        = & ~
        2\kappa_{\mu,n,\ell,t}(\alpha,\alpha')\big[1-\kappa_{\mu,n,\ell,t}(\alpha,\alpha')\big],\annot{By \cref{prp:kernel_moyal_weyl_coherent_inner_product}}
    \end{align}
    where the coherent state inner product (\cref{prp:kernel_moyal_weyl_coherent_inner_product}) gives
    \begin{align*}
        \kappa_{\mu,n,\ell,t}(\alpha,\alpha')\coloneq\Tr_{\mathcal{H}}\Big(\hat{P}_{k_{\mu,\ell,t}^{(n)}[\alpha]}\hat{P}_{k_{\mu,\ell,t}^{(n)}[\alpha']}\Big)=\exp\Big(-\frac{\big\|k_{\mu,\ell,t}^{(n)}[\alpha]-k_{\mu,\ell,t}^{(n)}[\alpha']\big\|^2_2}{2\theta}\Big).
    \end{align*}
    For a fixed quantile level $p\in(0,1)$, we define the quantiled key-key ETOC as
    \begin{align}
        \Gamma^{KK}_\ell(t)\coloneq Q_p\Big\{\big(C^{KK}_{\mu,n,\ell,t}\big)_{1\leq\mu\leq S,~1\leq n\leq H,~\alpha,\alpha'\in\mathcal{A}_\mu,\alpha\neq\alpha'}\Big\},
    \end{align}
    where $Q_p$ denotes the 100$p$-th percentile, i.e., it returns a fraction $p$ of the values are at or below it, and a fraction $(1-p)$ are above it.
\end{definition}

The key-key ETOC $\Gamma^{KK}_\ell(t)$ is small if $\kappa$ is near either zero or one.
These cases describe orthogonal states and identical states, respectively.
The maximum occurs at $\kappa=1/2$.

\begin{remark}
    A high quantile $q$ tracks the upper part of the distribution without depending only on its largest value.
    For example, $Q_{90\%}$ is a cutoff with $90\%$ of values below it and $10\%$ above it, subject to ties and interpolation.
    Hence, the level $p$ determines which part of the distribution we follow.
    $p$ is fixed, while the key pairs near the cutoff are able to change during training. 
\end{remark}




\subsection{Query-Key Equal-Time-Ordered Correlator (ETOC)}

In addition to key-key ETOCs, here we define query-key ETOCs to compare query and key projector changes over a time interval $\Delta$ at a fixed step $t$.
In this way, this query-key ETOC then characterizes the correlation between query and key projectors.

\begin{definition}[Query-Key ETOC]
\label{def:query_key_etoc}
    Consider a fixed layer $\ell$ and a set $\mathcal{P}\subset\R^{S\times d_{\text{in}}}$ of $N=|\mathcal{P}|$ input. 
    We define the averaged $n$-th head query and key projectors as
    \begin{align*}
        \hat{P}_{Q,\ell,t}^{(n)}(X)=\frac{1}{S}\sum_{i=1}^{S}\hat{P}_{q_{i,\ell,t}^{(n)}(X)},\qquad\hat{P}_{K,\ell,t}^{(n)}(X)=\frac{1}{S}\sum_{\mu=1}^{S}\hat{P}_{k_{\mu,\ell,t}^{(n)}(X)},
    \end{align*}
    where subscripts $Q$ and $K$ label token-averaged operators.
    Then the query-key ETOC is the average normalized squared correlator
    \begin{align}
    \label{eq:query-key-signal}
        \Gamma^{QK}_\ell(t)\coloneq\frac1{NH}\sum_{X\in\mathcal P}\sum_{n=1}^{H}\frac{\Tr\Big(\Big[\hat{V}^{(n)}_{Q,\ell,t}(X),\hat{V}^{(n)}_{K,\ell,t}(X)\Big]^\dagger\Big[\hat{V}^{(n)}_{Q,\ell,t}(X),\hat{V}^{(n)}_{K,\ell,t}(X)\Big]\Big)}{2\Tr\Big(\big[\hat{V}^{(n)}_{Q,\ell,t}(X)\big]^2\Big)\Tr\Big(\big[\hat{V}^{(n)}_{K,\ell,t}(X)\big]^2\Big)},
    \end{align}
    where $t\geq\Delta$, $\Delta$ is a positive integer interval, and non-zero
    \begin{align*}
        \hat{V}_{Q,\ell,t}^{(n)}(X)\coloneq\frac{\hat{P}_{Q,\ell,t}^{(n)}(X)-\hat{P}_{Q,\ell,t-\Delta}^{(n)}(X)}{\Delta},\quad \hat{V}_{K,\ell,t}^{(n)}(X)\coloneq\frac{\hat{P}_{K,\ell,t}^{(n)}(X)-\hat{P}_{K,\ell,t-\Delta}^{(n)}(X)}{\Delta}.
    \end{align*}
\end{definition}

\begin{remark}
    The averaged projectors consider the average kernel similarity across all query-key pairs.
    This gives one operator for each collection of queries or keys without selecting particular ones. 
    That is,
    \begin{align*}
        \Tr_{\mathcal{H}}\Big(\hat{P}^{(n)}_{Q,\ell,t}(X)\hat{P}^{(n)}_{K,\ell,t}(X)\Big)=\frac{1}{S^2}\sum_{i=1}^S\sum_{\mu=1}^S\kappa_{\text{RBF}}\big(q^{(n)}_{i,\ell,t}(X),k_{\mu,\ell,t}^{(n)}(X)\big).
    \end{align*}
    Hence, the query-key ETOC \eqref{eq:query-key-signal} measures the non-commutativity between the collective changes of averaged query/key projectors. 
    In this way, a rise in query-key ETOC indicates the emergence of stronger ordering dependence between their changes.
    We refer to this geometric relationship as the emergence of query-key correlation.
\end{remark}

The ETOCs above characterize complementary aspects of representation dynamics.
Specifically, $\Gamma_\ell^{KK}(t)$ summarizes the non-commutativity among key projectors, while $\Gamma_\ell^{QK}(t)$ measures the non-commutativity between query and key projector changes.
We summarize these below.


\subsection{Discussion: Non-Commutativity, Correlation and Grokking}

We interpret ETOCs and the emergence of correlation between queries and keys through  similarities.
Then we provide a way to probe geometric structures behind Transformer training.

\paragraph{Intuition of Key-Key ETOC.}
Key-key ETOCs (1) measure the discrepancy of projector orders through the trace, (2) contain inner-product correlation, and (3) indicate geometric information via \cref{def:key_key_etoc}.
In particular, the projector $\hat{P}_k$ extracts the component of an input feature along $\ket{k}$.
For two keys $k,k'$, applying these projectors in opposite orders gives
\begin{align*}
	\hat{P}_k\hat{P}_{k'}
	&=\bra{k}\ket{k'}\ket{k}\bra{k'},
	\qquad
	\hat{P}_{k'}\hat{P}_k
	=\bra{k'}\ket{k}\ket{k'}\bra{k}.
\end{align*}
The first product outputs a component along $\ket{k}$, while the reverse order outputs a component along $\ket{k'}$.
Hence, the correlation between keys in terms of $\kappa=\exp(-\frac{1}{2\theta}\|k-k'\|^2)$ emerge through key-key ETOCs via trace.
For geometry, this formulation provides the angle between their coherent-state directions via
\begin{align*}
    \angle(k,k'):=\cos^{-1}\big|\bra{k}\ket{k'}\big|=\cos^{-1}\Big[\exp\big(-\frac{\|k-k'\|^2}{4\theta}\big)\Big].
\end{align*}
Together, key-key ETOCs utilize non-commutativity between key projectors to characterize their correlations and the corresponding geometric structure.

\paragraph{Intuition of Query-Key ETOC.}
Similarly, query-key ETOCs describe the correlation between queries and keys, and their geometry in coherent space.
The inner product again provides the similarity and angles for queries and keys.
In particular, expanding the products of $\hat{V}^{(n)}_{Q,\ell,t}(X)$ and $\hat{V}^{(n)}_{K,\ell,t}(X)$ from \cref{def:query_key_etoc} yields closed chains such as
\begin{align*}
	\Tr\Big(\hat{P}_q\hat{P}_k\hat{P}_{q'}\hat{P}_{k'}\Big)=\bra{q}\ket{k}\bra{k}\ket{q'}\bra{q'}\ket{k'}\bra{k'}\ket{q}.
\end{align*}
Its squared magnitude equals the product of the four similarities along that chain, and their complex phases also contribute to the trace via \cref{prp:kernel_moyal_weyl_coherent_inner_product}.
Importantly, this phase records the oriented area of the chain and characterizes geometric information on the coherent space.
Hence, similar to key-key ETOCs, query-key ETOCs utilize the non-commutativity between queries and keys to determine geometry and correlations between them.

\paragraph{Implications to Transformer Training.}

During Transformer training, query and key projectors evolve in coherent space, and hence their non-commutativity.
In particular, changes in key-key similarities reshape which tokens offer similar matching targets and which remain distinguishable.
Change in query-key similarities indicate stronger or weaker matches between queries and these attention targets.
Geometrically, stronger matches correspond to closer query and key representations, and weaker matches correspond to greater separation.
From the intuition of ETOCs above, they characterize the correlations between keys and queries.
Hence, tracking both ETOCs provides complementary views of geometric representation learning: the organization of attention targets and their correlations during training.

For Grokking, the relevant question is how these geometric changes relate to the delayed improvement in test accuracy.
To characterize the transition from memorization to generalization at the geometric representation level, we compare the change times of these two ETOCs with the Grokking time $T_g$.
The next subsection defines a criterion for identifying these change times.


\subsection{ETOC Change Times}

To compare ETOC trajectories with Grokking time $T_g$, we define a timing rule for a significant change of ETOC as follows.
Here we summarize the observed trajectory by two constant levels and choose the split that best fits those levels.

\begin{definition}[Two-Mean Squared Error]
\label{def:etoc_two_mean_error}
    Consider layer $\ell$ and denote $s\in\{QK,KK\}$.
    Since our training time is discrete, we write the times $t_{j_0}<\cdots<t_m=T$ with $j_0\leq j\leq m$, where $j_0$ is the first available sample index, and $T$ is the observation endpoint.
    Then we define
    \begin{align*}
        \Gamma_{\ell,j}^s\coloneq\Gamma_\ell^s(t_j).
    \end{align*}
    For a candidate split after sample $k$ with $j_0\leq k<m$, we define two means
    \begin{align}
        \overline{\Gamma}_{\ell,j_0:k}^s\coloneq\frac{1}{k-j_0+1}\sum_{j=j_0}^{k}\Gamma_{\ell,j}^s,\qquad\overline{\Gamma}_{\ell,k+1:m}^s\coloneq\frac{1}{m-k}\sum_{j=k+1}^{m}\Gamma_{\ell,j}^s.
    \label{eq:etoc_two_means}
    \end{align}
    The total squared error is thereby
    \begin{align}
        \mathcal{E}_{\ell,k}^s\coloneq\sum_{j=j_0}^{k}\big(\Gamma_{\ell,j}^s-\overline{\Gamma}_{\ell,j_0:k}^s\big)^2+\sum_{j=k+1}^{m}\big(\Gamma_{\ell,j}^s-\overline{\Gamma}_{\ell,k+1:m}^s\big)^2.
    \label{eq:etoc_split_error}
    \end{align}
\end{definition}

For a fixed split, the mean gives the smallest squared error among all constant fits to each segment. 
Squaring the differences prevents positive and negative deviations from cancelling. 
The sum then measures how well the two constants fit all the observed values, with equal weight for each sample.

If the signal has two nearly constant levels, a split between them can give a small error. 
A split placed too early or too late puts values from both levels in one segment and can increase the error.
This motivates the following definition of ETOC change time.

\begin{definition}[ETOC Change Time]
\label{def:etoc_signal_change_time}
    From \cref{def:etoc_two_mean_error}, we define the splitting point as
    \begin{align}
        k_{\ell\star}^s(T)\coloneq\min\{\argmin_{j_0\leq k< m} \mathcal{E}_{\ell,k}^s\},\qquad \text{with }~s\in\{QK,KK\},
    \label{eq:etoc_minimizing_split}
    \end{align}
    where the minimization is over integer $k$.
    The outer minimum selects the earliest split if several splits have the same error.
    The query-key and key-key ETOC signal change times are, respectively,
    \begin{align}
        T^{QK}(T;\ell)\coloneq t_{k_{\ell\star}^{QK}(T)+1},\qquad T^{KK}(T;\ell)\coloneq t_{k_{\ell\star}^{KK}(T)+1}.
    \label{eq:etoc_signal_change_time}
    \end{align}
    In the following, we omit $\ell$ when the Transformer layer is fixed.
\end{definition}

The definitions above specify the rule for locating a change in both key-key and query-key ETOCs.
Algorithm~\ref{alg:etoc-grokking} summarizes how we apply them to a trained Transformer and compare the resulting times $T^{QK}$, $T^{KK}$ with the Grokking time $T_g$.
If the test-accuracy does not reach $(1-\delta)$ from \cref{def:grokking_time}, $T_g$ is undefined in this case.

\begin{algorithm}[!t]
    \caption{ETOC for Grokking}
    \label{alg:etoc-grokking}
    \begin{algorithmic}[1]
        \Require Transformer parameters $\{\Theta_t\}$; ETOC observation times
        $\mathcal T$ up to endpoint $T$; input sequence
        $\mathcal P$; fixed token sets $\{\mathcal A_\mu\}_{\mu=1}^S$;
        layer $\ell$; interval $\Delta$; quantile $p$; test-accuracy $\{\operatorname{Acc}_{\mathcal D_{\rm test}}(t)\}$; Grokking
        threshold $\delta$
        \Ensure ETOCs $\{\Gamma_\ell^{QK},\Gamma_\ell^{KK}\}$;
        change times $\{T^{QK}(T;\ell),T^{KK}(T;\ell)\}$; Grokking time $T_g$

        \ForAll{$t\in\mathcal T$}
            \State Compute query/key projectors $\{\hat{P}_q,\hat{P}_k\}$ at layer $\ell$ from
            $\Theta_t$ on $\mathcal P$, and those from
            $\Theta_{t-\Delta}$ when needed
            \Comment{\cref{def:quantum_state_map,def:moyal_weyl_transformer}}

            \If{$t\geq\Delta$}
                \State Compute $\Gamma_\ell^{QK}(t)$ from the interval
                changes of the averaged query and key projectors
                \Comment{\cref{def:query_key_etoc}}
            \EndIf

            \State Compute $\Gamma_\ell^{KK}(t)$ from distinct key pairs
            within each position and head, and take their $p$-quantile
            \Comment{\cref{def:key_key_etoc}}
        \EndFor

        \ForAll{$s\in\{QK,KK\}$}
            \State Order the available observations as
            $(\tau_1^s,\Gamma_{\ell,1}^s),\ldots,
            (\tau_{M_s}^s,\Gamma_{\ell,M_s}^s)$ up to $T$

            \If{$M_s\geq 2$}
                \For{$k=1,\ldots,M_s-1$}
                    \State Evaluate $\mathcal E_{\ell,k}^s$ by the two-mean fit in
                    \cref{def:etoc_two_mean_error}
                \EndFor

                \State $k_\star^s\gets\min\arg\min_{1\leq k<M_s}
                \mathcal E_{\ell,k}^s$

                \State $T^s(T;\ell)\gets\tau_{k_\star^s+1}^s$
            \Else
                \State Set $T^s(T;\ell)$ to undefined
            \EndIf
        \EndFor

        \State $T_g\gets\min\{t:
        \operatorname{Acc}_{\mathcal D_{\rm test}}(t)\geq1-\delta\}$
        \Comment{\cref{def:grokking_time}}

        \State Compare $T^{KK}(T;\ell)$, $T_g$, and
        $T^{QK}(T;\ell)$

        \State \Return both ETOCs $\{\Gamma_\ell^{QK},\Gamma_\ell^{KK}\}$ and the three timing quantities $\{T^{QK},T^{KK},T_g\}$
    \end{algorithmic}
\end{algorithm}


\begin{remark}
    $T^s$ summarizes a change in ETOC signal. 
    The definition does not require an abrupt jump, so a smooth curve also have a minimizing split. 
    Since the fitted means use all samples up to $T$, adding later samples is able to move the minimizing split $k_{\ell\star}^s$.
\end{remark}

\begin{figure}[!htb]
    \centering
    \includegraphics[width=1\linewidth]{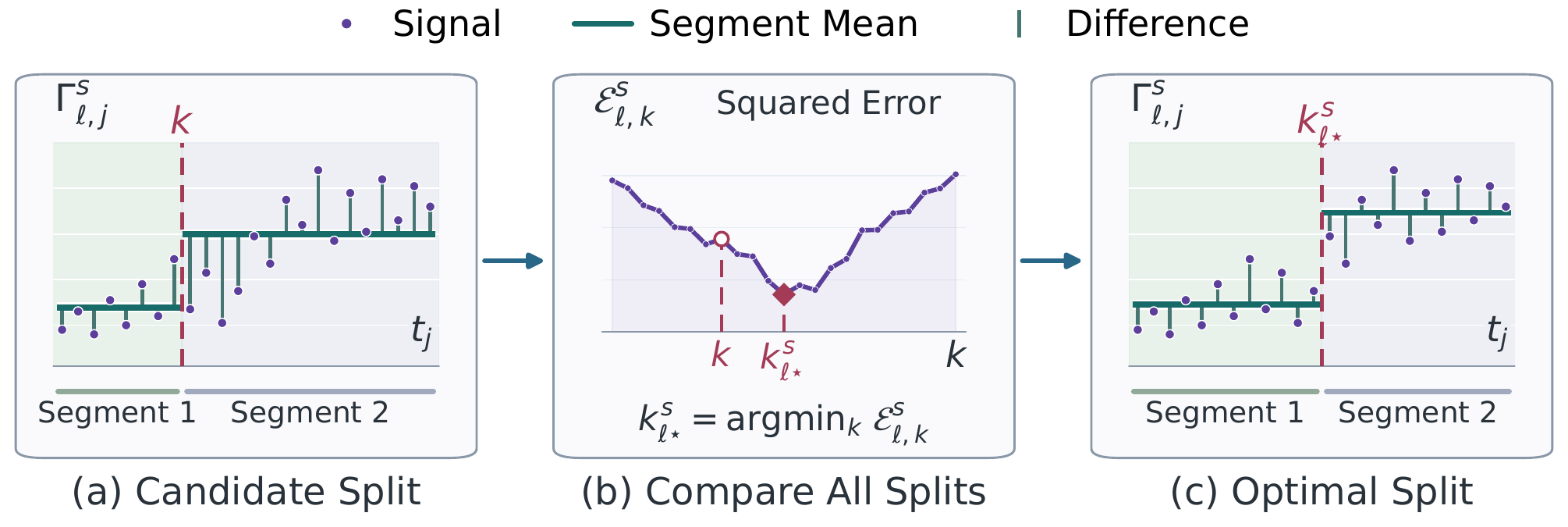}
    \caption{\textbf{Minimization of the Two-Mean Squared Errors.}
    {\small 
    \textbf{(a) Candidate Split.}
	A split at $k$ divides the ETOC signal into two segments.
	Horizontal lines indicate segment means, and vertical segments show differences between samples and their segment means.
	\textbf{(b) Compare All Splits.}
	The curve shows the total squared error $\mathcal{E}_{\ell,k}^{s}$ across candidate splits.
	The open circle marks the candidate in (a), and the filled diamond marks the minimizing split $k_{\ell\star}^{s}$.
	\textbf{(c) Optimal Split.}
	The minimizing split gives the best two-mean fit to the same signal.
	The first sample after this split defines the ETOC change time,
	$T^{s}(T;\ell)=t_{k_{\ell\star}^{s}(T)+1}$.
    }}
    \label{fig:etoc_change_time}
\end{figure}

The ETOC change times connect representation reorganization to the stages of Grokking:
\begin{itemize}
    \item At the early stage, the model achieves high accuracy on training examples but low accuracy on unseen examples.
    This refers to the memorization.
    \item At the later stage, the model achieves high accuracy on both training and unseen examples after Grokking.
    This refers to the generalization.
\end{itemize}
At the representation level, the segment means quantify the strength of ordering effects before and after each split.
For the key-key ETOC, a shift of the strength corresponds to a change in pairwise key-state overlaps and hence in key organization.
For the query-key ETOC, a shift of the strength corresponds to a change in the collective query-key changes.
Hence, the two change times summarize shifts in key organization and in the relations between collective representation changes.
We compare these times with loss functions and Grokking time $T_g$ to characterize the transition from memorization to generalization in \cref{sec:5exp} empirically.

\section{Experimental Studies}
\label{sec:5exp}

We study whether equal-time-ordered correlators (ETOCs) describe changes in query and key geometry around Grokking.
Here Grokking means that test performance improves after a delay relative to training performance \cite{power2022grokking}. 
In this section, our experiments of modular addition and non-zero modular multiplication tasks show that ETOCs change near generalization.
This suggests not only an approach to describe Grokking via quantum theory concretely, but also an online prediction method for locating Grokking.

\subsection{Modular Addition Experiment}
\label{sec:modular_add_exp}

In this subsection, we consider a modular addition task and test the relation between ETOCs and Grokking.
Specifically, we ask whether query, key projectors track a delayed generalization without providing the modular structure in advance.
We use the architecture and setup in 
\cref{sec:3attention}.

\paragraph{Data.}
We consider $a,b\in\{0,\ldots,112\}$, an output set $\mathcal Y=\{0,\ldots,112\}$, and $K=|\mathcal{Y}|=113$ counts these answers.
We denote the dataset as
\begin{align*}
	\mathcal D_+=\{((a,b,{=}),[a+b]_{113})\mid a,b\in\mathbb Z_{113}\},\quad\text{with }~
	\mathbb Z_{113}=\{0,\ldots,112\},
\end{align*}
where $[a+b]_{113}\coloneq (a+b)\mod{113}$.
Take input $(3,5,{=})$ for example, the correct class is $c=8$.
This dataset $\mathcal{D}_+$ contains $113^2=12{,}769$ ordered pairs.
We enumerate pairs in increasing $a$, then increasing $b$.
Python's \texttt{random.Random(0).shuffle} permutes this list.
The first $3{,}830$ examples form $\mathcal D_{\rm train}$, and
the remaining $8{,}939$ form $\mathcal D_{\rm test}$.

\paragraph{Architecture and Training.}
For the input tokens, $d_{\text{in}}=1$, so $X\in\R^S$ is a vector in this case.
For the input embedding, we choose 
\begin{align*}
    h_{A,0,t}(X)\coloneq E_{\text{token};t}[x_{A}]+E_{\text{pos};t}[A]\in\R^{d_{{\rm model}}},\quad\text{with }~A=i,\mu,
\end{align*}
where $E_{\text{tok};t}$ specifies a token value, $E_{\text{pos};t}$ specifies its position, and $t$ labels the iteration. 
Our network contains one Transformer block (\cref{sec:moyal_weyl_transformer}) of width $d_{\rm model}=128$.
Hence, we fix $\ell=0$.
It has $H=4$ attention heads, each with query, key, and value width $d_h=32$, and a FFN width $d_{f}=512$.
Indices $i,\mu\in\{1,\ldots,S\}$ denote token positions with $S=3$.
For the AdamW training process \eqref{eq:adamw}, we set the learning rate $\eta=10^{-3}$, coefficients
$(\beta_1,\beta_2)=(0.9,0.98)$, regularizer $\epsilon_{\text{Adam}}=10^{-8}$,
and weight-decay coefficient $\lambda=1$.
Decay acts on every trainable parameter, including biases and LN parameters.

\paragraph{Observables.}
We measure the cross-entropy loss (CE) \eqref{eq:cross_entropy_loss}, the ETOCs $\Gamma^{QK}_{\ell=0}$ (\cref{def:query_key_etoc}), $\Gamma^{KK}_{\ell=0}$ (\cref{def:key_key_etoc}), the signal change $T^{QK}$, $T^{KK}$ (\cref{def:etoc_signal_change_time}), and the Grokking time $T_g$ (\cref{def:grokking_time}) among eight random seeds.

\begin{figure}[!htb]
	\centering
	\includegraphics[width=1.0\textwidth]{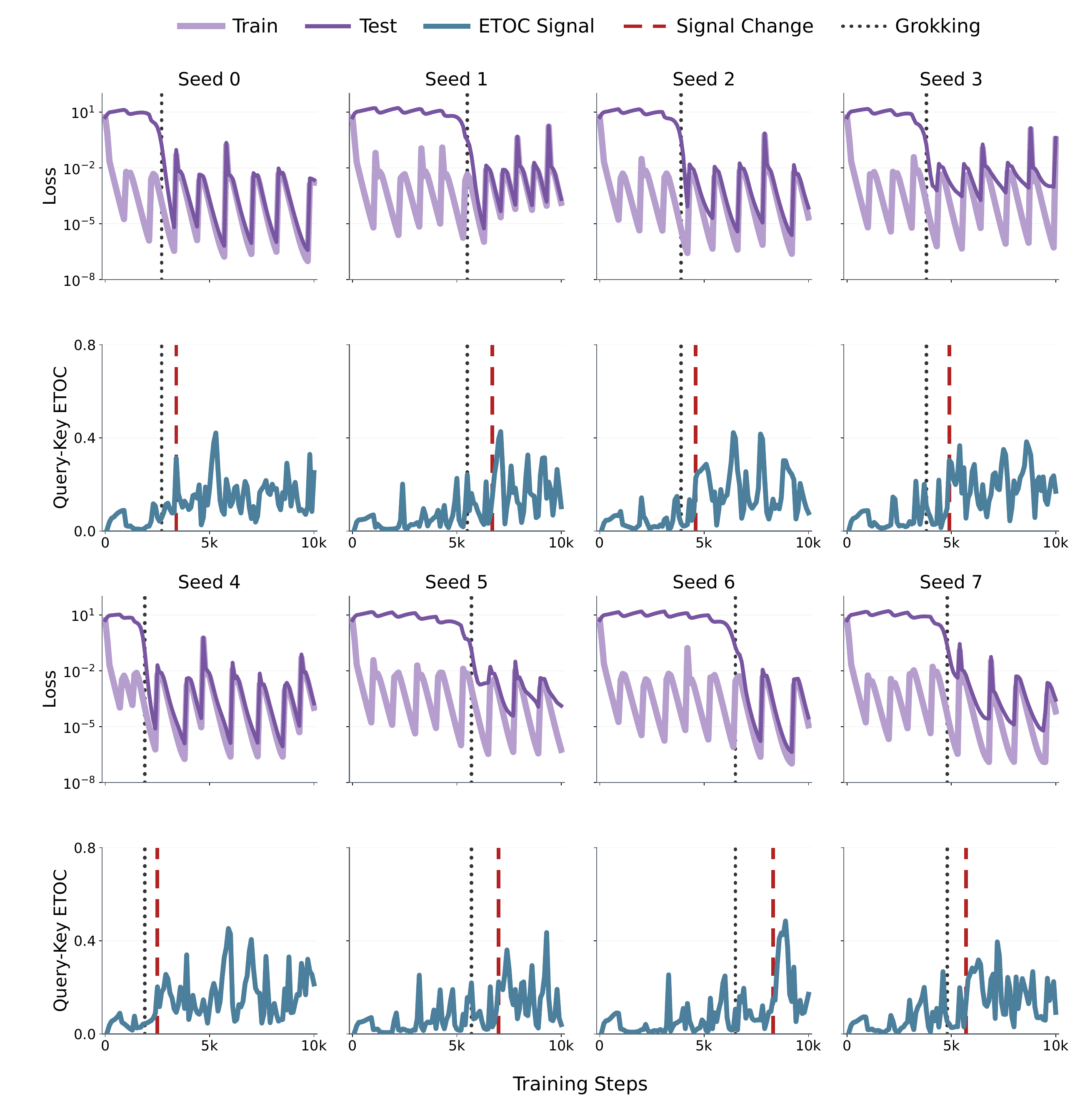}
	\caption{\textbf{Modular Addition: Loss and Query-Key Changes.}
	{\small
	We show the results of seeds $0$--$7$.
	Each run uses four-head Transformer and $10{,}000$ AdamW updates.
	\textbf{Loss rows.} Rows one and three show Train and Test cross entropy
	on the same logarithmic scale.
	\textbf{Query-key ETOC rows.} Rows two and four show the normalized squared
	commutator of query/key projector changes over $100$ updates.
	The query-key ETOC curve averages $64$ inputs and four heads after normalization.
	The red dashed line marks its fitted change time.
    The gray dotted Grokking line marks the point where its test-accuracy exceeds $90\%$. 
	}
	}
	\label{fig:addition}
\end{figure}

\begin{figure}[!htb]
	\centering
	\includegraphics[width=1.0\textwidth]{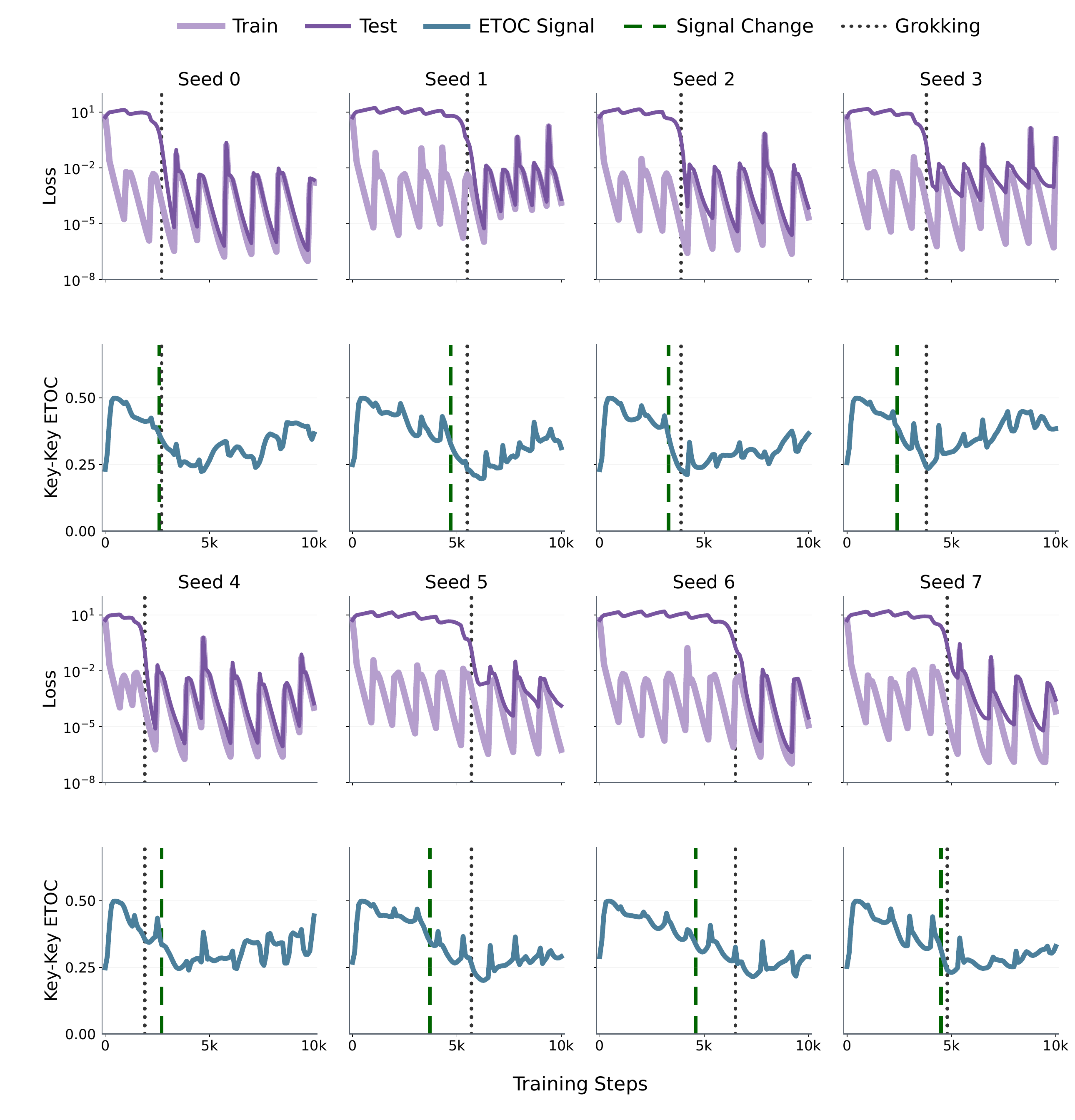}
	\caption{\textbf{Modular Addition: Loss and Key-Key ETOC.}
	{\small
	We show the results of seeds $0$--$7$.
	The runs and loss curves are the same as in \cref{fig:addition}.
	\textbf{Loss rows.} Rows one and three show Train and Test cross entropy on a logarithmic scale.
	\textbf{Key-key ETOC rows.} Rows two and four show the 90th percentile of squared commutators
	between normalized key projectors at the same training update.
	The pool contains $18{,}816$ ordered distinct-token pairs from two operand positions
	and four heads. Each pair stays within one position and head.
	The green dashed line marks its fitted change time.
	The gray dotted Grokking line marks the point where its test-accuracy exceeds $90\%$.
	}}
	\label{fig:addition-key-key}
\end{figure}

\paragraph{Results.}
We plot the result of modular addition experiment in \cref{fig:addition,fig:addition-key-key}.

For all of the seeds here, training CE decreases before
testing CE reaches its late-training scale (generalization).
This validates the Grokking phenomenon in this task.
By combining \cref{fig:addition,fig:addition-key-key}, we observe
\begin{align}
\label{eq:ordering}
	T^{KK}(10{,}000)<T_g<T^{QK}(10{,}000)
\end{align}
in seven of eight seeds.

\begin{remark}
    Seed 4 is the only exception that its Grokking time is earlier than the key-key ETOC change time.
    We notice that its corresponding training loss does not exhibit multiple peaks toward the late-training scale before the Grokking.
    This indicates that the keys organize well at the very beginning, so there is no substantial memorization.
    The testing loss declines with training loss in this case.
    Hence, this leads to an exception for the ordering \eqref{eq:ordering}.
\end{remark}

\begin{figure}[!htb]
	\centering
	\includegraphics[width=1\textwidth]{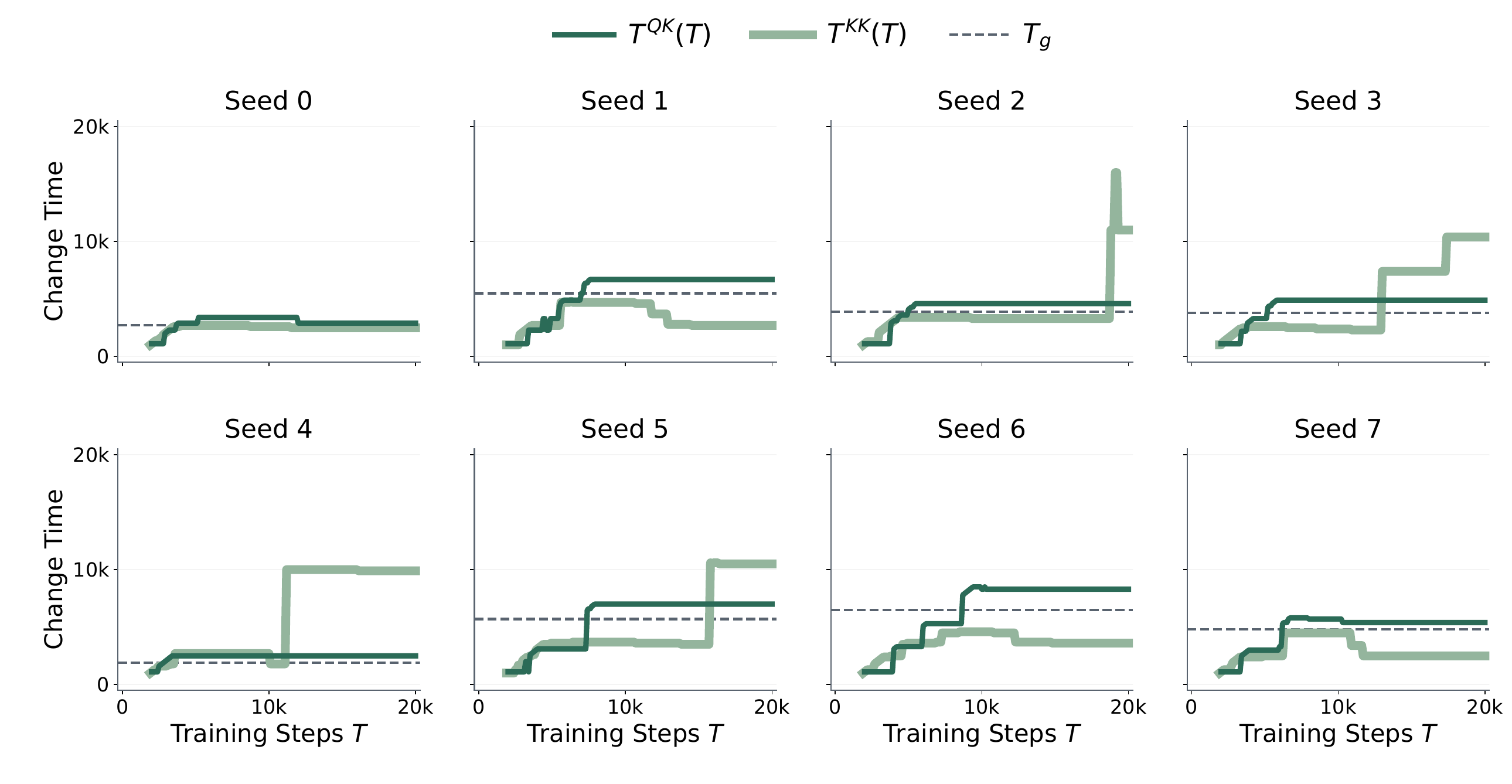}
	\caption{\textbf{Modular Addition: Change Times versus Total Updates.}
	{\small Panels show seeds $0$--$7$ in row order.
	The horizontal axis is the total update $T$.
	The curves show $T^{QK}(T)$ and $T^{KK}(T)$ from all signal records up to $T$.
	The dotted horizontal line is the Grokking time $T_g$. 
    }}
	\label{fig:addition-horizon}
\end{figure}

To test persistence of the ordering, we consider all these eight seeds and refit their locations of $T^{KK}$, $T^{QK}$ for longer total updates $T$.
We show the result in \cref{fig:addition-horizon}.

In \cref{fig:addition-horizon}, early stable windows indicates a reliable region that places $T_g$ between the $T^{KK}$ and $T^{QK}$ locations.
In comparison, the large-update regime shows a loss of ordering.
Specifically, at $T=20{,}000$, only four of eight seeds satisfy $T^{KK}<T_g<T^{QK}$.
Furthermore, the curves in some seeds jump dramatically at high $T$.
This is the consequence of \cref{def:etoc_signal_change_time}.
The definition of ETOC change assigns equal weight to every record, i.e., adding a late segment changes the splitting points.
This makes the evaluated change time at large updates regime unreliable.

From \cref{fig:addition,fig:addition-key-key,fig:addition-horizon}, the first stable windows suggest $T^{KK}<T_g<T^{QK}$.
This indicates that the key geometry reorganizes before generalization, and the query-key structure changes afterward.

\subsection{Modular Multiplication Experiment}
\label{sec:modular_multipli_exp}

In this subsection, we consider a non-zero modular multiplication task and test the relation between ETOCs and Grokking.
Specifically, we ask whether query, key projectors track delayed generalization without providing the modular structure in advance.
We use the architecture and setup in \cref{sec:3attention}.

\paragraph{Data.}
We consider $a,b\in\{1,\ldots,112\}$ without zero, an output set $\mathcal Y=\{1,\ldots,112\}$, and $K=|\mathcal{Y}|=112$ counts these answers.
We denote the dataset as
\begin{align*}
	\mathcal D_\times
	&=\{((a,b,{=}),[ab]_{113})\mid a,b\in\mathbb{F}_{113}\}.
\end{align*}
where $[ab]_{113}\coloneq(ab)\mod{113}$.
Take input $(3,5,{=})$ for example, the correct class is $c=15$.
Since $113$ is prime, the product of two nonzero residues stays nonzero.
This dataset $\mathcal{D}_\times$ contains $112^2=12{,}544$ ordered inputs.
We enumerate input pairs in increasing $a$, then increasing $b$.
NumPy's \texttt{default\_rng(0).permutation} permutes their indices.
The first $3{,}763$ examples form $\mathcal D_{\rm train}$, and
the remaining $8{,}781$ form $\mathcal D_{\rm test}$.

\paragraph{Architecture and Training.}
For the input tokens, $d_{\text{in}}=1$, so $X\in\R^S$ is a vector in this case.
For the input embedding, we choose
\begin{align*}
    h_{A,0,t}(X)\coloneq E_{\text{token};t}[x_{A}]+E_{\text{pos};t}[A]\in\R^{d_{{\rm model}}},\quad\text{with }~A=i,\mu,
\end{align*}
where $E_{\text{tok};t}$ specifies a token value, $E_{\text{pos};t}$ specifies its position, and $t$ labels the iteration. 
Our network contains one Transformer block (\cref{sec:moyal_weyl_transformer}) of width $d_{\rm model}=128$.
Hence, we fix $\ell=0$.
It has $H=4$ attention heads, each with query, key, and value width $d_h=32$, and a FFN width $d_{f}=512$.
Indices $i,\mu\in\{1,\ldots,S\}$ denote token positions with $S=3$.
For the AdamW training process \eqref{eq:adamw}, we set the learning rate $\eta=10^{-3}$, coefficients
$(\beta_1,\beta_2)=(0.9,0.98)$, regularizer $\epsilon_{\text{Adam}}=10^{-8}$,
and weight-decay coefficient $\lambda=1$.
Decay acts on every trainable parameter, including biases and LN parameters.

\paragraph{Observables.}
We measure the cross-entropy loss (CE) \eqref{eq:cross_entropy_loss}, the ETOCs $\Gamma^{QK}_{\ell=0}$ (\cref{def:query_key_etoc}), $\Gamma^{KK}_{\ell=0}$ (\cref{def:key_key_etoc}), the signal change $T^{QK}$, $T^{KK}$ (\cref{def:etoc_signal_change_time}), and the Grokking time $T_g$ (\cref{def:grokking_time}) among eight random seeds.

\begin{figure}[!htb]
	\centering
	\includegraphics[width=1.0\textwidth]{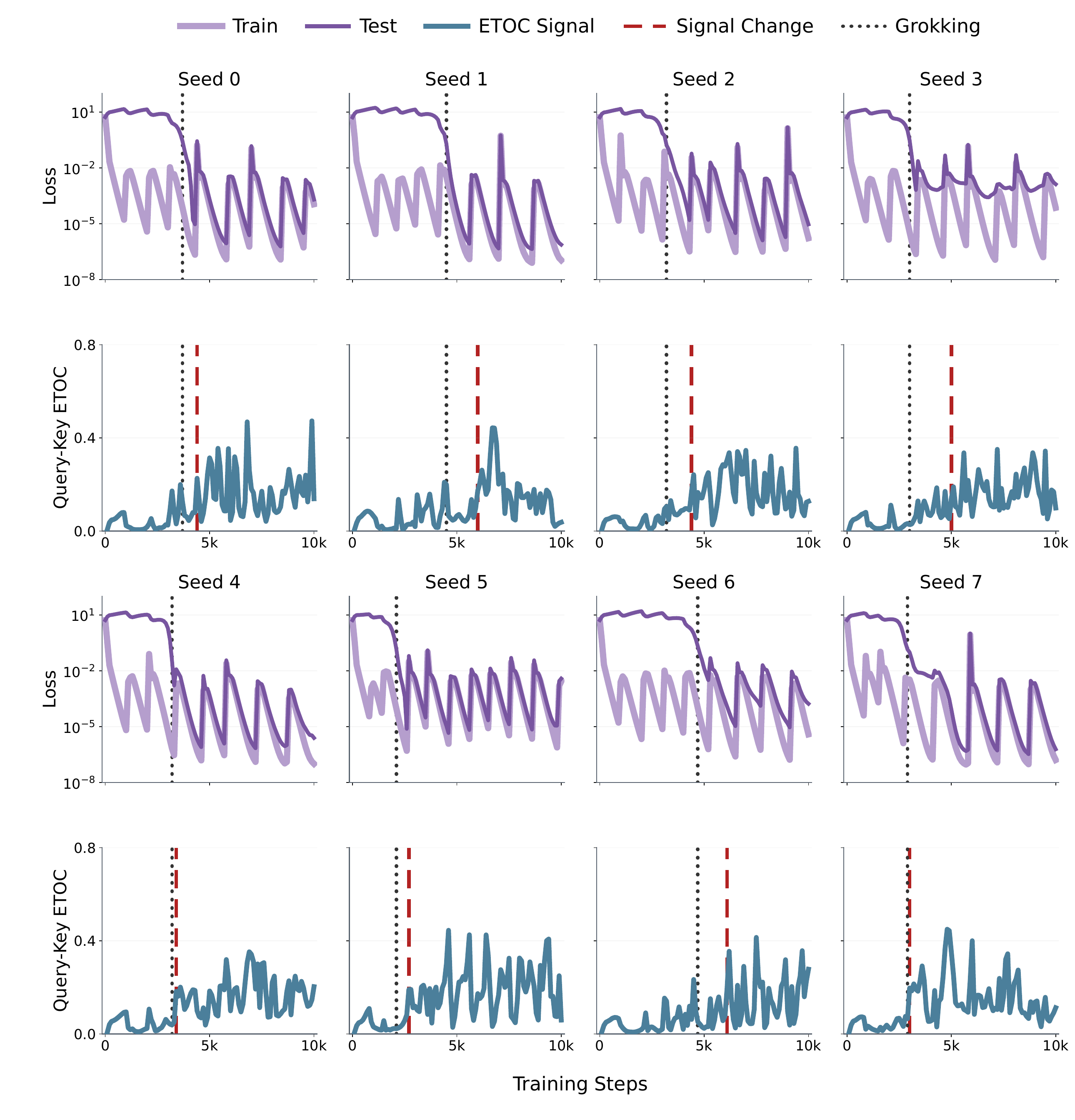}
	\caption{\textbf{Non-Zero Modular Multiplication: Loss and Query/Key Changes.}
	{\small
	We show the results of seeds $0$--$7$.
	Each run uses one four-head Transformer block and $10{,}000$ AdamW updates.
	\textbf{Loss rows.} Rows one and three show Train and Test cross entropy
	on the same logarithmic scale.
	\textbf{Query-key ETOC rows.} Rows two and four show the normalized squared
	commutator of query/key projector changes over $100$ updates.
	The query-key ETOC curve averages $64$ inputs and four heads after normalization.
	The red dashed line marks its fitted change time.
    The gray dotted Grokking line marks the point where its test-accuracy exceeds $90\%$. 
	}
	}
	\label{fig:multiplication}
\end{figure}

\begin{figure}[!htb]
	\centering
	\includegraphics[width=1.0\textwidth]{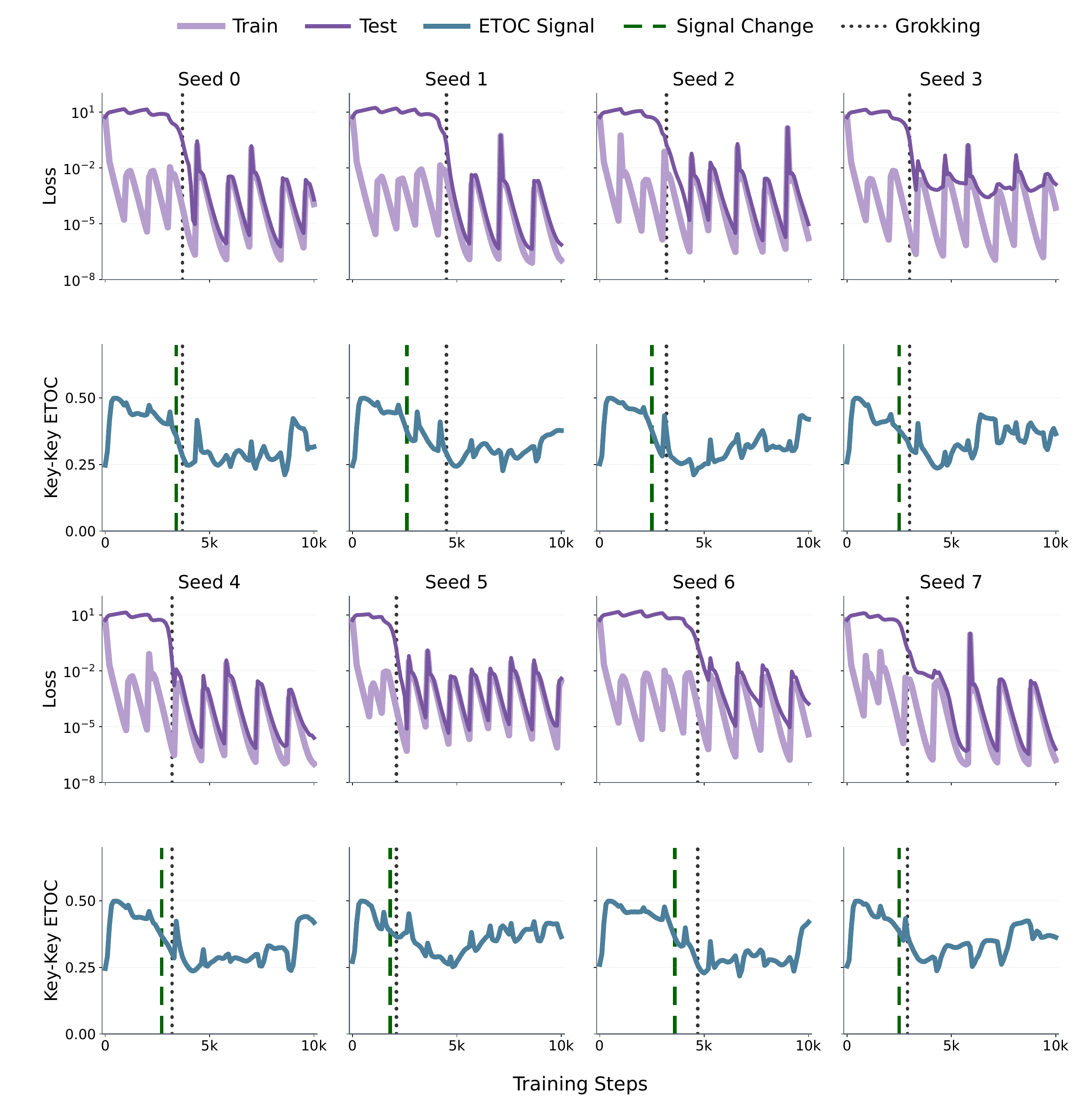}
	\caption{\textbf{Non-Zero Modular Multiplication: Loss and Key-Key ETOC.}
	{\small
	We show the results of seeds $0$--$7$.
	The runs and loss curves are the same as in \cref{fig:multiplication}.
	\textbf{Loss rows.} Rows one and three show Train and Test cross entropy on a logarithmic scale.
	\textbf{Key-key ETOC rows.} Rows two and four show the 90th percentile of squared commutators
	between normalized key projectors at the same training update.
	The pool contains $19{,}632$ ordered distinct-token pairs from two operand positions
	and four heads. 
    Each pair stays within one position and head.
	The green dashed line marks its fitted change time.
	The gray dotted Grokking line marks the point where its test-accuracy exceeds $90\%$.
	}}
	\label{fig:multiplication-key-key}
\end{figure}

\begin{figure}[!htb]
	\centering
	\includegraphics[width=1\textwidth]{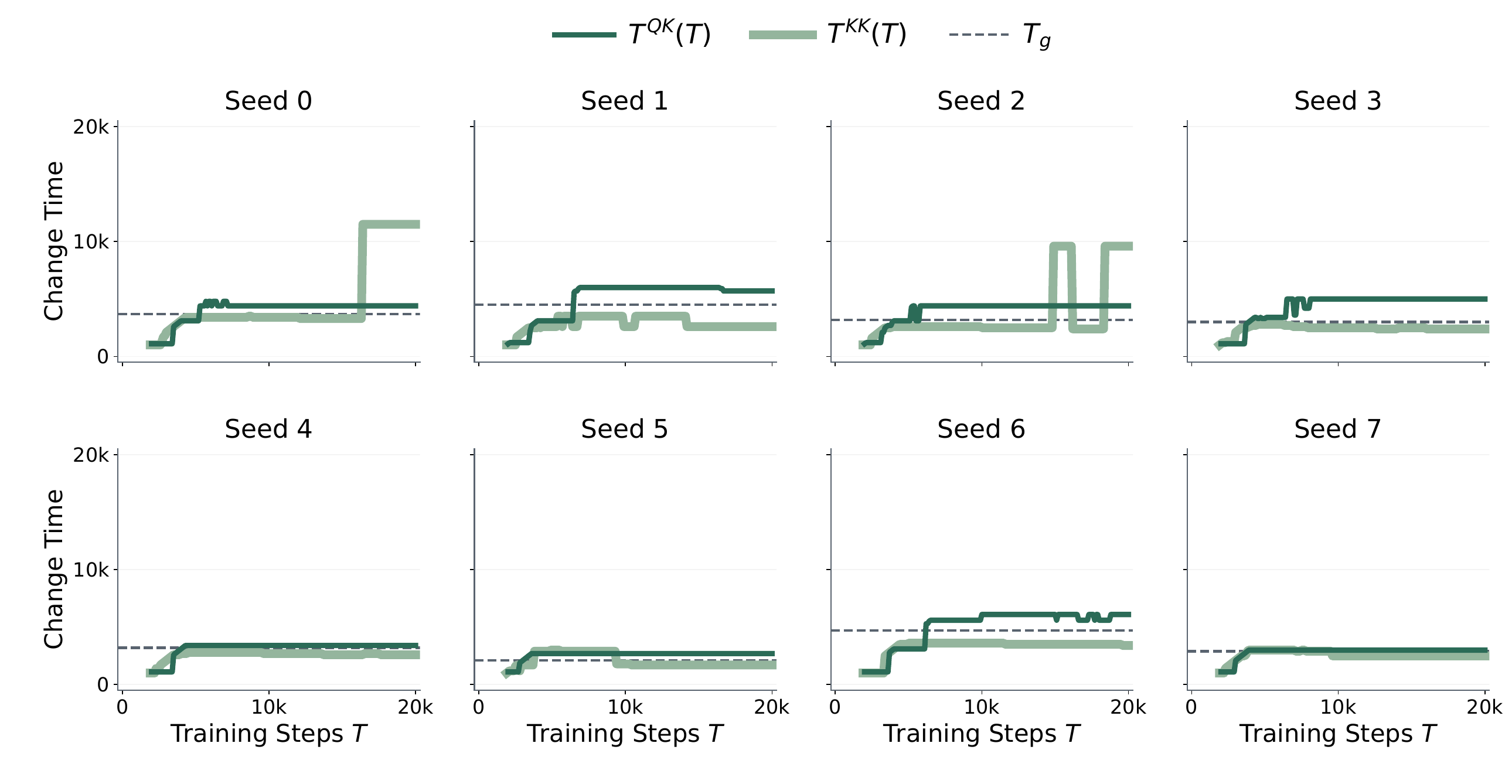}
	\caption{\textbf{Modular Multiplication: Change Times versus Total Updates.}
	{\small Panels show seeds $0$--$7$ in row order.
	The horizontal axis is the total updates $T$.
	The curves show $T^{QK}(T)$ and $T^{KK}(T)$ from all signal records up to $T$.
	The dotted horizontal line is the Grokking time $T_g$.
    }}
	\label{fig:multiplication-horizon}
\end{figure}

\paragraph{Results.}
We plot the result of non-zero modular multiplication experiment as \cref{fig:multiplication,fig:multiplication-key-key,fig:multiplication-horizon}.

For all of the seeds here, training CE decreases before
testing CE reaches its late-training scale (generalization).
This validates the Grokking phenomenon in this task.
By combining \cref{fig:multiplication,fig:multiplication-key-key}, we observe
\begin{align*}
	T^{KK}(10{,}000)<T_g<T^{QK}(10{,}000)
\end{align*}
in all eight seeds.
To test persistence of the ordering, we consider all these eight seeds and refit their locations of $T^{KK}$, $T^{QK}$ for longer total updates $T$.
We show the result in \cref{fig:multiplication-horizon}.

In \cref{fig:multiplication-horizon}, early stable windows indicates a reliable region that places $T_g$ between the $T^{KK}$ and $T^{QK}$ locations.
In comparison, the large-update regime shows a loss of ordering.
Specifically, at $T=20{,}000$, only six of eight seeds satisfy $T^{KK}<T_g<T^{QK}$.
Furthermore, the curves in some seeds jump dramatically at high $T$.
This is the consequence of \cref{def:etoc_signal_change_time}.
The definition of ETOC change assigns equal weight to every record, i.e., adding a late segment changes the splitting points and makes the points unreliable.

From \cref{fig:multiplication,fig:multiplication-key-key,fig:multiplication-horizon}, the first stable windows suggest $T^{KK}<T_g<T^{QK}$.
This indicates that the key geometry reorganizes before test generalization, and the collective query-key structure changes afterward.

Together, the two arithmetic tasks support an association between generalization and changes in projector geometry.


\subsection{Non-Grokking Experiments}
\label{sec:exp_copy}

To answer how does our analysis differ in a non-Grokking task, we choose first-input copying as the example.
That is, for an input $(a,b,=)$, the output is $a$ and is independent of $b$.

\paragraph{Data.}
The input format remains $(a,b,{=})$, but the target is the first operand $a$, i.e.,
\begin{align*}
	\mathcal D_{\text{no-Grok}}
	&=\{((a,b,{=}),a):a,b\in\mathbb Z_{113}\}.
\end{align*}
For example, $(3,5,{=})$ has target $3$, regardless of the second operand.
Hence, the task only requires copying one token, rather than combining two operands.
We choose $3,830$ examples for $\mathcal{D}_{\text{train}}$, and $8,939$ for $\mathcal{D}_{\text{test}}$.
The answer remains in $\mathcal Y=\mathbb Z_{113}$, so $K=|\mathcal{Y}|=113$ stays fixed.

\paragraph{Architecture and Training.}
For the input tokens, $d_{\text{in}}=1$, so $X\in\R^S$ is a vector in this case.
For the input embedding, we choose
\begin{align*}
    h_{A,0,t}(X)\coloneq E_{\text{token};t}[x_{A}]+E_{\text{pos};t}[A]\in\R^{d_{{\rm model}}},\quad\text{with }~A=i,\mu,
\end{align*}
where $E_{\text{tok};t}$ specifies a token value, $E_{\text{pos};t}$ specifies its position, and $t$ labels the iteration. 
Our network contains one Transformer block (\cref{sec:moyal_weyl_transformer}) of width $d_{\rm model}=128$.
Hence, we fix $\ell=0$.
It has $H=4$ attention heads, each with query, key, and value width $d_h=32$, and a FFN width $d_{f}=512$.
Indices $i,\mu\in\{1,\ldots,S\}$ denote token positions with $S=3$.
For the AdamW training process \eqref{eq:adamw}, we set the learning rate $\eta=10^{-3}$, coefficients
$(\beta_1,\beta_2)=(0.9,0.98)$, regularizer $\epsilon_{\text{Adam}}=10^{-8}$,
and weight-decay coefficient $\lambda=1$.
Decay acts on every trainable parameter, including biases and LN parameters.

\paragraph{Observables.}
We measure the cross-entropy loss (CE) \eqref{eq:cross_entropy_loss}, the ETOCs $\Gamma^{QK}_{\ell=0}$ (\cref{def:query_key_etoc}), $\Gamma^{KK}_{\ell=0}$ (\cref{def:key_key_etoc}), the signal change $T^{QK}$, $T^{KK}$ (\cref{def:etoc_signal_change_time}), and the Grokking time $T_g$ (\cref{def:grokking_time}) among eight random seeds.

\begin{figure}[!htb]
	\centering
	\includegraphics[width=1.0\textwidth]{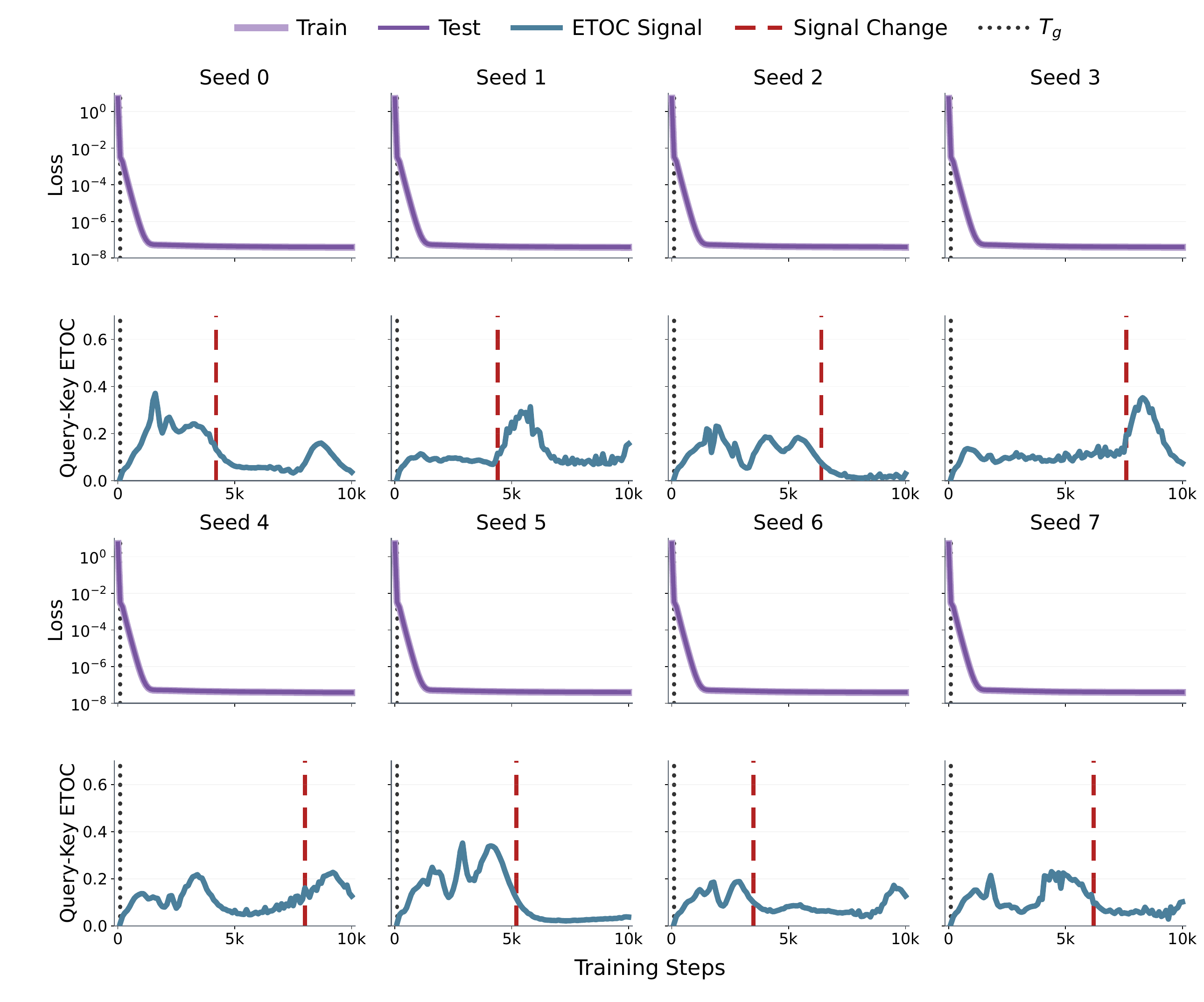}
	\caption{\textbf{Copy Task: Loss and Query-Key ETOC.}
	{\small The input is $(a,b,{=})$, and the target is $a$.
	We show the results of seeds $0$--$7$.
	Each run uses one four-head Transformer block and the first $10{,}000$ AdamW updates.
	\textbf{Loss rows.} Rows one and three show Train and Test cross entropy on logarithmic axes.
	\textbf{Query-key ETOC rows.} Rows two and four show $\Gamma^{QK}_{\ell=0}(t)$ for projector changes over $100$ updates, with a mean over $64$ inputs and four heads.
	The red dashed line marks the fitted change time $T^{QK}$.
	The gray dotted Grokking line marks the point where its test-accuracy exceeds $90\%$. 
    Hence, this result indicates the absence of Grokking.}}
	\label{fig:copy-qk}
\end{figure}

\begin{figure}[!htb]
	\centering
	\includegraphics[width=1.0\textwidth]{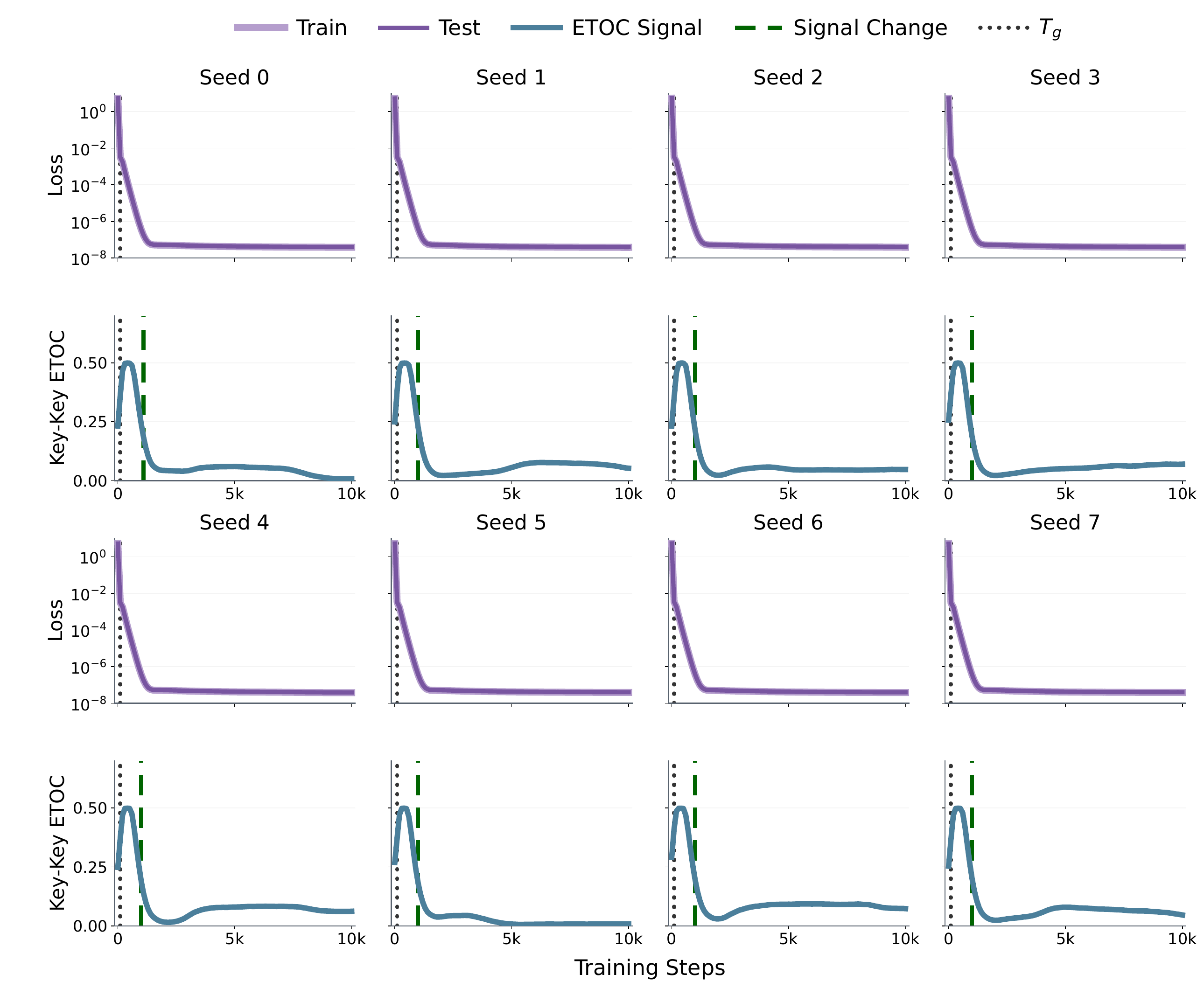}
	\caption{\textbf{Copy Task: Loss and Key-Key ETOC.}
	{\small The input is $(a,b,{=})$, and the target is $a$.
    We show the results of seeds $0$--$7$.
	Each run uses one four-head Transformer block.
	\textbf{Loss rows.} Rows one and three show Train and Test cross entropy on logarithmic axes through $10{,}000$ updates.
	\textbf{Key-key ETOC rows.} Rows two and four show $\Gamma^{KK}_{\ell=0}(t)$, the 90th percentile of $18{,}816$ ordered distinct-key commutators.
	The green dashed line marks the fitted change time $T^{KK}$.
	The gray dotted line marks the point where its test-accuracy exceeds $90\%$. 
    Hence, this result indicates the absence of Grokking.}}
	\label{fig:copy-kk}
\end{figure}

\begin{figure}[!htb]
	\centering
	\includegraphics[width=1\textwidth]{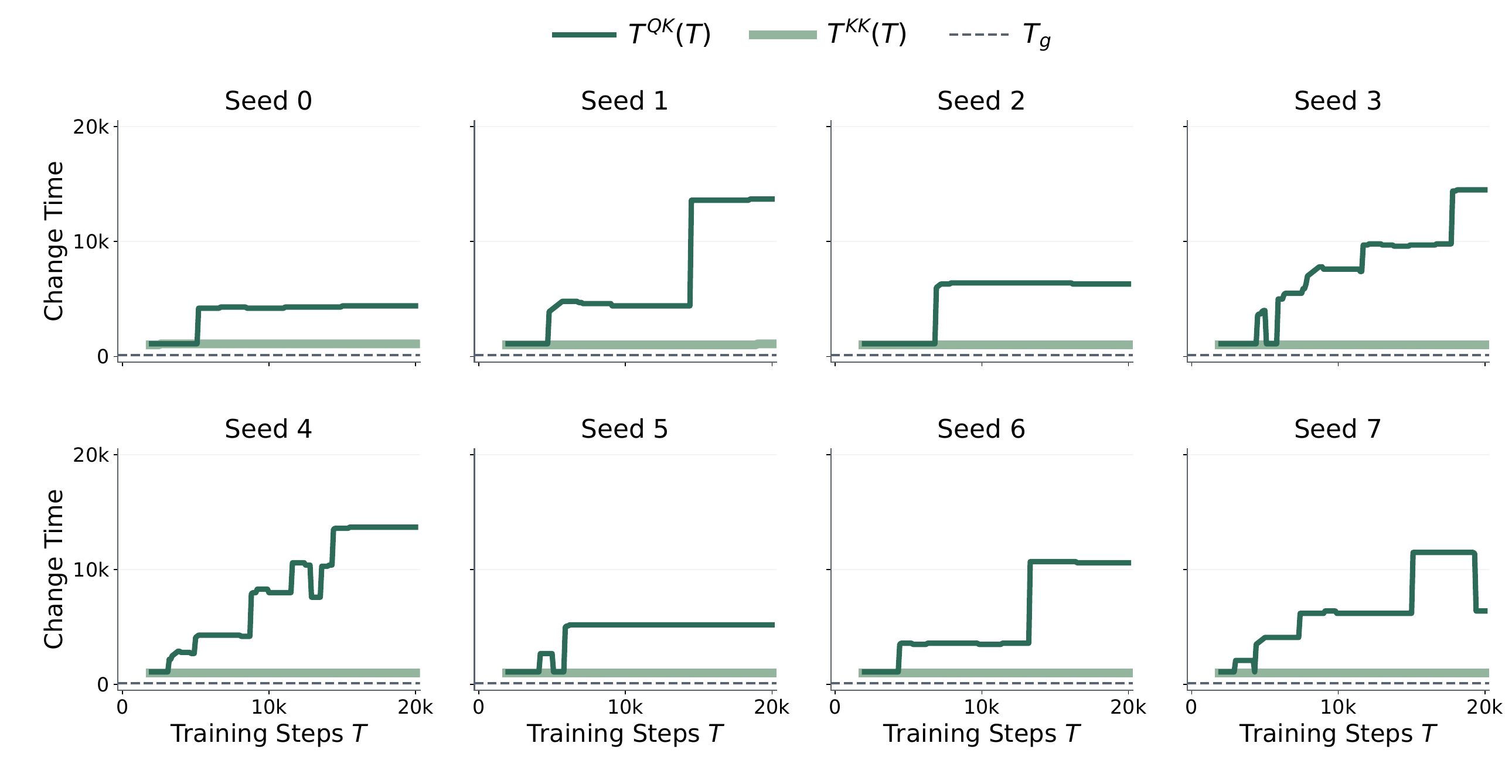}
	\caption{\textbf{Copy Task: Change Times versus Total Updates.}
	{\small Panels show seeds $0$--$7$.
	The horizontal axis is the total updates $T$. 
	The curves show $T^{QK}(T)$ and $T^{KK}(T)$ from all signal records up to $T$.
    The dotted horizontal line is the Grokking time $T_g$.
    }}
	\label{fig:copy-horizon}
\end{figure}

\paragraph{Results.}
We plot the result of non-Grokking copy experiment as \cref{fig:copy-qk,fig:copy-kk,fig:copy-horizon}.

For all of the seeds here, testing CE declines with training CE simultaneously.
This shows no Grokking phenomenon in this task.
By combining \cref{fig:copy-qk,fig:copy-kk}, we observe that $T_g$ and $T^{KK}$ are small in all seeds, while $T^{QK}$ exhibits distinct behavior across seeds.
Hence, the ordering $T^{KK}<T_g<T^{QK}$ does not hold in this case.
To test the long-time behavior, we consider all these eight seeds and refit their locations of $T^{KK}$, $T^{QK}$ for longer total updates $T$.
We show the result in \cref{fig:copy-horizon}.


In \cref{fig:copy-horizon}, $T^{KK}$ remains small, while $T^{QK}$ exhibits jumps at large $T$.
This indicates that the key geometry does not reorganize at all.
Consequently, there is no memorization procedure and therefore no Grokking phenomenon.

\subsection{Online Prediction}

In this subsection, we conduct online prediction experiments for Grokking.
In particular, we test whether changes in key-key ETOCs provide advance warnings of Grokking.
We use the history available during training, and evaluate whether Grokking follows each warning in $2{,}000$ steps.

\paragraph{Setup.}
We test whether ETOCs provide an advance warning of Grokking. 
The two-mean fit for change-time localization requires observations on both sides of a split. 
Instead, an online warning requires a decision only from the training history. 
To make this decision, we replay the recorded query-key and key-key signals in temporal order and detect rapid changes over a short window.

\paragraph{Data and Architecture.}
We reuse the same data, same embedding, same one-layer Transformer and training of the modular addition (\cref{sec:modular_add_exp}), modular non-zero multiplication (\cref{sec:modular_multipli_exp}), and copying experiment (\cref{sec:exp_copy}).
In this subsection, we use 8 addition runs, 8 multiplication runs, and 4 copying runs with measurements every 100 steps.

\paragraph{Observables.}
We measure the following quantities during the training:
\begin{itemize}
    \item \textbf{Key-Key Change.}
	We drop the measurements below step $1{,}000$, and we fit a straight line to the five most recent key-key ETOC values.
	The absolute difference between the line's first and last fitted values measures the change.
	A change of at least $0.025$ in key-key ETOC activates the warning procedure.

	\item \textbf{Warning Time $T_w$.}
	We issue a warning when the fitted change has magnitude at least $0.025$ at two consecutive checks.
	The second check defines our warning time $T_w$.

	\item \textbf{Advance Warning $(T_g-T_w)$.}
	For this experiment, $T_g$ is defined as in \cref{def:grokking_time}. 
	The difference $(T_g-T_w)$ is the training steps between the warning and Grokking.

	\item \textbf{Success and False-Alarm Rate.}
	We count a warning as successful if Grokking follows within $2{,}000$ steps, i.e., $0<(T_g-T_w)\leq2{,}000$.
	This defines the success rate.
	For the copy tasks, any warning counts as a false alarm.
	This defines the false-alarm rate.
    
\end{itemize}

\paragraph{Results.}

With the $2{,}000$-step window, the rule succeeds in $7/8$ addition runs and $6/8$ multiplication runs, for $13/16$ successes ($81.25\%$).
All four copy experiments receive no warning.
We summarize the success rates and the false-alarming rates in \cref{tab:online_prediction}.

\begin{figure}[t]
    \centering
    \includegraphics[width=1\linewidth]{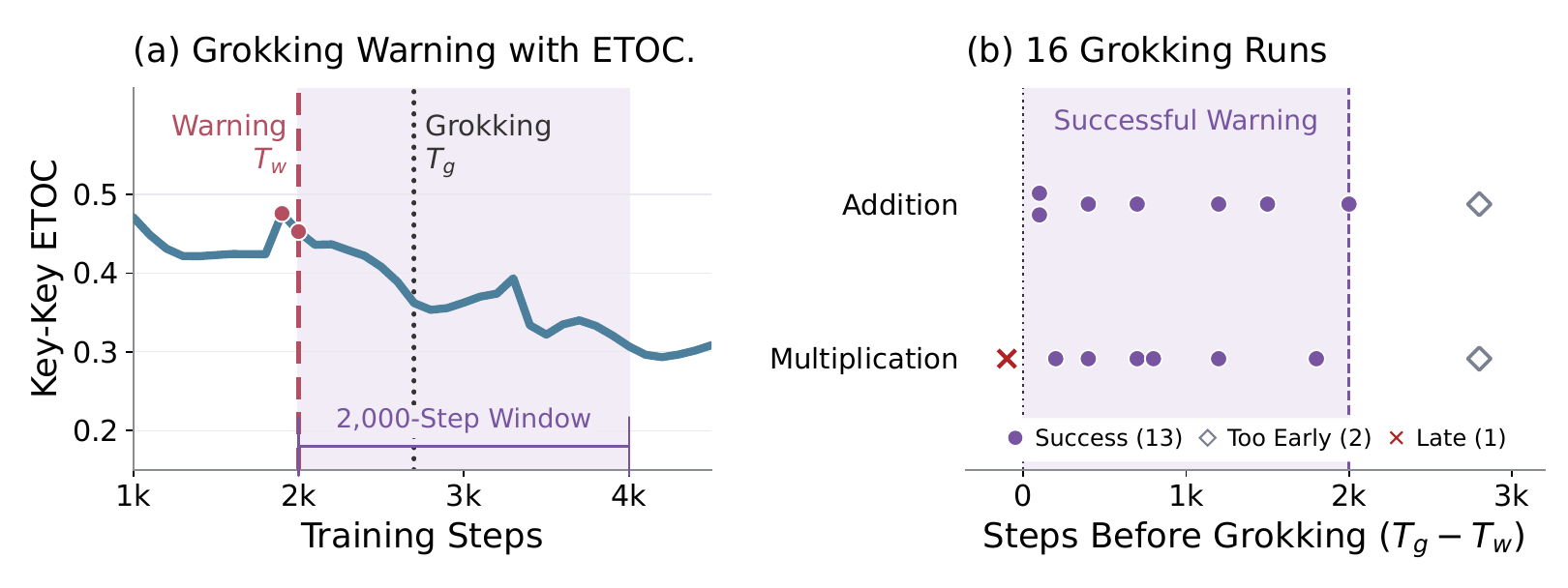}
    \caption{\textbf{Online Prediction Experiment of Grokking.}
	{\small
	\textbf{(a) Grokking Warning with ETOC.} 
    The curve shows the key-key ETOC for addition modulo $113$, seed $16$.
	The two marked observations correspond to consecutive checks that satisfy the change threshold.
    This defines a warning at $T_w=2{,}100$.
	Shading and the bracket mark the $2{,}000$-step forecast window after the warning.
	\textbf{(b) 16 Grokking Runs.} 
    Each marker gives the number of steps from our warning to Grokking.
	The shaded region contains successful warnings under $0<(T_g-T_w)\leq2{,}000$ window.
	Here $T_g$ is the Grokking time from \cref{def:grokking_time}.
	}}
    \label{fig:online_prediction}
\end{figure}

\begin{table}[t]
	\centering
	\setlength{\tabcolsep}{5pt}
	\renewcommand{\arraystretch}{1.2}
	\begin{tabular*}{0.96\linewidth}{@{\extracolsep{\fill}}lccrrr@{}}
		\toprule
		& \multicolumn{2}{c}{Warning Rates} & \multicolumn{3}{c}{Steps Before Grokking} \\
		\cmidrule(lr){2-3}\cmidrule(l){4-6}
		Task & Success & False Warning & Median & Shortest & Longest \\
		\midrule
		Addition & 7/8 (87.5\%) & --- & 700 & 100 & 2,000 \\
		Multiplication & 6/8 (75\%) & --- & 750 & 200 & 1,800 \\
		\addlinespace[3pt]
		\textbf{Arithmetic Total} & \textbf{13/16 (81.25\%)} & --- & 700 & 100 & 2,000 \\
		\midrule
		Copy Controls & --- & 0/4 (0\%) & --- & --- & --- \\
		\bottomrule
	\end{tabular*}
    \caption{\textbf{Online Prediction with a 2,000-Step Window.}
	{\small
	Success requires $0<(T_g-T_w)\leq2{,}000$ on arithmetic runs.
	The false-warning rate counts copy experiments with any warning.
	Median, shortest, and longest intervals refer to successful warnings.
	A dash marks a metric that does not apply to that task.
	}}
    \label{tab:online_prediction}
\end{table}

\clearpage
\section{Discussion and Conclusion}
\label{sec:conclusion}

We establish the duality between reproducing kernel Hilbert space (RKHS) representations in machine learning and coherent-state representations in quantum physics.
Concretely, the kernel in RKHS corresponds to the inner product squared in coherent space.
Through the Moyal-Weyl coherent space, we construct a quantum theory for Transformer and express both softmax and Gaussian attention through query and key projectors (\cref{thm:quantum_softmax_rbf}).
Inspired by quantum chaos, we use equal-time-ordered correlators (ETOCs) (\cref{def:query_key_etoc,def:key_key_etoc}) to characterize non-commutativity and geometry in the learned representations.
When both query-key $T^{QK}$ and key-key ETOC change-time $T^{KK}$ stabilize, the experiments satisfy
\begin{align}
\label{eq:tkk_tg_tqk}
    T^{KK}<T_g<T^{QK},
\end{align}
where $T_g$ is the Grokking time.
This gives an approach to locate Grokking via attention geometry.


\paragraph{Geometric Interpretation.}
The magnitudes of coherent-state inner products determine how far apart the quantum states lie and the relative angle between those quantum states.
During Transformer training, the inner products between keys and queries change dynamically.
This corresponds to a dynamical shaping of quantum geometry.
Through this perspective, the earlier key-key ETOC change \eqref{eq:tkk_tg_tqk}
indicates a reorganization of key geometry.
Thus, this is the formation of representations that support generalization.
On the other hand, the later query-key ETOC change \eqref{eq:tkk_tg_tqk} indicates how the averaged query and key projectors evolve together. 
We interpret it as a response of their collective geometry after generalization improves. 
Together, these observations give a geometric interpretation of the transition from memorization toward generalization.


\paragraph{A New Physics-for-AI Route.}
Our framework establishes an exact duality of Transformers without relying on a large-width limit (e.g., NN-QFT correspondence \cite{halverson2021neural}) or other approximation schemes like mean-field methods (e.g., \cite{mei2018mean}).
The distinction between the prior work rests on the following formulation:
\begin{itemize}
    \item \textbf{Neural-Network-Quantum-Field-Theory (NN-QFT) Correspondence.} 
    The NN-QFT correspondence identifies neural-network covariance kernels with two-point functions in quantum field theory around the large-width limit \cite{halverson2021neural,demirtas2024neural}.
    These large-width perturbative formulations organize corrections through Feynman diagrams and expansions in inverse-width or the depth-to-width ratio \cite{dyer2019asymptotics,grosvenor2021edge,cohen2021learning,roberts2022principles,guillen2025finite}.
    Applying these approximations requires controlling truncation errors in the network's parameter regime.
    
    \item \textbf{Asymptotic Theories.}
    Large-width and mean-field approaches describe feature learning in $\mu$P, Transformer training dynamics, and neural scaling laws \cite{mei2018mean,yang2020feature,bordelon2023dynamics,bordelon2024infinite,bordelon2025feature,zhang2025neural}.
    Applying these theories to a finite network requires controlling deviations from the limits.

    \item \textbf{Our Coherent-State Construction.}
    Our approach here does not require any perturbative parameter to dual a Transformer to a quantum theory.
    That is, our quantum duality and ETOC constructions hold at finite network width, finite $\theta$ without a perturbative expansion (\cref{thm:quantum_softmax_rbf,def:query_key_etoc,def:key_key_etoc}).
    Here $\theta$ corresponds to temperature or covariance in attention \cref{thm:quantum_softmax_rbf}.
\end{itemize}
Our perspective opens a route to quantum analysis of Transformer representation dynamics without an infinite-width or weak-coupling approximation.


\paragraph{Limitations.}
Our work have two main limitations.
First, our experiments only examine modular addition, non-zero modular multiplication, and first-input copying tasks. 
Second, although we establish a quantum theory for Transformers with arbitrary depth, all of our experiments use only a single Transformer block, rather than a deep Transformer.
We leave the experiments of deeper models and broader tasks for future work.


\paragraph{Practical Implications.}
Our results provide the following practical implications.
\begin{itemize}
    \item \textbf{Quantum-RKHS Duality.}
    We connect RKHS in machine learning to quantum mechanics.
    This establishes a perspective to study kernel regression in terms of operators and geometry in quantum mechanics.
    
    \item \textbf{Quantum Perspective on Transformer.}
    We provide a way to study Transformer representations through quantum mechanical formalism.
    This encode additional operator, geometric, and topological information for each Transformer layer.
    Accordingly, it provides a range of potential research routes based on this perspective.
    
    \item \textbf{Locating Grokking.}
    By computing query-key and key-key ETOCs, their stabilized change time interval locates the Grokking time.
    Furthermore, if the key-key ETOC change time remains low and static, the system may not Grok.

    \item \textbf{Predicting Grokking Online.}
    By computing recent key-key ETOC change magnitudes, this issue an advanced warning for Grokking during training.
    This uses only the key-key ETOC history, rather than the data after Grokking.
    
\end{itemize}

\section*{AI Usage Disclosure}
We used large language models (LLMs) to aid, polish writing and debug codes, such as improving clarity, grammar, and conciseness. 
We also used LLMs for retrieval and discovery, for example, exhausting literature to identify potential missing related work. 
Finally, we used LLMs to brainstorm ideas for selected experiments in \cref{sec:5exp}, including suggestions for the model architecture, the number of attention heads, and the hidden dimensions.
All technical content, proofs, and results are original contributions by the authors. The authors take full responsibility for the paper's content and conclusions.

\section*{Acknowledgments}
The authors thank Hikaru Kawai, Pei-Ming Ho, Dino Feng and Andrew Chen for their insights and valuable conversations.
Y.C. is also grateful to Pei-Yuan Wu's lectures on kernel methods.
This sparked the initial idea for this research.
H.L. is supported in part by the National Science and Technology Council (NSTC) grants 115-2811-M-007-070.



\appendix
\label{sec:append}
\part*{Appendix}
{
\setlength{\parskip}{-0em}
\startcontents[sections]
\printcontents[sections]{ }{1}{}
}
\section*{Impact Statement}
By the theoretical nature of this work, we do not anticipate any negative social impact.


\section{Related Work}
\label{sec:related_work}

\paragraph{RKHS and Quantum Mechanics.}

An RKHS represents a positive semi-definite kernel as an inner product \citep{Aronszajn1950TheoryOR}.
\citet{ali:1995zz} explain how quantum-state transformation realize quantum Hilbert states as reproducing kernel spaces, and \citet{chatterjee2016generalized,schuld2019quantum,abanov2025quantum} supports quantum states as kernel feature vectors.
Specifically, the RKHS of feature maps is isometric to the span of quantum states in the complex Hilbert space.
This dual is established via projectors in Hilbert--Schmidt operator space \citep{schuld2021supervised}.
Accordingly, we express attention kernels through projector inner products and study their operator geometry.


\paragraph{Non-Commutative Coherent Space.}
Geometric deformation quantization provides an approach from symplectic manifold to non-commutative coherent space via Weyl quantization map \cite{emmrich1994differential}.
Specifically, this formalism replaces ordinary multiplication rule for functions with a non-commutative star product with the Poisson structure in the classical level.
This provides a natural representation and reference to define coherent states.
\citet{perelomov1977coherent,Steinacker:2024unq} therefore develop the resulting operator geometry of coherent spaces thoroughly.
Specifically, coherent state inner product magnitudes describe separation, and phases contain symplectic and geometric information.
Additionally, commutation relation of the coherent space is able to extend to a general Lie algebra with the corresponding geometry \cite{perelomov1977coherent}.
These structures also provide matrix-model approaches to quantum gravity: how gravity and spacetime emerge from matrices \cite{ishibashi1997IKKT,banks1999m,liao2026saddle,ho2026general}.
In particular, coherent-state expectation values describe positions, and their inner products encode non-commutative geometric relations \cite{steinacker2021quantum}.
However, these works mainly focus on physics side, rather than the machine learning tasks nor architectures.
Hence, we adopt those abundant ideas to construct the simplest coherent space formalism for Transformers.


\paragraph{Mechanisms of Grokking.}
\citet{power2022grokking} discover delayed generalization, the Grokking, after fitting small algorithmic datasets.
To explain this delay, \citet{nanda2023progress} identify Fourier periodic structures in modular addition and measure their formation before test accuracy improve.
Complementarily, \citet{kumar2024grokking} link grokking to delayed feature learning in polynomial regression.
However, these studies interpret Grokking through analyzing the training behaviors, rather than the relations among query and key representations.
Beyond algorithmic tasks, \citet{liu2022omnigrok} induce Grokking on MNIST images, IMDb reviews, and QM9 molecules tasks using selected data sizes and large initial weight norms.
In their account, training loss falls quickly at large weight norms, and weight decay or implicit regularization then slowly reduces the norm toward lower test loss.
On the other hand, from a statistical-mechanical viewpoint, \citet{rubin2024grokking} derive first-order transitions in two-layer teacher--student models under equilibrium and large-system assumptions.
Together, these results link Grokkings to representation learnings. 
However, the mechanism depends on specific training regime or simplified architectures, and these works do not probe the non-commutative geometry of representations behind Transformers. 
Our work therefore studies this geometry of representations via the quantum-mechanical formalism.


\paragraph{Early Indicators of Grokking.}
Beyond explaining why Grokking occurs, another goal is to identify them during training.
For this purpose, \citet{tan2023understanding} study how input perturbations change the information of learned representation.
They suggest that small or decreasing differences in early training indicate that a model will Grok eventually.
On the other hand, \citet{xu2026early} report early signals of Grokking from comparing different orders of mini-batch gradient updates via two tasks: (1) SCAN, which translates commands into action sequences (2) Dyck-1, which predicts the number of unmatched opening parentheses at each position. 
However, they do not measure non-commutativity among the query and key representations and lack the physical meaning in terms of memorization toward generalization.
Hence, we provide a concrete measure (ETOC) to locate Grokking using query and key geometric dynamics.


\paragraph{Quantum Chaos.}
Quantum mechanics provides abundant information than in the classical theory.
One of the benefits is the quantum chaos analysis.
Specifically, classical Lyapunov exponents characterize the chaos of a classical trajectory, while out-of-time-ordered correlators (OTOCs) connect Lyapunov exponents and go beyond through the non-commutativity of operators \cite{rozenbaum2017lyapunov}.
OTOCs probe operator spreading and provide measures of scrambling when initial local information spreads across a many-body system \citep{xu2024scrambling}.
Specifically, physicists adopt OTOCs to study integrable systems \cite{xu2020does}, thermal chaos bounds \cite{maldacena2016bound}, scrambling in Sachdev–Ye–Kitaev (SYK) models \cite{maldacena2016remarks}, butterfly effects in black holes \cite{shenker2014black}, and so on.
These related works motivate our ETOCs as geometric diagnostics of learned representations to study Grokking in Transformers.

\section{Brief Review on Hilbert Space and Algebra}
\label{sec:brief_review_hilbert_space}

We review Hilbert-space and algebraic notions to describe attention through quantum states and operators. 
We first distinguish bounded and unbounded operators, then introduce operator algebras and group representations, and finally connect continuous frames
with reproducing kernels.


\subsection{Hilbert Space and Operators}

This subsection defines the Hilbert space and operators.
We further provide several examples of unbounded and bounded operators. 

\begin{definition}[Cauchy Sequence and Hilbert Space]
    A sequence $\{u_n\}_{n\geq1}$ in a vector space $V$ is Cauchy if for every $\epsilon>0$, there exists $N$ such that $\|u_n-u_m\|_V<\epsilon$ whenever $n,m\geq N$.
    A vector space with inner product structure is a Hilbert space if every Cauchy sequence converges to an element of the space.
    This convergence condition is called the completeness.
\end{definition}

In this definition of Hilbert space, we do not require finite-dimensionality or a choice of basis.

\begin{definition}[Linear Map, Endomorphism, and Operator]
\label{def:linear_map_operator}
    Let $V,W$ be two vector spaces.
    \begin{itemize}
        \item \textbf{Linear map.}
        A map $\hat{A}:V\to W$ is linear if $\hat{A}(\alpha u+\beta v)=\alpha \hat{A}u+\beta \hat{A}v$ for all $u,v\in V$ and $\alpha,\beta\in\mathbb{C}$.
        
        \item \textbf{Endomorphism.}
        A linear map $\hat{A}:V\to V$ is an endomorphism of $V$. 
        An endomorphism need not be injective, surjective, nor invertible.
        We call an endomorphism $\hat{A}$ an operator in this paper.
    \end{itemize}
    
\end{definition}

\begin{definition}[Boundedness]
\label{def:bounded}
    Consider a linear map $\hat{A}:\mathcal{H}_1\to\mathcal{H}_2$ with two Hilbert spaces $\mathcal{H}_1,\mathcal{H}_2$.
    We say $\hat{A}$ is bounded if there exists $C<\infty$ such that
    \begin{align*}
        \|\hat{A} u\|_{\mathcal{H}_2}\leq C\|u\|_{\mathcal{H}_1}\quad\forall u\in\mathcal{H}_1.
    \end{align*}
    We denote the space of bounded linear maps as $\mathcal{B}(\mathcal{H}_1,\mathcal{H}_2)$.
    For bounded operators, we denote the space as $\mathcal{B}(\mathcal{H})\coloneq\mathcal{B}(\mathcal{H},\mathcal{H})$.
\end{definition}

Every linear map between finite-dimensional Hilbert spaces is bounded. 

\begin{example}[Momentum Operator is Unbounded]
    Consider the Hilbert space $\mathcal{H}=L^2(\R)$ with momentum operator $\hat{P}=-i\frac{d}{dx}$ and its domain 
    \begin{align*}
        \text{Dom}(\hat{P})=\{\psi\in L^2(\R)\mid\frac{d}{dx}\psi\in L^2(\R)\}.
    \end{align*}
    If we consider $\psi_n(x)\coloneq e^{inx}\phi(x)$ with $\|\psi_n\|_{L^2}=1$, then
    \begin{align*}
        \hat{P}\psi_n =-i\frac{d}{dx}\Big(e^{inx}\phi(x)\Big)=ne^{inx}\phi(x)-i e^{inx}\frac{d\phi(x)}{dx}.
    \end{align*}
    Hence, $\|\hat{P}\psi_n\|_{L^2}\geq n\|\phi\|_{L^2}-\|\frac{d}{dx}\phi\|_{L^2}\to\infty$ as $n\to\infty$.
    There is no finite constant $C$ such that $\|\hat{P}\psi_n\|_{L^2}\leq C\|\psi_n\|_{L^2}$, and thus $\hat{P}$ is unbounded.
\end{example}

\begin{definition}[Adjoint, Hermiticity, and Projection]
\label{def:adj_hermitian_projection}
    Let $\hat{A}\in\mathcal{B}(\mathcal{H})$ be a bounded operator with a Hilbert space $\mathcal{H}$.
    Then we define adjoint, Hermiticity, and projection as follows.
    \begin{itemize}
        \item \textbf{Adjoint.}
        We denote the adjoint of $\hat{A}$ as $\hat{A}^\dagger$.
        It satisfies
        \begin{align}
            \langle \hat{A} u,v\rangle_{\mathcal{H}}=\langle u,\hat{A}^\dagger v\rangle_{\mathcal{H}}\qquad\forall u,v\in\mathcal{H}.
        \end{align}

        \item \textbf{Hermiticity.}
        If $\hat{A}$ is Hermitian, then it satisfies $\hat{A}^\dagger=\hat{A}$.
        
        \item \textbf{Projection.}
        If $\hat{A}$ is a projection, then it satisfies $\hat{A}^2=\hat{A}=\hat{A}^\dagger$.
    \end{itemize}
\end{definition}

\begin{remark}
    If $\mathcal H$ is finite-dimensional, every linear operator
	$\hat{A}:\mathcal{H}\to\mathcal{H}$ is bounded, and its adjoint
	corresponds to the conjugate transpose of its matrix in an
	orthonormal basis.
	In infinite dimensions, both bounded and unbounded operators exist.
\end{remark}

\begin{remark}
    For an unbounded operator $\hat{A}$, Hermiticity requires both $\hat{A}u = \hat{A}^\dagger u$ and additional $\text{Dom}(\hat{A}) = \text{Dom}(\hat{A}^\dagger)$ on this domain for every $u\in\text{Dom}(\hat{A})$ \cite{bcHall2020quantumformath}. 
\end{remark}

\begin{example}[Translation Operator is Bounded]
\label{exp:translation_opt_bdd}
    Consider the Hilbert space $\mathcal{H}=L^2(\R)$ with translation operator $\hat{U}(a)\coloneq\exp(-ia\hat{P})$ and momentum operator $\hat{P}=-i\frac{d}{dx}$.
    Let $\psi\in L^2(\R)$ be a normalized function $\|\psi\|_{L^2}=1$.
    Then, acting $\hat{U}(a)$ on $\psi$, we have
    \begin{align*}
        \hat{U}(a)\psi(x)=e^{-a\frac{d}{dx}}\int_{\R}\tilde{\psi}(p)e^{-ipx}~\frac{\dd p}{2\pi}=\int_{\R}\tilde{\psi}(p)e^{-ip(x-a)}~\frac{\dd p}{2\pi}=\psi(x-a).
    \end{align*}
    Hence, $\|\hat{U}\psi\|_{L^2}=\|\hat{U}\|_{L^2}\|\psi\|_{L^2}=1$ indicates that $\hat{U}$ is bounded on an infinite-dimensional $\mathcal{H}$.
\end{example}

In our construction, the Hilbert space of coherent space remains infinite-dimensional even at finite Transformer width.
The coordinate operators $\hat{M}^a$ are unbounded, while the coherent-state projectors $\hat{P}_x$ and spatial translation operators $\hat{U}_x$ belong to $\mathcal{B}(\mathcal{H})$.


\subsection{Operator Algebras and Group Representations}

This subsection describes how operators compose and how group operations represent on a Hilbert space. 
A matrix representation depends on a chosen basis, but these algebraic relations do not.

\begin{definition}[Operator Algebra and Commutator]
    An operator algebra $\mathfrak{g}$ is a complex linear subspace of $\mathcal{B}(\mathcal{H})$.
    For $\hat{A},\hat{B}\in\mathfrak{g}$, there commutator is
    \begin{align*}
        \big[\hat{A},\hat{B}\big]\coloneq\hat{A}\hat{B}-\hat{B}\hat{A}.
    \end{align*}
    The algebra is commutative if $[\hat{A},\hat{B}]=0$ for all $\hat{A},\hat{B}\in\mathfrak{g}$.
    The commutator is bilinear, anti-symmetric and satisfies the Jacobi identity, i.e.,
    \begin{align*}
        \big[\hat{A},\big[\hat{B},\hat{C}\big]\big]+\big[\hat{B},\big[\hat{C},\hat{A}\big]\big]+\big[\hat{C},\big[\hat{A},\hat{B}\big]\big]=0.
    \end{align*}
    This defines a Lie bracket on $\mathfrak{g}$.
\end{definition}

\begin{proposition}[Baker-Campbell-Hausdorff Formula]
\label{prp:bch_formula}
    Consider two operators $\hat{A},\hat{B}\in\mathcal{B}(\mathcal{H})$.
    If $[\hat{A},[\hat{A},\hat{B}]]=[\hat{B},[\hat{A},\hat{B}]]=0$, we have a simplified Baker-Campbell-Hausdorff (BCH) formula
    \begin{align*}
        e^{\hat{A}}e^{\hat{B}}=e^{\hat{A}+\hat{B}+\frac12[\hat{A},\hat{B}]}.
    \end{align*}
    In particular, this condition holds when $[\hat{A},\hat{B}]$ is a scalar multiple of the identity operator.
    For the proof, please see \cite{hall2015lie}, Section 5.
\end{proposition}

\begin{definition}[Group and Representation]
\label{def:group_rep}
    To illustrate how to represent an operator as a matrix, we define the "group" and its "representation" as follows.
    \begin{itemize}
        \item \textbf{Group}.
        A group is a set $G$ with a binary operation $\circ$ satisfying
        \begin{align*}
            &g_1\circ(g_2\circ g_3)=(g_1\circ g_2)\circ g_3\qquad\forall g_1,g_2,g_3\in G,\annot{Associativity}\\
            &\exists e\in G~\text{ s.t. }~ g\circ e=e\circ g=g\qquad\forall g\in G,\annot{Identity element}\\
            &\forall g\in G~\exists g^{-1}\in G~\text{ s.t. }~g\circ g^{-1}=g^{-1}\circ g=e.\annot{Inverse element}
        \end{align*}
        
        \item \textbf{Representation.}
        A representation of $G$ on a Hilbert space $\mathcal{H}$ is a homomorphism $\rho:G\to\text{GL}(\mathcal{H})$, where $\text{GL}(\mathcal{H})$ is a general linear group
        \begin{align*}
            \text{GL}(\mathcal{H})\coloneq\{\hat{A}\in\mathcal{B}(\mathcal{H})\mid \hat{A}^{-1}\in\mathcal{B}(\mathcal{H})\}.
        \end{align*}
        Hence, $\rho(g\circ h)=\rho(g)\rho(h)$ for all $g,h\in G$, and $\rho(e)=\one$.

        \item \textbf{Unitary representation.}
        A representation $\rho$ is unitary if 
        \begin{align*}
            \rho(g)\in\{\hat{U}\in\mathcal{B}(\mathcal{H})\mid \hat{U}^\dagger \hat{U}=\hat{U}\hat{U}^\dagger=\hat{\one}_\mathcal{H}\},
        \end{align*}
        for all $g\in G$.
    \end{itemize}
\end{definition}

For example, $\hat{U}(a)\hat{U}(b)=\hat{U}(a + b)$, so the translations \cref{exp:translation_opt_bdd} give a unitary representation of the additive group $\R$. 
A unitary group representation and a reference state supply the group orbit in coherent-state constructions \cite{perelomov1977coherent}. 
A resolution of the identity additionally requires the frame condition below.


\subsection{Continuous Frames and Reproducing Kernels}

This subsection introduces the frame to resolve the identity and the positivity condition that construct an reproducing kernel Hilbert space (RKHS).

\begin{proposition}[Riesz Representation Theorem \cite{conway2007functional,paulsen2016rkhs}]
\label{prp:riesz_rep_thm}
    Let $\mathcal{H}$ be a Hilbert space.
    Then for every bounded linear functional $f:\mathcal{H}\to\mathbb{C}$, there exists a unique $h_f\in\mathcal{H}$ such that
    \begin{align*}
        f(u)=\langle h_f,u\rangle_{\mathcal{H}}\qquad\forall u\in\mathcal{H},
    \end{align*}
    and $\|h_f\|_{\mathcal{H}}=\sup_{\|u\|\leq1}|f(u)|$.
\end{proposition}

For an RKHS, applying this theorem to the bounded function $f(x)$ gives the kernel $K(\cdot,x)$.
The following frame condition gives a resolution of the identity for a family of states.

\begin{definition}[Weakly Measurable and Continuous Parseval Frame \cite{ALI19931,conway2007functional}]
\label{def:parseval_frame}
    Let $(\mathcal{X},\Sigma,\mu)$ be a measure space, where $\Sigma$ is a $\sigma$-algebra of subsets of $\mathcal{X}$, $\mu$ is a measure on $\Sigma$, and $\mathcal{H}$ is a Hilbert space.
    A map $\psi:\mathcal{X}\to\mathcal{H}$ is weakly measurable if $x\mapsto\langle u,\psi(x)\rangle_{\mathcal{H}}$ is measurable for every $u\in\mathcal{H}$ and $x\in\mathcal{X}$.
    In the weak-measurable perspective, we write
    \begin{align*}
        \langle u,v\rangle_{\mathcal{H}}=\int_{\mathcal{X}}\langle u,\psi(x)\rangle_{\mathcal{H}}\langle\psi(x),v\rangle_{\mathcal{H}}~\dd\mu(x)\qquad\forall u,v\in\mathcal{H}.
    \end{align*}
    In physics notation, we denote $\psi(x)$ as $\ket{x}\in\mathcal{H}$ and write this condition as
    \begin{align*}
        \int_{\mathcal{X}}\ket{x}\bra{x}\dd\mu(x)=\hat{\one}_{\mathcal{H}},
    \end{align*}
    where the integral is weakly through the inner-product identity.
    We call this family of $\mathcal{H}$ with continuous index set $\mathcal{X}$ a continuous Parseval frame.    
\end{definition}

The frame condition considers the completeness of a family of states with a chosen measure.
Kernel positivity is a separate condition.

\begin{definition}[Hermitian and Positive Semi-Definite (PSD) Kernel]
    Let $\mathcal{X}$ be a non-empty set.
    A function $K:\mathcal{X}\times\mathcal{X}\to\mathbb{C}$ is Hermitian if 
    \begin{align*}
        K(x,y)=\overline{K(y,x)}\qquad\forall x,y\in\mathcal{K}.
    \end{align*}
    It is positive semi-definite (PSD) kernel if 
    \begin{align*}
        \sum_{i,j=1}^{N}\overline{c_i} c_j K(x_i,x_j)\geq 0,\qquad\forall N\in\mathbb{N}\;\forall x_i\in\mathcal{X}\;\forall c_i\in\mathbb{C}.
    \end{align*}
    Equivalently, every $\{K(x_i,x_j)\}_{i,j=1}^N$ is a Hermitian PSD matrix.
\end{definition}

\begin{proposition}[Moore-Aronszajn Correspondence \cite{Aronszajn1950TheoryOR}]
    The reproducing kernel of every RKHS (\cref{def:reproducing_kernel_rkhs}) is Hermitian PSD.
    Conversely, every Hermitian PSD kernel on a non-empty set $\mathcal{X}$ is the reproducing kernel of a unique RKHS.
\end{proposition}

For any map $x\mapsto\ket{x}\in\mathcal{H}$, the kernel $K(x,y)=\bra{x}\ket{y}$ is a Hermitian PSD kernel (\cref{prp:coherent-rkhs}). 
Since PSD condition is separate from \cref{def:parseval_frame}, the quantum-RKHS duality does not require additional or stronger Parseval-frame condition.

\section{Transformer Training}
\label{sec:transformer_training}

Here we introduce the Transformer training process
with a fixed architecture in \cref{sec:moyal_weyl_transformer}.
During training, some parameters or projections update via gradient-based algorithm.
We denote them as a trainable parameter set
\begin{align*}
    \Theta_t\coloneq\Big\{\big\{W_{Q,\ell,t}^{(n)},W_{K,\ell,t}^{(n)},W_{V,\ell,t}^{(n)}\}_{n=1}^H,\big\{\gamma_{1;r,,\ell,t},\beta_{1;r,\ell,t}\big\}_{r=1}^{d_{{\rm model}}},W_{O,\ell,t},\cdots\Big\}_{\ell=1}^L.
\end{align*}
Below we consider AdamW optimizer \cite{kingma2014adam} for Transformer training.
Given a dataset $\mathcal{D}$ of $|\mathcal{D}|$ input-target pairs $(X,c)$, we define cross-entropy loss (CE) and its corresponding accuracy as
\begin{align}
    \mathcal L_{\mathcal D}(\Theta_t)
    = & ~
    -\frac1{|\mathcal D|}\sum_{(X,c)\in\mathcal D}\log\pi_{\Theta_t}(c\mid X),\label{eq:cross_entropy_loss}\\
    \text{Acc}_{\mathcal D}(t)
    = & ~
    \frac1{|\mathcal D|}\sum_{(X,c)\in\mathcal D}\one\left\{\arg\max_{c'\in\mathcal Y}\pi_{\Theta_t}(c'\mid X)=c\right\},\notag
\end{align}
where $\pi_{\Theta_t}$ is the class probability \eqref{eq:class_score_probability}, and the indicator $\one\{A\}$ equals one when statement $A$ holds and zero otherwise. 
$\mathcal{L}_{\mathcal{D}}$ measures the probability penalty on the correct class, and $\text{Acc}_{\mathcal{D}}$ measures the correctness of class predictions.
Our training procedure then uses full-batch AdamW in each update \cite{loshchilov2017decoupled}. 
That is, let $g_t=\nabla_{\Theta_t}\mathcal L_{\mathcal D}(\Theta_t)$ be the loss gradient, $\tilde{m}_t$ and $\tilde{v}_t$ be
\begin{align*}
	m_{t+1}&\coloneq\beta_1m_t+(1-\beta_1)g_t,&
	v_{t+1}&\coloneq\beta_2v_t+(1-\beta_2)g_t\odot g_t,\\
	\tilde{m}_{t+1}&\coloneq\frac{m_{t+1}}{1-\beta_1^{t+1}},&
	\tilde{v}_{t+1}&\coloneq\frac{v_{t+1}}{1-\beta_2^{t+1}},
\end{align*}
where the hats remove the bias from the zero initial averages $m_0=v_0=0$, and the symbol $\odot$ denotes coordinate-wise multiplication.
AdamW then utilizes these quantities to realize a parameter update rule during training via
\begin{align}
\label{eq:adamw}
    \Theta_{t+1}\coloneq(1-\eta\lambda)\Theta_t-\eta\frac{\tilde{m}_{t+1}}{\sqrt{\tilde{v}_{t+1}}+\epsilon_{\text{Adam}}}.
\end{align}


\section{Proofs of Main Text}
\label{sec:proof}

In this appendix, we provide the detailed proofs of theorems and propositions in the main text.


\subsection{Proof of \texorpdfstring{\cref{prp:creation_annihilation}}{}}
\label{sec:proof_creation_annihilation}

\begin{proposition}[\cref{prp:creation_annihilation} Restated: Canonical Commutation Relation]
    Consider the canonical coordinates in a $2d$-dimensional Moyal-Weyl quantum plane.
    For each block, we have the commutation relation
    \begin{align*}
        \big[\hat{Q}^i,\hat{P}^j\big]=i\theta_i\delta^{ij}\hat{\one},\qquad\big[\hat{Q}^i,\hat{Q}^j\big]=\big[\hat{P}^i,\hat{P}^j\big]=0.
    \end{align*}
    We call these relations the canonical commutation relations.
    These relations allow us to construct lowering and raising operators as
    \begin{align}
    \label{eq:app_annihilation_creation}
        \hat{a}_i\coloneq\frac{1}{\sqrt{2\theta_i}}(\hat{Q}^i+i\hat{P}^i),\qquad \hat{a}^\dagger_i=\frac{1}{\sqrt{2\theta_i}}(\hat{Q}^i-i\hat{P}^i),
    \end{align}
    respectively.
    Specifically, they satisfy the following commutation relation
    \begin{align*}
        [\hat{a}_i,\hat{a}^\dagger_j]=\delta_{ij}\hat{\one}.
    \end{align*}
\end{proposition}

\begin{proof}
    This results is well-known in quantum mechanics but we still restate here for completeness. 
    If we define $\hat{Y}^a\coloneq(O^\top \hat{M})^a=\sum_{b=1}^{2d}O_{b}{}^a\hat{M}^b$, then
    \begin{align*}
        [\hat{Y}^a,\hat{Y}^b] = & ~
        \Big[\sum_{n=1}^{2d}O_n{}^a\hat{M}^n,\sum_{m=1}^{2d}O_m{}^b\hat{M}^m\Big]\\
        = & ~
        \sum_{n,m=1}^{2d}O_n{}^aO_m{}^b\big[\hat{M}^n,\hat{M}^m\big]\\
        = & ~
        i\sum_{n,m=1}^{2d}O_n{}^aO_m{}^b\theta^{nm}\hat{\one}\annot{By \eqref{eq:moyal_weyl_commutation}}\\
        = & ~
        i(O^\top\theta O)^{ab}\hat{\one}
    \end{align*}
    Then from \eqref{eq:block_diagonalize_theta}, we identify 
    \begin{align}
    \label{eq:canonical_q_p_from _x}
        \hat{Q}^i=\hat{Y}^{2i-1}=(O^\top \hat{M})^{2i-1},\qquad \hat{P}^i=\hat{Y}^{2i}=(O^\top \hat{M})^{2i},
    \end{align} 
    with $i=1,\cdots,d$.
    In this way, the block diagonal representation gives 
    \begin{align*}
        [\hat{Q}^i,\hat{P}^j] = & ~ 
        \big[\hat{Y}^{2i-1},\hat{Y}^{2i}\big]=i(O^\top\theta O)^{2i-1~2j}\hat{\one}=i\theta_i\delta^{ij}\hat{\one}.
    \end{align*}
    Similarly, the block structure gives $[\hat{Q}^i,\hat{Q}^j]=[\hat{P}^i,\hat{P}^j]=0$.
    For the annihilation and creation operators, we compute
    \begin{align*}
        [\hat{a}_i,\hat{a}_j^\dagger] = & ~
        \frac{1}{2\sqrt{\theta_i\theta_j}}\big[\hat{Q}^i+i\hat{P}^i,\hat{Q}^i-i\hat{P}^i\big]\annot{By \eqref{eq:app_annihilation_creation}}\\
        = & ~
        \frac{1}{2\sqrt{\theta_i\theta_j}}\Big(\big[\hat{Q}^i,\hat{Q}^j\big]-i\big[\hat{Q}^i,\hat{P}^j\big]+i\big[\hat{P}^i,\hat{Q}^j\big]+\big[\hat{P}^i,\hat{P}^j\big]\Big)\\
        = & ~
        \frac{1}{2\sqrt{\theta_i\theta_j}}\cdot\big(-2i\cdot i\theta_i\delta^{ij}\hat{\one}\big)\\
        = & ~
        \delta^{ij}\hat{\one}.
    \end{align*}
    This completes the proof.
\end{proof}

\subsection{Proof of \texorpdfstring{\cref{prp:kernel_moyal_weyl_coherent_inner_product}}{}}
\label{sec:proof_kernel_moyal_weyl_coherent_inner_product}

\begin{lemma}[Expectation Value of Moyal-Weyl Coordinate \cite{glauber1963coherent}]
\label{prp:expectation_coherent_state}
    Consider coherent states $\ket{x}\in\mathcal{H}$ in \cref{def:moyal_weyl_coherent_state}.
    Then we have the following expectation value 
    \begin{align}
    \label{eq:expectation_value_x}
        \bra{x}\hat{M}\ket{x}=x.
    \end{align}
\end{lemma}

\begin{proof}
    We start from expressing the inner product with respect to the coherent state $\ket{x}$.
    \begin{align*}
        \bra{x}\hat{M}\ket{x} = & ~
        \bra{0}\hat{U}^\dagger_x\hat{M}\hat{U}_x\ket{0}=\bra{0}e^{-ix^\top\theta^{-1}\hat{M}}\hat{M}e^{ix^\top\theta^{-1}\hat{M}}\ket{0}\\
        = & ~
        \bra{0}\Big(\hat{M}-ix^\top\theta^{-1}[\hat{M},\hat{M}]+\cdots\Big)\ket{0}\annot{By \cref{prp:bch_formula}}\\
        = & ~
        \bra{0}\hat{M}\ket{0}-ix^\top\theta^{-1}\bra{0}i\theta\hat{\one}\ket{0}\annot{By \eqref{eq:moyal_weyl_commutation}}\\
        = & ~
        \bra{0}\hat{M}\ket{0}+x.
    \end{align*}
    Since for every $\hat{M}^a$, it is either $\hat{Q}^i=\sqrt{\frac{\theta_i}{2}}(\hat{a}_i+\hat{a}_i^\dagger)$ or $\hat{P}^i=-i\sqrt{\frac{\theta_i}{2}}(\hat{a}_i-\hat{a}_i^\dagger)$, we notice that
    \begin{align*}
        \big(O^\top\bra{0}\hat{M}\ket{0}\big)_i=\bra{0}\big(O^\top \hat{M}\big)_i\ket{0}=\begin{cases}\sqrt{\frac{\theta_i}{2}}\big(\bra{0}\hat{a}_i\ket{0}+\bra{0}\hat{a}_i^\dagger\ket{0}\big)=0,\\-i\sqrt{\frac{\theta_i}{2}}\big(\bra{0}\hat{a}_i\ket{0}-\bra{0}\hat{a}_i^\dagger\ket{0}\big)=0.\end{cases}
    \end{align*}
    In this way, we obtain $\bra{0}\hat{M}\ket{0}=0$ and hence $\bra{x}\hat{M}\ket{x}=x$.
    This completes the proof.
\end{proof}

\begin{proposition}[\cref{prp:kernel_moyal_weyl_coherent_inner_product} Restated: Inner Product in Moyal-Weyl Quantum Plane \cite{Steinacker:2024unq}]
    The coherent states follow the inner product structure
    \begin{align*}
        \bra{x}\ket{y}=\exp\Big\{-\frac{1}{4}\sum_{a,b=1}^{2d}(x-y)_a(\theta^\top\theta)^{-1/2}_{ab}(x-y)_b\Big\}\exp\Big\{-\frac{i}{2}\sum_{a,b=1}^{2d}x_a(\theta^{-1})_{ab}y_b\Big\}.
    \end{align*}
\end{proposition}

\begin{proof}
    Firstly, from \cref{def:moyal_weyl_plane}, we note that
    \begin{align*}
        \Theta\coloneq O^\top\theta O=\bigoplus_{i=1}^d\Theta_i=\bigoplus_{i=1}^d\begin{pmatrix}0&\theta_i\\-\theta_i&0\end{pmatrix},
    \end{align*}
    where $O$ is a $2d$-dimensional orthogonal matrix.
    Hence, $O^\top=O^{-1}$.
    In this way, 
    \begin{align*}
        \theta^{-1} = \big((O^\top)^{-1}\Theta O^{-1}\big)^{-1}=(O\Theta O^\top)^{-1}=O\Theta^{-1}O^\top,
    \end{align*}
    with
    \begin{align*}
        \Theta^{-1}=\bigoplus_{i=1}^d\Theta^{-1}_i=\bigoplus_{i=1}^d\begin{pmatrix}0&-\theta_i^{-1}\\\theta^{-1}_i&0\end{pmatrix}.
    \end{align*}
    Then, we express the phase of Moyal-Weyl coherent states as
    \begin{align*}
        ix^\top\theta^{-1}\hat{M} = & ~
        ix^\top O\Theta^{-1}O^\top \hat{M}=i(O^\top x)^\top\Theta^{-1}(O^\top \hat{M})=i(x')^\top\Theta^{-1}\hat{Y},
    \end{align*}
    where $\hat{Y}^a=(O^\top \hat{M})^a$ with $\hat{Q}^i=\hat{Y}^{2i-1}$ and $\hat{P}^i=\hat{Y}^{2i}$.
    In this way, we block diagonalize the $2d$-dimensional matrix into $(q^i,p^i)$-coordinates.
    Explicitly, for $i$-th block we have
    \begin{align}
        i\begin{pmatrix}q^i&p^i\end{pmatrix}\begin{pmatrix}0&-\theta^{-1}_i\\\theta^{-1}_i&0\end{pmatrix}\begin{pmatrix}\hat{Q}^i\\\hat{P}^i\end{pmatrix} = & ~
        i\sum_{i=1}^d\theta^{-1}_i(p^i\hat{Q}^i-q^i\hat{P}^i)\label{eq:i_th_block_translation_opt}\\
        = & ~
        \sum_{i=1}^d\frac{1}{2\theta_i}\Big[\big(q^i+ip^i\big)\big(\hat{Q}^i-i\hat{P}^i\big)-\big(q^i-ip^i\big)\big(\hat{Q}^i+i\hat{P}^i\big)\Big]\notag\\
        = & ~
        \sum_{i=1}^d\alpha_i \hat{a}^\dagger_i-\alpha^\star_i\hat{a}_i\annot{By \eqref{eq:annihilation_creation}},\notag
    \end{align}
    where
    \begin{align*}
        \alpha_i(x)\coloneq\frac{1}{\sqrt{2\theta_i}}\big[q^i(x)+ip^i(x)\big],\qquad\alpha^\star_i(x)\coloneq\frac{1}{\sqrt{2\theta_i}}\big[q^i(x)-ip^i(x)\big].
    \end{align*}
    Hence, we obtain
    \begin{align*}
        ix^\top\theta^{-1}\hat{M}=\sum_{i=1}^d\big(\alpha_i(x)\hat{a}^\dagger_i-\alpha^\star_i(x) \hat{a}_i\big).
    \end{align*}
    Then 
    \begin{align}
        \hat{U}_x = & ~
        e^{ix^\top\theta^{-1}\hat{M}}\annot{By \cref{def:moyal_weyl_coherent_state}}\notag\\
        = & ~
        \exp\Big\{\sum_{i=1}^d\big(\alpha_i\hat{a}_i^\dagger-\alpha^\star_i\hat{a}_i\big)\Big\}\notag\\
        = & ~
        \exp\Big\{\sum_{i=1}^d\alpha_i\hat{a}^\dagger_i\Big\}\exp\Big\{-\sum_{i=1}^d\alpha^\star_i\hat{a}_i\Big\}\exp\Big\{\frac{1}{2}\Big[\sum_{i=1}^d\alpha_i\hat{a}^\dagger_i,\sum_{j=1}^d\alpha^\star_j\hat{a}_j\Big]\Big\}\notag\\
        & ~
        \times\exp\Big\{\frac{1}{12}\Big[\sum_{i=1}^d\alpha_i\hat{a}^\dagger_i,\cdots\Big]\Big\}\cdots\annot{By \cref{prp:bch_formula}}\notag\\
        = & ~
        \exp\Big\{\sum_{i=1}^d\alpha_i\hat{a}^\dagger_i\Big\}\exp\Big\{-\sum_{i=1}^d\alpha^\star_i\hat{a}_i\Big\}\exp\Big\{\frac{1}{2}\sum_{i,j=1}^d\alpha_i\alpha^\star_j\Big[\hat{a}^\dagger_i,\hat{a}_j\Big]\Big\}\cdots\notag\\
        = & ~
        \exp\Big\{\sum_{i=1}^d\alpha_i\hat{a}^\dagger_i\Big\}\exp\Big\{-\sum_{i=1}^d\alpha^\star_i\hat{a}_i\Big\}\exp\Big\{\frac{1}{2}\sum_{i,j=1}^d\alpha_i\alpha^\star_j(-\one_{\{i,j\}})\Big\}\times1\annot{By \eqref{eq:creation_annihilation_commutation_relation}}\notag\\
        = & ~
        \exp\Big\{\sum_{i=1}^d\alpha_i(x)\hat{a}^\dagger_i\Big\}\exp\Big\{-\sum_{i=1}^d\alpha^\star_i(x)\hat{a}_i\Big\}\exp\Big\{-\frac{1}{2}\|x\|_g^2\Big\},\label{eq:u_x_coherent_unitary_transform}
    \end{align}
    where
    \begin{align}
    \label{eq:x_norm_with_metric}
        \|x\|_g^2\coloneq\sum_{a,b=1}^{2d}g_{ab}x^ax^b\coloneq\sum_{i=1}^d\alpha_i^\star(x)\alpha_i(x).
    \end{align}
    Here $g_{ab}$ is called the Mahalanobis metric \cite{weinberg2009distance}.
    It determines how we measure the distance in coherent space.
    Specifically, we obtain the explicit form of $g_{ab}$ via
    \begin{align*}
        \sum_{i=1}^d\alpha^\star_i(x)\alpha_i(x) = & ~
        \sum_{i=1}^d\frac{1}{2\theta_i}\big[q^i(x)+ip^i(x)\big]\big[q^i(x)-ip^i(x)\big]\\
        = & ~
        \sum_{i=1}^d\frac{1}{2\theta_i}\big[(q^i)^2+(p^i)^2\big]\\
        = & ~
        (x')^\top\Big(\bigoplus_{i=1}^d\frac{1}{2\theta_i}\one_2\Big)x'\\
        = & ~
        (O^\top x)^\top\Big(\bigoplus_{i=1}^d\frac{1}{2\theta_i}\one_2\Big)O^\top x\\
        = & ~
        x^\top \Big[O\Big(\bigoplus_{i=1}^d\frac{1}{2\theta_i}\one_2\Big)O^\top\Big]x\coloneq x^\top gx.
    \end{align*}
    Furthermore, notice that $\Theta^\top\Theta=\bigoplus_{i=1}^d\theta^2_i\one_2$.
    Hence, $(\theta^\top\theta)^{1/2}=O[\bigoplus_{i=1}^d\theta_i\one_2]O^\top$.
    This gives
    \begin{align}
    \label{eq:metric}
        g=\frac{1}{2}(\theta^\top\theta)^{-1/2}.
    \end{align}
    Finally, to derive the explicit form of inner product $\bra{x}\ket{y}$, we consider
    \begin{align*}
        \bra{x}\ket{y} = & ~
        \bra{0}\hat{U}_x^\dagger \hat{U}_y\ket{0}=\bra{0}\hat{U}_{-x}\hat{U}_y\ket{0}=\bra{0}e^{-ix^\top\theta^{-1}\hat{M}}e^{iy^\top\theta^{-1}\hat{M}}\ket{0}\annot{By \cref{def:moyal_weyl_coherent_state}}\\
        = & ~
        \Big\langle0\Big|\exp\Big\{i(y-x)^\top\theta^{-1}\hat{M}+\frac{1}{2}\Big[-ix^\top\theta^{-1}\hat{M},iy^\top\theta^{-1}\hat{M}\Big]+\cdots\Big\}\Big|0\Big\rangle\annot{By \cref{prp:bch_formula}}\\
        = & ~
        \Big\langle0\Big|\exp\Big\{i(y-x)^\top\theta^{-1}\hat{M}+\frac{1}{2}\sum_{a,b,n,m=1}^{2d}x_a(\theta^{-1})_{ab}y_n(\theta^{-1})_{nm}\cdot\big(i\theta^{bm}\hat{\one}\big)\Big\}\Big|0\Big\rangle\annot{By \cref{def:moyal_weyl_plane}}\\
        = & ~
        \Big\langle0\Big|\exp\Big\{i(y-x)^\top\theta^{-1}\hat{M}-\frac{i}{2}\sum_{a,b=1}^{2d}x_a(\theta^{-1})_{ab}y_b\hat{\one}\Big\}\Big|0\Big\rangle\annot{By $(\theta^{-1})_{nm}=-(\theta^{-1})_{mn}$}\\
        = & ~
        e^{-\frac{i}{2}x^\top\theta^{-1}y}\bra{0}e^{i(y-x)^\top\theta^{-1}\hat{M}}\ket{0}\\
        = & ~
        e^{-\frac{i}{2}x^\top\theta^{-1}y}\bra{0}e^{\sum_{i=1}^d\alpha_i(y-x)\hat{a}_i^\dagger}e^{-\frac{1}{2}\|x-y\|_g^2}e^{-\sum_{i=1}^d\alpha^\star_i(y-x)\hat{a}_i}\ket{0}\annot{By \eqref{eq:u_x_coherent_unitary_transform}}\\
        = & ~
        \exp\Big\{-\frac{i}{2}x^\top\theta^{-1}y-\frac{1}{2}\|x-y\|_g^2\Big\}\cdot\bra{0}\ket{0}\annot{By $\hat{a}_i\ket{0}=0$ and $\bra{0}\hat{a}^\dagger_i=0$}\\
        = & ~
        \exp\Big\{-\frac{i}{2}x^\top\theta^{-1}y-\frac{1}{4}(x-y)^\top(\theta^\top\theta)^{-1/2}(x-y)\Big\}\annot{By \eqref{eq:x_norm_with_metric} and \eqref{eq:metric}}.
    \end{align*}
    This completes the proof.
\end{proof}



\subsection{Proof of \texorpdfstring{\cref{prp:normalization_trace}}{}}
\label{sec:proof_normalization_trace}

\begin{proposition}[\cref{prp:normalization_trace} Stated: Identity Matrix and Trace in Moyal-Weyl Quantum Plane \cite{Steinacker:2024unq,ali:1995zz,folland1989harmonic}]
    Consider a Moyal-Weyl quantum plane $\R^{2d}_\theta$ with coherent states $\ket{x}\in\mathcal{H}$.
    We deal with the normalization consistency by counting the number of coherent-states.
    With the non-commutativity scale $\varepsilon=[\det((\theta^\top\theta)^{1/2})]^{1/2d}=(\prod_{i=1}^d\theta_i)^{1/d}$, this gives the identity operator
    \begin{align}
    \label{eq:app_identity_operator_moyal_weyl}
        \hat{\one}=\int_{\R^{2d}_\theta} \hat{P}_x~\dd\mu(x)=\int_{\R^{2d}}\ket{x}\bra{x}~\frac{\dd^{2d}x}{(2\pi\varepsilon)^d}.
    \end{align}
    Furthermore, for $\hat{A}=\sum_{r=1}^m\ket{u_r}\bra{v_r}\in\text{Mat}(\mathcal{H})$ with $\ket{u_r},\ket{v_r}\in\mathcal{H}$, the trace satisfies
    \begin{align*}
        \Tr_{\mathcal{H}}(\hat{A})=\int_{\R^{2d}}\bra{x}\hat{A}\ket{x}~\frac{\dd^{2d}x}{(2\pi\varepsilon)^d}.
    \end{align*}
\end{proposition}

\begin{proof}
    We prove the claims one by one as follows.
    \begin{enumerate}
        \item Consider the normalized vacuum state $\ket{0}$ from \cref{def:moyal_weyl_coherent_state}.
        Then if we set $\dd\mu(x)=\mathcal{A}\dd^{2d}x$ with a normalization constant $\mathcal{A}$, then
        \begin{align*}
            1 = & ~
            \bra{0}\ket{0}=\bra{0}\one\ket{0}=\mathcal{A}\int_{\R^{2d}}\bra{0}\ket{x}\bra{x}\ket{0}~\dd^{2d}x\\
            = & ~
            \mathcal{A}\int_{\R^{2d}}\exp\Big\{-\frac{1}{2}\sum_{a,b=1}^{2d}x_a(\theta^\top\theta)^{-1/2}_{ab}x_b\Big\}~\dd^{2d}x\annot{By \cref{prp:kernel_moyal_weyl_coherent_inner_product}}\\
            = & ~
            \mathcal{A}\times\sqrt{\det\big[2\pi(\theta^\top\theta)^{1/2}\big]}\annot{By Gaussian integral}\\
            = & ~
            \mathcal{A}\times(2\pi\varepsilon)^d.
        \end{align*}
        Hence, we have
        \begin{align*}
            \one=\mathcal{A}\int_{\R^{2d}}\ket{x}\bra{x}~\dd^{2d}x=\int_{\R^{2d}}\ket{x}\bra{x}~\frac{\dd^{2d}x}{(2\pi\varepsilon)^d}.
        \end{align*}

        \item From the scaling \eqref{eq:app_identity_operator_moyal_weyl}, the normalized trace operation follows the same scaling.
    \end{enumerate}
    This completes the proof.
\end{proof}


\subsection{Proof of \texorpdfstring{\cref{thm:quantum_softmax_rbf}}{}}
\label{sec:proof_quantum_softmax_rbf}

\begin{theorem}[\cref{thm:quantum_softmax_rbf} Restated: Quantum Description of RBF and Softmax]
    Consider the projection map $\hat{P}_x=\ket{x}\bra{x}$.
    If we define the scaled projection map as $\hat{\mathcal{P}}_x\coloneq\exp(\|x\|_g^2)\hat{P}_x$ with $\|x\|_g^2\coloneq\frac{1}{2}x^\top(\theta^\top\theta)^{-1/2}x$, 
   then we recover the softmax and RBF kernel by
    \begin{align*}
        \kappa_{\text{Soft}}(q,k_\mu)=\Tr_{\mathcal{H}}\Big(\hat{\mathcal{P}}_q\hat{\mathcal{P}}_{k_\mu}\Big),\quad\kappa_{\text{RBF}}=\Tr_{\mathcal{H}}\Big(\hat{P}_q\hat{P}_{k_\mu}\Big).
    \end{align*}
    In particular, the inverse-temperature and covariance matrix correspond to
    \begin{align*}
        \beta\one_{2d}=(\theta^\top\theta)^{-1/2},\quad\Sigma=(\theta^\top\theta)^{1/2}.
    \end{align*}
\end{theorem}

\begin{proof}
    By definition, we compute
    \begin{align*}
        \Tr_{\mathcal{H}}\Big(\hat{\mathcal{P}}_q\hat{\mathcal{P}}_{k_\mu}\Big) = & ~
        \Tr_{\mathcal{H}}\Big(e^{\|q\|^2_g+\|k_\mu\|_g^2}\ket{q}\bra{q}\ket{k_\mu}\bra{k_\mu}\Big)\\
        = & ~
        \exp\Big(\frac{1}{2}q^\top(\theta^\top\theta)^{-1/2}q+\frac{1}{2}k_\mu^\top(\theta^\top\theta)^{-1/2}k_\mu\Big)\bra{q}\ket{k_\mu}\bra{k_\mu}\ket{q}\\
        = & ~
        \exp\Big(\frac{1}{2}q^\top(\theta^\top\theta)^{-1/2}q+\frac{1}{2}k_\mu^\top(\theta^\top\theta)^{-1/2}k_\mu\Big)\\
        & ~
        \times\exp\Big(-\frac{1}{2}(q-k_\mu)^\top(\theta^\top\theta)^{-1/2}(q-k_\mu)\Big)\annot{By \cref{prp:kernel_moyal_weyl_coherent_inner_product}}\\
        = & ~
        \exp\Big(q^\top(\theta^\top\theta)^{-1/2}k_\mu\Big).
    \end{align*}
    Compared to \eqref{eq:softmax_rbf}, we identify this with softmax attention weight $\kappa_{\text{Soft}}(q,k_\mu)$ with $\beta\one_{2d}=(\theta^\top\theta)^{-1/2}$.
    For the RBF Gaussian attention, we compare the inner product in \cref{prp:kernel_moyal_weyl_coherent_inner_product}.
    This identifies
    \begin{align*}
        |\bra{q}\ket{k_\mu}|^2=\Tr_{\mathcal{H}}\Big(\hat{P}_q\hat{P}_{k_\mu}\Big)=\exp\Big(-\frac{1}{2}(q-k_\mu)^\top(\theta^\top\theta)^{-1/2}(q-k_\mu)\Big)=\kappa_{\text{RBF}}(q,k_\mu).
    \end{align*}
    Then the covariance matrix $\Sigma$ corresponds to $(\theta^\top\theta)^{1/2}$.
    This completes the proof.
\end{proof}

\subsection{Proof of \texorpdfstring{\cref{prp:existence}}{}}
\label{sec:proof_existence}

\begin{proposition}[\cref{prp:existence} Restated: Existence of Moyal-Weyl Realization]
    Consider a RBF Gaussian kernel $\kappa_{\text{RBF}}(x,y)=\exp[-\frac{1}{2}(x-y)^\top\Sigma^{-1}(x-y)]$ with covariance matrix $\Sigma\succ0$.
    Then a Moyal-Weyl duality realization with a linear quantum-state map $x\mapsto\ket{Lx}_\theta$ always exists, where
    \begin{align*}
        L^\top(\theta^\top\theta)^{-1/2}L=\Sigma^{-1},
    \end{align*}
    and $\theta$ is even-dimensional, anti-symmetric and invertible matrix.
    If one requires $L=\one$ in the same dimension, then such a realization exists if and only if every eigenspace of $\Sigma$ is even-dimensional. 
    This existence holds for the softmax kernel $\kappa_{\text{soft}}=\exp(\beta x^\top y)$ as well.
\end{proposition}

\begin{proof}
    Consider an injective linear map $L:\R^n\to\R^N$, where $n$ is arbitrary, and $N$ is even.
    Then a simplest realization is to choose $N=2n$ with 
    \begin{align*}
        \theta=\begin{pmatrix}0&-\one_n\\\one_n&0\end{pmatrix},\quad\text{and }~L=\begin{pmatrix}\Sigma^{-1/2}\\0\end{pmatrix}.
    \end{align*}
    Then
    \begin{align*}
        L^\top(\theta^\top\theta)^{-1/2}L = & ~
        \begin{pmatrix}\Sigma^{-1/2}&0\end{pmatrix}\begin{pmatrix}\one_n&0\\0&\one_n\end{pmatrix}\begin{pmatrix}\Sigma^{-1/2}\\0\end{pmatrix}\\
        = & ~
        \Sigma^{-1/2}\Sigma^{-1/2}\\
        = & ~
        \Sigma^{-1}.
    \end{align*}
    On the other hand, by Moyal-Weyl inner product (\cref{prp:kernel_moyal_weyl_coherent_inner_product}), we have
    \begin{align*}
        {}_\theta\bra{Lx}\ket{Ly}{}_\theta=\exp\Big[-\frac{1}{4}(x-y)^\top L^\top(\theta^\top\theta)^{-1/2}L(x-y)\Big]\exp\big(-\frac{i}{2}x^\top L^\top\theta^{-1}Ly\big).
    \end{align*}
    Together, the square $|{}_\theta\bra{Lx}\ket{Ly}{}_\theta|^2$ gives a Gaussian with $\Sigma^{-1}=L^\top(\theta^\top\theta)^{-1/2}L$.
    Hence, every arbitrary SPD RBF kernel has Moyal-Weyl realization.
    This includes the case where $n$ is odd.
    For a softmax kernel, we construct the scaled projection map as
    \begin{align*}
        \hat{\mathcal{P}}_{Lx}=e^{\|Lx\|^2_g}\hat{P}_{Lx}=\exp\Big[\frac{1}{2}x^\top L^\top(\theta^\top\theta)^{-1/2}Lx\Big]\ket{Lx}{}_\theta{}_\theta\bra{Lx}.
    \end{align*}
    Then the inner product gives
    \begin{align*}
        \Tr_{\mathcal{H}}\Big(\hat{\mathcal{P}}_{Lx}\hat{\mathcal{P}}_{Ly}\Big) = & ~
        \exp\big(\|Lx\|^2_g+\|Ly\|^2_g\big)|{}_\theta\bra{Lx}\ket{Ly}{}_\theta|^2\\
        = & ~
        \exp\Big[\frac{1}{2}x^\top L^\top(\theta^\top\theta)^{-1/2}Lx+\frac{1}{2}y^\top L^\top(\theta^\top\theta)^{-1/2}Ly\Big]\\
        & ~
        \times\exp\Big[-\frac{1}{2}(x-y)^\top L^\top(\theta^\top\theta)^{-1/2}L(x-y)\Big]\\
        = & ~
        \exp\Big[x^\top L^\top(\theta^\top\theta)^{-1/2}Ly\Big].
    \end{align*}
    This gives a softmax with $\beta=L^\top(\theta^\top\theta)^{-1/2}L$.
    Hence, the similar statement holds for softmax kernels.
    This completes the proof.
\end{proof}


\subsection{Proof of \texorpdfstring{\cref{prp:non_uniqueness}}{}}
\label{sec:proof_non_uniqueness}

To prepare our main proof, we first introduce a helper lemma.

\begin{lemma}[Properties of $J\coloneq\theta\Sigma^{-1}$]
\label{lem:J_properties}
    Define $J\coloneq\theta\Sigma^{-1}\in\R^{2d\times 2d}$ with $\Sigma=(\theta^\top\theta)^{1/2}$.
    Then it satisfies
    \begin{align*}
        J^2=-\one_{2d},\quad J^\top=-J,\quad J^\top J=\one_{2d}.
    \end{align*}
    Furthermore, we have $J\Sigma=\Sigma J$.
\end{lemma}

\begin{proof}
    First, notice that $\Sigma=(\theta^\top\theta)^{1/2}=(-\theta^2)^{1/2}$ is a matrix function of $\theta^2$.
    Thus,
    \begin{align}
    \label{eq:theta_sigma_commute}
        \Sigma\theta=\theta\Sigma.
    \end{align}
    Then we compute each equality one-by-one:
    \begin{enumerate}
        \item $J^2=-\one_{2d}$.
        \begin{align*}
            J^2 = & ~
            \theta\Sigma^{-1}\theta\Sigma^{-1}\\
            = & ~
            \theta^2\Sigma^{-2}\annot{By \eqref{eq:theta_sigma_commute}}\\
            = & ~
            -\Sigma^2\Sigma^{-2}=-\one_{2d}.
        \end{align*}

        \item $J^\top=-J$.
        \begin{align*}
            J^\top=(\theta\Sigma^{-1})^\top=(\Sigma^{-1})^\top\theta^\top=-\Sigma^{-1}\theta=-\theta\Sigma^{-1}=-J.
        \end{align*}

        \item $J^\top J=\one_{2d}$
        \begin{align*}
            J^\top J=-J^2=\one_{2d}.
        \end{align*}

    \end{enumerate}
    
    Finally, for $J\Sigma=\Sigma J$, we start from
    \begin{align*}
        J\Sigma-\Sigma J = & ~(\theta\Sigma^{-1})\Sigma-\Sigma(\theta\Sigma^{-1})\\
        = & ~
        \theta(\Sigma^{-1}\Sigma-\Sigma\Sigma^{-1})+(\theta\Sigma-\Sigma\theta)\Sigma^{-1}\\
        = & ~0.\annot{By \eqref{eq:theta_sigma_commute}}
    \end{align*}
    This completes the proof.
\end{proof}

Now we are ready for the main proof.
\begin{proposition}[\cref{prp:non_uniqueness} Restated: Non-Uniqueness of Moyal-Weyl Realization]
    Whenever the realization exists, it is not unique even if $L=\one$.
    That is, if the multiplicity of an eigenvalue $\lambda$ of $\Sigma=(\theta^\top\theta)^{1/2}$ is $2m_\lambda$, then we parametrize the non-commutative structure by
    \begin{align*}
        \Theta(\Sigma)\simeq\prod_\lambda O(2m_\lambda)/U(m_\lambda),\quad\Theta(\Sigma)\coloneq\{\theta:(\theta^\top\theta)^{1/2}=\Sigma\},
    \end{align*}
    where $O(2m_\lambda)$ is a set of orthogonal matrices, and $U(m_\lambda)$ is a set of unitary matrices.
    All these structures induce the same RBF and softmax kernel, but produce distinct phases in general. 
\end{proposition}

\begin{proof}
    Since the covariance matrix $\Sigma$ is invertible and symmetric, we decompose it into eigenspaces, i.e., $\R^{2d}=\bigoplus_\lambda E_\lambda$ with $\dim(E_\lambda)=2m_\lambda$.
    If we define $J=\theta\Sigma^{-1}$, then \cref{lem:J_properties} gives $J\Sigma=\Sigma J$.
    This indicates the simultaneous block diagonalization, i.e., $J=\bigoplus_\lambda J_\lambda$ with $J_\lambda:E_\lambda\to E_\lambda$.
    That means at every $E_\lambda$, we can write $\theta_\lambda=\lambda J_\lambda$ such that $\theta^\top\theta=\bigotimes_\lambda\theta^\top_\lambda\theta_\lambda=\Sigma^2$.
    For each block $J_\lambda$, it satisfies \cref{lem:J_properties}.
    Next, for each $J_\lambda$, we consider an arbitrary rotation matrix $R\in O(2m_\lambda)$, where $O(2m_\lambda)$ is the set of orthogonal matrices with dimension $2m_\lambda$.
    Then after this rotation $J'_\lambda=RJ_\lambda R^\top$, we have
    \begin{align*}
        (J'_\lambda)^2 = & ~
        RJ_\lambda R^\top RJ_\lambda R^\top=RJ_\lambda^2 R^\top=-R\one_{2m_\lambda}R^\top=-\one_{2m_\lambda},\\
        (J'_\lambda)^\top = & ~
        (RJ_\lambda R^\top)^\top=RJ_\lambda^\top R^\top=-RJ_\lambda R^\top=-J'_\lambda,\\
        \text{and}\quad (J'_\lambda)^\top J'_\lambda = & ~
        -J'_\lambda J_\lambda'=\one_{2m_\lambda}.
    \end{align*}
    Hence, rotations do not break the algebraic relation of $J_\lambda$, and hence preserves $\theta_\lambda^\top\theta_\lambda=\lambda^2\one_{2m_\lambda}$. 
    Thus, the choice of $J_\lambda$ is $O(2m_\lambda)$.
    However, it has redundancy from $RJ_\lambda R^\top=J_\lambda$.
    This happens when $R$ is complex linear and preserves inner product, so it is a unitary transformation $R\in U(m_\lambda)$.
    Together, each $E_\lambda$ can choose independent $J_\lambda$ from $O(2m_\lambda)/U(m_\lambda)$.
    Hence, for the domain of $\theta\in\Theta(\Sigma)$ with the whole $\Sigma$, we have the non-commutative structure on
    \begin{align*}
        \theta\in\Theta(\Sigma)\simeq\prod_\lambda\big( O(2m_\lambda)/U(m_\lambda)\big).
    \end{align*}
    Different $\theta$ from this set gives the same $\Sigma=(\theta^\top\theta)^{1/2}$, so RBF and softmax kernel remain the same.
    This suggests the non-uniqueness. 
    However, \cref{prp:kernel_moyal_weyl_coherent_inner_product} shows that the coherent-state phase is $\exp(-\frac{i}{2}x^\top\theta^{-1}y)$, so the phases are different in general.
    This completes the proof.
\end{proof}




\def\arxivfont{\rm}
\bibliographystyle{plainnat}

\bibliography{refs}

\end{document}